\documentclass{article}

     \PassOptionsToPackage{numbers, compress}{natbib}

     \usepackage[final]{neurips_2022}

\usepackage[utf8]{inputenc} 
\usepackage[T1]{fontenc}    
\usepackage{hyperref}       
\usepackage{url}            
\usepackage{booktabs}       
\usepackage{amsfonts}       
\usepackage{nicefrac}       
\usepackage{microtype}      
\usepackage{xcolor}         

\usepackage{microtype}
\usepackage{graphicx}
\usepackage{multirow}
\usepackage{booktabs} 
\usepackage{subfig}
\usepackage{adjustbox}

\usepackage{amsmath}
\usepackage{amssymb}
\usepackage{mathtools}
\usepackage{amsthm}

\theoremstyle{plain}
\newtheorem{theorem}{Theorem}[section]

\theoremstyle{definition}
\newtheorem{definition}[theorem]{Definition}

\theoremstyle{remark}
\newtheorem{remark}[theorem]{Remark}

\usepackage{algorithm,algpseudocode}

\title{A Geometric Perspective on Variational Autoencoders}

\author{%
  Clément Chadebec\\
  Université Paris Cité, INRIA, Inserm, SU\\
  Centre de Recherche des Cordeliers\\
  \texttt{clement.chadebec@inria.fr} \\
  \And
  Stéphanie Allassonnière\\
  Université Paris Cité, INRIA, Inserm, SU \\
  Centre de Recherche des Cordeliers\\
  \texttt{stephanie.allassonniere@inria.fr}
}

\begin{document}

\maketitle

\begin{abstract}
This paper introduces a new interpretation of the Variational Autoencoder framework by taking a fully geometric point of view. We argue that vanilla VAE models unveil naturally a Riemannian structure in their latent space and that taking into consideration those geometrical aspects can lead to better interpolations and an improved generation procedure. This new proposed sampling method consists in sampling from the uniform distribution deriving intrinsically from the learned Riemannian latent space and we show that using this scheme can make a vanilla VAE competitive and even better than more advanced versions on several benchmark datasets. Since generative models are known to be sensitive to the number of training samples we also stress the method's robustness in the low data regime.
\end{abstract}

\section{Introduction}

Variational Autoencoders (VAE) \citep{kingma_auto-encoding_2014, rezende_stochastic_2014} are powerful generative models that map complex input data in a much lower dimensional space referred to as the latent space while driving the latent variables to follow a given prior distribution. Their simplicity to use in practice has made them very attractive models to perform various tasks such as high-fidelity image generation \citep{razavi_generating_2019}, speech modeling \citep{blaauw_modeling_2016}, clustering \citep{yang_deep_2019} or data augmentation \citep{chadebec_data_2021}.

Nonetheless, when taken in their simplest version, it was noted that these models produce blurry samples on image generation tasks most of the time. This undesired behavior may be due to several limitations of the VAE framework. First, the training of a VAE aims at maximizing the Evidence Lower BOund (ELBO) which is only a lower bound on the true likelihood and so does not ensure that we are always actually improving the true objective \citep{burda_importance_2016, alemi_deep_2016, higgins_beta-vae_2017, cremer_inference_2018, zhang_advances_2018}. Second, the prior distribution may be too simplistic \citep{dai_diagnosing_2018} leading to poor data generation and there exists no guarantee that the actual distribution of the latent code will match a given prior distribution inducing distribution mismatch \citep{connor2021variational}. Hence, trying to tackle those limitations through richer posterior distributions \citep{salimans_markov_2015, rezende_variational_2015} or better priors \citep{tomczak_vae_2018} represents a major part of the proposed improvements over the past few years. However, the tractability of the ELBO constrains the choice in distributions and so finding a trade-off between model expressiveness and tractability remains crucial. In this paper, we take a rather different approach and focus on the geometrical aspects a vanilla VAE is able to capture in its latent space. In particular, we propose the following contributions:

\begin{itemize}
    \item We show that VAEs unveil naturally a latent space with a structure that can be modeled as a Riemannian manifold through the learned covariance matrices in the variational posterior distributions and that such modeling can lead to better interpolations.
    \item We propose a natural sampling scheme consisting in sampling from a uniform distribution defined on the learned manifold and given by the Riemannian metric. We show that this procedure improves the generation process from a \emph{vanilla} VAE significantly without complexifying the model nor the training. The proposed sampling method outperforms more advanced VAE models in terms of Frechet Inception Distance \citep{heusel_gans_2017} and Precision and Recall \citep{sajjadi_assessing_2019} scores on four benchmark datasets. We also discuss and show that it can benefit more recent VAEs as well.
    \item We show that the method appears robust to dataset size changes and outperforms even more strongly peers when only \emph{smaller} sample sizes are considered.  
    \item We discuss the link of the proposed metric to the \emph{pull-back} metric. 
\end{itemize}

\section{Variational autoencoders}

Considering that we are given $x \in \mathbb{R}^D$ a set of data points deriving from an unknown distribution $p(x)$, a VAE aims at inferring $p$ with a parametric model $\{p_{\theta}, \theta \in \Theta\}$ using a maximum likelihood estimator. A key assumption behind the VAE is to assume that the generation process involves latent variables $z$ living in a lower dimensional space such that the generative model writes
\[
    z \sim p(z) \hspace{5mm};\hspace{5mm} x \sim p_{\theta}(x|z)\,,
\]
where $p$ is a prior distribution over the latent variables often taken as a standard Gaussian and $p_{\theta}(x|z)$ is referred to as the decoder and is most of the time taken as a parametric distribution the parameters of which are estimated using neural networks. Hence, the likelihood $p_{\theta}$ writes:
\begin{equation}\label{Eq:Objective}
    p_{\theta}(x) = \int \limits_{\mathcal{Z}} p_{\theta}(x|z)p(z) dz \,.
\end{equation}
As this integral is most of the time intractable so is $p_{\theta}(z|x)$, the posterior distribution. Hence, Variational Inference~\citep{jordan_introduction_1999} is used and a simple parametrized variational distribution $q_{\phi}(z|x)$ is introduced to approximate the posterior $p_{\theta}(z|x)$. $q_{\phi}(z|x)$ is referred to as the \emph{encoder} and, in the vanilla VAE, $q_{\phi}$ is chosen as a multivariate Gaussian whose parameters $\mu_{\phi}$ and $\mathbf{\Sigma}_{\phi}$ are again given by neural networks. An unbiased estimate $\hat{p}_{\theta}$ of the likelihood $ p_{\theta}(x)$ can then be derived using importance sampling with $q_{\phi}(z|x)$ and the ELBO objective follows using Jensen's inequality:
\begin{equation}\label{eq: ELBO}
      \log~p_{\theta}(x) = \log \mathbb{E}_{z \sim q_{\phi}}\big[\hat{p}_{\theta}\big]
                         \geq \mathbb{E}_{z \sim q_{\phi}}\big[\log \hat{p}_{\theta}\big] \geq~
\mathbb{E}_{z \sim q_{\phi}} \log p_{\theta}(x| z) - \mathrm{KL}( q_{\phi}(z|x) \Vert p(z)) = \mathcal{L}
\end{equation}
The ELBO is now tractable since both $q_{\phi}(z|x)$ and $p_{\theta}(x|z)$ are known and so can be optimized with respect to the \textit{encoder} and \textit{decoder} parameters.
\begin{remark}\label{Rem: remarrk 1}
In practice, $p_{\theta}(x|z)$ is chosen depending on the modeling of the input data but is often taken as a simple distribution ($\emph{e.g}$ fixed variance Gaussian, Bernoulli ...) and a weight $\beta$ can be applied to balance the weight of the $\mathrm{KL}$ term \citep{higgins_beta-vae_2017}. Hence, the ELBO can also be seen as a two terms objective \citep{ghosh_variational_2020}. The first one is a reconstruction term given by $p_{\theta}(x|z)$ while the second one is a regularizer given by the KL between the variational posterior $q_{\phi}$ and the prior $p$. For instance, in the case of a fixed variance Gaussian for $p_{\theta}(x|z)$ we have
\begin{equation}\label{eq: ELBO rewrite}
    \mathcal{L}_{\text{REC}} = \Vert x - \mu_{\theta}(z) \Vert_2^2,\hspace{2mm} \mathcal{L}_{\text{REG}} = \beta \cdot \mathrm{KL}( q_{\phi}(z|x) \Vert p(z))\,.
\end{equation}
\end{remark}

\section{Related work}

A natural way to improve the generation from VAEs consists in trying to use more complex priors \citep{hoffman_elbo_2016} than the standard Gaussian distribution used in the initial version such that they better match the true distribution of the latent codes. For instance, using a Mixture of Gaussian \citep{nalisnick_approximate_2016, dilokthanakul_deep_2017} or a Variational Mixture of Posterior (VAMP) \citep{tomczak_vae_2018} as priors was proposed. In the same vein, hierarchical latent variable models \citep{sonderby_ladder_2016, klushyn_learning_2019} or prior learning \citep{chen_variational_2016, aneja_ncp-vae_2020} have recently emerged and aimed at finding the best suited prior distribution for a given dataset. Acceptance/rejection sampling method was also proposed to try to improve the expressiveness of the prior distribution \citep{bauer_resampled_2019}. Some recent works linking energy-based models (EBM) and VAEs \citep{xiao2020vaebm} or modeling the prior as an EBM \citep{pang_learning_2020} have demonstrated promising results and are also worth citing.

On the ground that the latent space must adapt to the data as well, \emph{geometry-aware} latent space modelings as hypershpere \citep{davidson_hyperspherical_2018}, torus \citep{falorsi_explorations_2018} or Poincaré disk \citep{mathieu_continuous_2019} or discrete latent representations \citep{razavi_generating_2019} were proposed. Other recent contributions proposed to see the latent space as a Riemannian manifold where the Riemannian metric is given by the Jacobian of the generator function \citep{arvanitidis_latent_2018, chen_metrics_2018, shao_riemannian_2018}. This metric was then used directly within the prior modeled by Brownian motions \citep{kalatzis_variational_2020}. Others proposed to learn the metric directly from the data throughout training thanks to \emph{geometry-aware} normalizing flows \citep{chadebec_geometry-aware_2020} or learn the latent structure of the data using transport operators \citep{connor2021variational}. While these geometry-based methods show interesting properties of the learned latent space they either require the computation of a time consuming model-dependent function, the Jacobian, or add further parameters to the model to learn the metric or transport operators adding some computational burden. 

Arguing that VAEs are essentially autoencoders regularized with a Gaussian noise, \citet{ghosh_variational_2020} proposed another interesting interpretation of the VAE framework and showed that other types of regularization may be of interest as well. Since the generation process from these autoencoders is no longer relying on the prior distribution, the authors proposed to use ex-post density estimation by fitting simple distributions such as a Gaussian mixture in the latent space. While this paves the way for consideration of other ways to generate data, it mainly reduces the VAE framework to an autoencoder while we believe that it can also unveil interesting geometrical aspects. 

Another widely discussed improvement of the model consists in trying to tweak the approximate posterior in the ELBO so that it better matches the true posterior using MCMC methods \citep{salimans_markov_2015} or normalizing flows \citep{rezende_variational_2015}. For instance, methods using Hamiltonian equations in the flows to target the true posterior \citep{caterini_hamiltonian_2018} were proposed.

Finally, while discussing the potential link between PCA and autoencoders some intuitions arose on the impact of both the intrinsic structure of the variance of the data \citep{rakowski2021disentanglement} and the shape of the covariance matrices in the posterior distributions \citep{rolinek2019variational} on disentanglement in the latent space. We also believe that these covariance matrices indeed play a crucial role in the modeling of the latent space but in this paper, we instead propose to see their inverse as the value of a Riemannian metric.

\section{Proposed method}

In this section, we show that a vanilla VAE unveils naturally a Riemannian structure in its latent space through the learned covariance matrices in the variational posterior distribution. We then propose a new natural generation scheme guided by this estimated geometry and consisting in sampling from a uniform distribution deriving intrinsically from the learned Riemannian manifold.

\subsection{A word on Riemannian geometry}\label{Sec: Word on Riemann}
First, we briefly recall some basic elements of Riemannian geometry needed in the rest of the paper. A more detailed discussion on Riemannian manifolds may be found in Appendix~\ref{appA}. A $d$-dimensional manifold $\mathcal{M}$ is a manifold which is locally homeomorphic to a $d$-dimensional Euclidean space. If the manifold $\mathcal{M}$ is further differentiable it possesses a tangent space $T_{z}$ at any $z \in \mathcal{M}$ composed of the tangent vectors of the curves passing by $z$. If $\mathcal{M}$ is equipped with a smooth inner product $g: z\to \langle \cdot | \cdot \rangle_{z}$ defined on its tangent space $T_{z}$ for any $z \in \mathcal{M}$ then $\mathcal{M}$ is called a Riemannian manifold and $g$ is the associated Riemannian metric. Then, a local representation of $g$ at any $z \in \mathcal{M}$ is given by the positive definite matrix $\mathbf{G}(z)$ (See Appendix~\ref{appA}). If $\mathcal{M}$ is connected, a Riemannian distance between two points $z_1$, $z_2$ of $\mathcal{M}$ can be defined
\begin{equation}\label{eq: Riemannian distance}
    \mathrm{dist}_{\mathbf{G}}(z_1, z_2) = \inf \limits _{\gamma} \int \limits _a ^b \sqrt{\dot{\gamma}(t)^{\top} \mathbf{G}(\gamma(t))\dot{\gamma}(t)} dt = \inf  \limits _{\gamma} L(\gamma)\hspace{1em}\mathrm{s.t.}\hspace{1em}z_1=\gamma(a), z_2 = \gamma(b)\,,
\end{equation}
where $L$ is the length of curves $\gamma: \mathbb{R} \to \mathcal{M}$ traveling from $z_1$ to $z_2$. Curves minimizing $L$ and parametrized proportionally to the arc length are \emph{geodesic}. The manifold $\mathcal{M}$ is said to be \textit{geodesically complete} if all geodesic curves can be extended to $\mathbb{R}$. In an Euclidean space, $\mathbf{G}$ reduces to $I_d$ and the distance becomes the classic Euclidean one. A simple extension of this Euclidean framework consists in assuming that the metric is given by a constant positive definite matrix $\mathbf{\Sigma}$ different from $I_d$. In such a case the induced Riemannian distance is the well-known Mahalanobis distance $\mathrm{dist}_{\mathbf{\Sigma}}(z_1, z_2) = \sqrt{(z_2-z_1)^{\top}\mathbf{\Sigma}(z_2-z_1)}$~.
\subsection{The Riemannian Gaussian distribution}
\label{sec: The Riemannian Gaussian distribution}
Given the Riemannian manifold $\mathcal{M}$ endowed with the Riemannian metric $\mathbf{G}$ and a chart $z$, an infinitesimal volume element may be defined on each tangent space $T_{z}$ of the manifold $\mathcal{M}$ as follows
\begin{equation}\label{eq: Riemannian volume element}
    d \mathcal{M}_z = \sqrt{\det \mathbf{G}(z)} dz\,,
\end{equation}
with $dz$ being the Lebesgue measure. This defines a canonical measure on the manifold and allows to extend the notion of random variables to Riemannian manifolds whose density can be defined with respect to that Riemannian measure (see Appendix~\ref{appA}). Hence, a Riemannian Gaussian distribution on $\mathcal{M}$ can be defined using the Riemannian distance of Eq.~\eqref{eq: Riemannian distance} instead of the Euclidean one.
\begin{equation}\label{eq: Riemannian Gaussian}
    \mathcal{N}_{\mathrm{riem}}(z|\sigma, \mu) = \frac{1}{C} \exp\Big(- \frac{\mathrm{dist}_{\mathbf{G}}(z, \mu)^2}{2 \sigma}\Big) \,, \hspace{2em} C= \int \limits_{\mathcal{M}} \exp\Big(- \frac{\mathrm{dist}_{\mathbf{G}}(z, \mu)^2}{2 \sigma}\Big) d \mathcal{M}_{z}\,,
    \end{equation}
where $d \mathcal{M}_{z}$ is the volume element defined in Eq.~\eqref{eq: Riemannian volume element}. Thus, a multivariate normal distribution with covariance matrix $\mathbf{\Sigma}$ is only a specific case of the Riemannian distribution with $\sigma = 1$ and defined on the manifold $\mathcal{M}=(\mathbb{R}^d, \mathbf{G})$ where $\mathbf{G}$ is the constant Riemannian metric $\mathbf{G}(z) = \mathbf{\Sigma}^{-1},~\forall z \in \mathcal{M}$. 

\subsection{Geometrical interpretation of the VAE framework}

Within the VAE framework, the variational distribution $q_{\phi}(z|x)$ is often chosen as a simple multivariate Gaussian distribution defined on $\mathbb{R}^d$ with $d$ being the latent space dimension. Hence, as explained in the previous section, given an input data point $x_i$, the posterior $q_{\phi}(z|x_i) = \mathcal{N}(\mu(x_i), \mathbf{\Sigma}(x_i))$ can also be seen as a Riemannian Gaussian distribution where the Riemannian distance is simply the distance with respect to the metric tensor $\mathbf{\Sigma}^{-1}(x_i)$. Hence, the VAE framework can be seen as follow: As with an autoencoder, the VAE provides a code $\mu(x_i)$ which is a lower dimensional representation of an input data point $x_i$. However, it also gives a tensor $\mathbf{\Sigma}^{-1}(x_i)$ depending on $x_i$ which can be seen as the value of a Riemannian metric $\mathbf{G}$ at $\mu(x_i)$ \emph{i.e.}
\[
    \mathbf{G}(\mu(x_i)) = \mathbf{\Sigma}^{-1}(x_i)\,.
\]
This metric is crucial since it impacts the notion of distance in the latent space now seen as the Riemannian manifold $\mathcal{M} = (\mathbb{R}^d, \mathbf{G})$ and so changes the directions that are favored in the sampling from the posterior distribution $q_{\phi}(z|x_i)$.
Then, a sample $z$ is drawn from a standard (\emph{i.e.} $\sigma=1$ in Eq.~\eqref{eq: Riemannian Gaussian}) Riemannian Gaussian distribution and fed to the decoder. Since we only have access to a finite number of metric tensors $\mathbf{\Sigma}^{-1}(x_i)$, as a first approximation the VAE model assumes that the metric is locally constant close to $\mu(x_i)$ and so the Riemannian distance reduces to the Mahalanobis distance in the posterior distribution. This drastically simplifies the training process since now Riemannian distances have closed form and so are easily computable. Interestingly, the VAE framework will impose through the ELBO expression given in Eq.~\eqref{eq: ELBO rewrite}, that $z$ gives a sample $x \sim p_{\theta}(x|z)$ close to $x_i$ when decoded. Since $z$ has a probability density function imposing higher probability for samples having the smallest Riemannian distance to $\mu$, the VAE imposes in a way that latent variables that are close in the latent space with respect to the metric $\mathbf{G}$ will also provide samples that are close in the data space $\mathcal{X}$ in terms of L2 distance as noticed in Remark.~\ref{Rem: remarrk 1}. Noteworthy is that the latter distance can be amended through the choice of the decoder $p_{\theta}(x|z)$. This is an interesting property since it allows the VAE to directly link the learned Riemannian distance in the latent space to the distance in the data space. The regularization term in Eq.~\eqref{eq: ELBO rewrite} ensures that the covariance matrices do not collapse to $\mathbf{0}_d$ and constraints the latent codes to remain close to the origin easing optimization. Finally, at the end of training, we have a lower dimensional representation of the training data given by the means of the posteriors $\mu(x_i)$ and a family of metric tensors ($\mathbf{G}_i = \mathbf{\Sigma}^{-1}(x_i)$) corresponding to the value of a Riemannian metric defined locally on the latent space. Inspired from  \citep{hauberg_geometric_2012}, we propose to build a smooth continuous Riemannian metric defined on the entire latent space as follows:
\begin{equation}\label{eg: Riemannian metric}
    \mathbf{G}(z) = \sum \limits_{i=1}^N \mathbf{\Sigma}^{-1}(x_i) \cdot \omega_i(z) + \lambda \cdot e^{-\tau \lVert z\rVert_2^2} \cdot I_d \,,\hspace{1em}
    \omega_i(z) = \exp \Bigg ( -\frac{\mathrm{dist}_{\mathbf{\Sigma}^{-1}(x_i)}(z, \mu(x_i))^2}{\rho^2}\Bigg)\,,
\end{equation}
where $\mathrm{dist}_{\mathbf{\Sigma}^{-1}(x_i)}(z, \mu(x_i))^2 = (z - \mu(x_i))^{\top} \mathbf{\Sigma}^{-1}(x_i)(z - \mu(x_i))$ is the Riemannian distance between $z$ and $\mu(x_i)$ with respect to the locally constant metric $\mathbf{G}(\mu(x_i))=\mathbf{\Sigma}^{-1}(x_i)$. Since the sum in Eq.~\eqref{eg: Riemannian metric} is made on the total number of training samples $N$, the number of centroids ($\mu(x_i)$) and so of reference metric tensors can be decreased for huge datasets by selecting only $k < N$ elements\footnote{This may be performed with $k$-medoids algorithm.} and increasing $\rho$ to reduce memory usage. We provide an ablation study on the impact of $\lambda$, the number of centroids and their choice along with a discussion on the choice for $\rho$ in Appendix~\ref{appF}. The parameter $\tau$ is only there to ensure that the volume of $ (\mathbb{R}^d, \textbf{G})$ is finite, a property that is needed in Sec.~\ref{Sec: Geometry-Aware Sampling}, and its value can be set as close as desired to zero so the norm of $z$ does not influence the metric close to the centroids. In practice, it is set below computer precision (\emph{i.e.} $\tau\approx 0$). Rigorously, the metric defined in Eq.~\eqref{eg: Riemannian metric} should have been used during the training process. Nonetheless, this would have made the training longer and trickier since it would involve i) the computation of Riemannian distances that have no longer closed form and so make the resolution of the optimization problem in Eq.~\eqref{eq: Riemannian distance} needed, ii) the sampling from Eq.~\eqref{eq: Riemannian Gaussian} which is not trivial and iii) the computation of the regularization term. Moreover, for small values of $\beta$ in Eq.~\eqref{eq: ELBO rewrite}, the samples generated from the variational distribution $z \sim \mathcal{N}(\mu(x_i), \mathbf{\Sigma}(x_i))$ can be assumed to be concentrated around $\mu(x_i)$ and so we have the following first-order Taylor expansion around $\mu(x_i)$ 
\begin{equation}\label{Taylor expansion of metric}
    \mathbf{G}(z) \approx \mathbf{\Sigma}^{-1}(x_i) + \sum \limits_{j=1, j\neq i}^N \mathbf{\Sigma}^{-1}(x_j) \cdot \underbrace{\omega_j(\mu(x_i))}_{\approx 0} +
    \mathbf{\Sigma}^{-1}(x_i) \cdot \underbrace{\mathbf{J}_{\omega_i}(\mu(x_i))}_{=0}(z-\mu(x_i))\,,
\end{equation}
where $\mathbf{J}_{\omega_i}(\mu(x_i))$ is the Jacobian of the interpolant $\omega_i$ evaluated at $\mu(x_i)$. Note that we have further assumed small enough $\rho$ and $\lambda$ to neglect the influence of the other $\mathbf{\Sigma}(x_j)$ in Eq.~\eqref{eg: Riemannian metric}. Hence by approximating the value of the metric during training by its value at $\mu(x_i)$ (\emph{i.e.} $\mathbf{\Sigma}^{-1}(x_i)$), the VAE training remains unchanged, stable and computationally reasonable since Riemannian Gaussians become multivariate Gaussians in $q_{\phi}(z|x)$ as explained before. Noteworthy is the fact that following the discussion on the role of the KL loss in the VAE framework and the experiments conducted in \citep{ghosh_variational_2020}, in our vision of the VAE, the prior distribution is only seen as a regularizer though the KL term and other latent space regularization schemes may have been also envisioned. In the following, we keep the proposed vision and do not amend the training.

\subsection{Link with the \emph{pull-back} metric}
It has been shown that a natural Riemannian metric on the latent space of generative models can be the \emph{pull-back} metric given by $\mathbf{G}(z) = \mathbf{J}_g(z)^{\top}\mathbf{J}_g(z)$ \citep{arvanitidis_latent_2018} and induced by the decoder mapping $g: \mathbb{R}^d \to \mathbb{R}^D$ outputing the parameters of the conditional distribution $p_{\theta}(x|z)$. Actually, there exists a strong relation linking the metric proposed in this paper to the \emph{pull-back} metric. Indeed, assuming that samples from the variational posterior $z \sim q_{\phi}(z|x) = \mathcal{N}(\mu(x), \mathbf{\Sigma}(x))$ remain close to $\mu(x)$ (\emph{e.g.} by setting a small $\beta$ in Eq.~\eqref{eq: ELBO rewrite}) allows to consider an approximation of the log density $h(z) \coloneqq \log p_{\theta}(x| z)$ next to $\mu(x)$ for a given $x$ \citep{kumar2020implicit}.
\begin{equation*}
        h(z) \approx ~h(\mu(x)) + \mathbf{J}_{h}(\mu(x))(z - \mu(x))~+
        ~\frac{1}{2}(z - \mu(x))^{\top} \mathbf{H}_{h}(\mu(x))(z - \mu(x))\,,
\end{equation*}
where $\mathbf{J}_{h}(\mu(x))$ is the Jacobian and $\mathbf{H}_{h}(\mu(x))$ is the Hessian of h. Using this and remarking that
\begin{equation*}\label{eq: ELBO approx}
    \mathbb{E}_{z \sim q_{\phi}} \Big [\mathbf{J}_{h}(\mu)(z - \mu)\Big ]=0\, \hspace{1em} \text{and} \hspace{1em}
        \mathbb{E}_{z \sim q_{\phi}} \Big [ (z - \mu)^{\top} \mathbf{H}_{h}(\mu)(z - \mu) \Big ]= \mathrm{Tr}(\mathbf{H}_{h}(\mu)\mathbf{\Sigma})\,,
\end{equation*}
  makes the ELBO in Eq.~\eqref{eq: ELBO rewrite} write:
\begin{equation}
\begin{aligned}
    \mathcal{L} &\approx h(\mu(x)) + \frac{1}{2}\mathrm{Tr}(\mathbf{H}_{h}(\mu(x))\mathbf{\Sigma}(x))
    - \beta~\mathrm{KL}(q_{\phi}(z|x)\Vert p(z))\,.
\end{aligned}
\end{equation}
Assuming a standard Gaussian prior, \citet{kumar2020implicit} showed that $\widetilde{\mathbf{\Sigma}}$ maximizing the ELBO is
\begin{equation}
    \begin{aligned}
        \widetilde{\mathbf{\Sigma}}(x) = \Big (I_d - \frac{1}{\beta}\mathbf{H}_{h}(\mu(x)) \Big)^{-1}\,,
    \end{aligned}
\end{equation}
and if we further assume some regularity on the neural networks used for the decoder mapping $g$ (\emph{e.g.} piece-wise linear activation functions) we have
\begin{equation}\label{eq: sigma opt}
    \begin{aligned}
         \widetilde{\mathbf{\Sigma}}(x) = \Big (I_d - \frac{1}{\beta}\mathbf{J}_g(\mu(x))^{\top}\mathbf{H}_{p}(g(\mu(x)))\mathbf{J}_g(\mu(x)) \Big)^{-1}\,,
    \end{aligned}
\end{equation}
where $\mathbf{H}_{p}(g(\mu(x)))$ is the Hessian of $\log p_{\theta}(x;g(z))$. A standard case for the VAE is to assume that $ p_{\theta}(x|z)= \mathcal{N}(\mu_{\theta}(z), \sigma\cdot I_D)$ and so gives $\mathbf{H}_{p}(g(\mu(x))) = -\frac{1}{\sigma}\cdot I_D$~. If we further set $\sigma = \frac{1}{\beta}$, Eq.~\eqref{eq: sigma opt} gives a relation between the \emph{pull-back} and the metric we propose
\[
    \widetilde{\mathbf{\Sigma}}^{-1}(x) = \mathbf{J}_g(\mu(x))^{\top}\mathbf{J}_g(\mu(x)) + I_d\,.
\]
Hence, the proposed metric is closely linked to the \emph{pull-back} metric and may be useful to approximate it (at least close to the $\mu(x)$) and so avoid the computation of a potentially costly function.

\subsection{Geometry-aware sampling}\label{Sec: Geometry-Aware Sampling}
Assuming that the VAE has learned a latent representation of the data in a space seen as a Riemannian manifold, we propose to exploit this strong property to enhance the generation procedure. A natural way to sample from such a latent space would consist in sampling from the uniform distribution intrinsically defined on the learned manifold. Similar to the Gaussian distribution presented in Sec.~\ref{sec: The Riemannian Gaussian distribution}, the notion of uniform distribution can indeed be extended to Riemannian manifolds.  Given a set $\mathcal{A} \subset \mathcal{M}$ having a finite volume, a Riemannian uniform distribution on $\mathcal{A}$ writes \citep{pennec_intrinsic_2006}
\[
    p_{\mathcal{A}}(z) = \frac{\mathbf{1}_{\mathcal{A}}(z) }{\mathrm{Vol(\mathcal{A})}} = \frac{\mathbf{1}_{\mathcal{A}}(z)}{\int _{\mathcal{M}} \mathbf{1}_{\mathcal{A}}(z) d \mathcal{M}_z}\,.
\]
This density is taken with respect to $d\mathcal{M}_z$, the Riemannian measure but using Eq.~\eqref{eq: Riemannian volume element} and a coordinate system $z$ allows to obtain a pdf defined with respect to the Lebesgue one. Moreover, since the volume of the whole manifold $\mathcal{M}=(\mathbb{R}^d, \mathbf{G})$ is finite, we can now define a \emph{uniform distribution} on $\mathcal{M}$
\[
    \mathcal{U}_{\text{Riem}}(z) = \frac{\sqrt{\det \mathbf{G}(z)}}{\int_{\mathbb{R}^d} \sqrt{\det\mathbf{G}(z)dz}}\,.
\]
Since the Riemannian metric has a closed form expression given by Eq.~\eqref{eg: Riemannian metric} sampling from this distribution is quite easy and may be performed using the HMC sampler \citep{neal_hamiltonian_2005} for instance. Now we are able to sample from the intrinsic uniform distribution which is a natural way of exploring the estimated manifold and the sampling is guided by the geometry of the latent space. A discussion on practical outcomes can be found in Appendix~\ref{appB}. Noteworthy is the fact that this approach can also be easily applied to more recent VAE models having a Gaussian posterior (\emph{e.g.} \citep{burda_importance_2016, larsen_autoencoding_2015, tomczak_vae_2018,makhzani2015adversarial}). We detail this and show that the proposed method can also benefit these models in Appendix~\ref{appG}.

\begin{figure}[t]
    \centering
    \captionsetup[subfigure]{position=above, labelformat = empty}
    \subfloat{\includegraphics[width=1.6in]{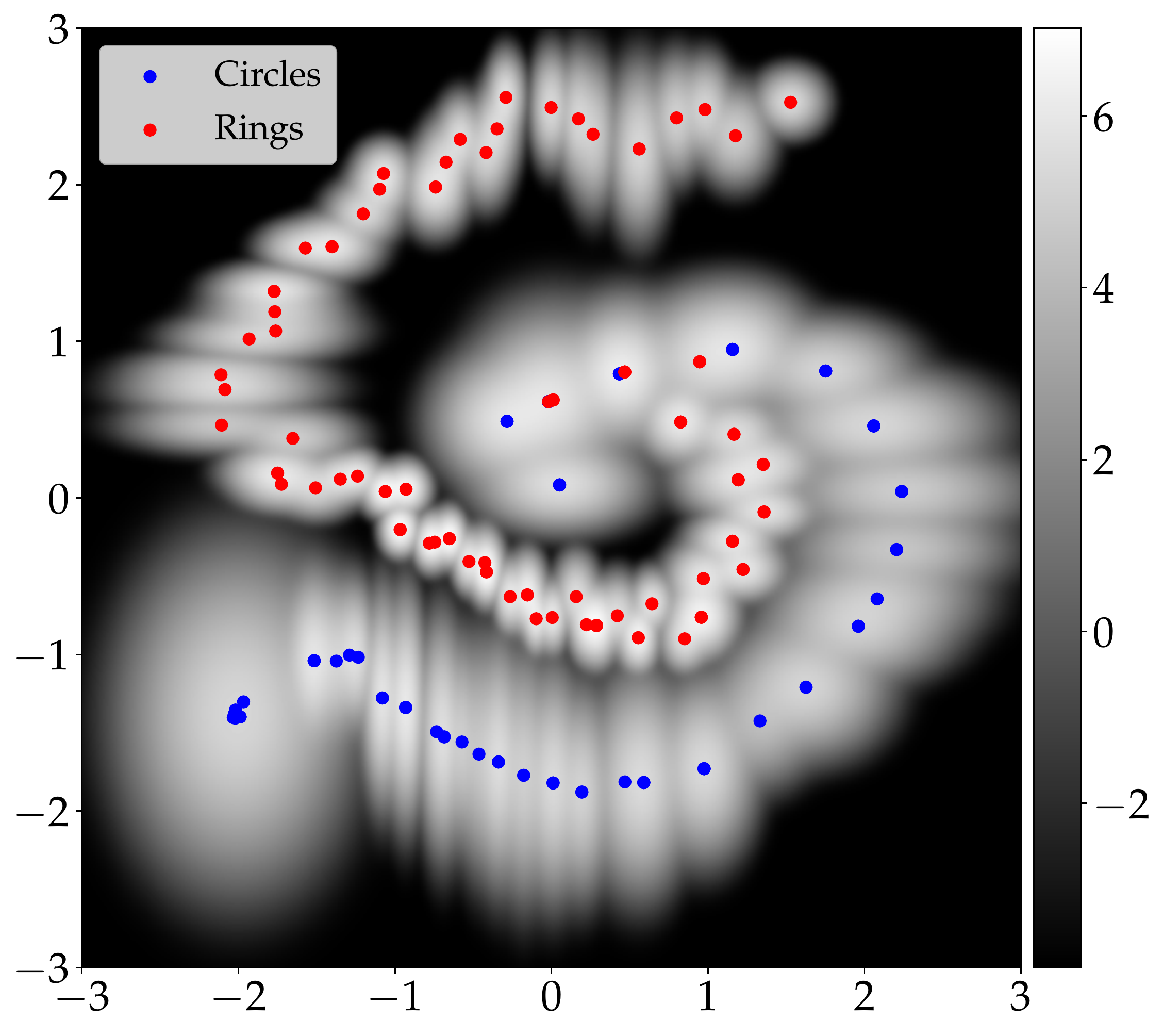}}
    \subfloat[(a)]{\includegraphics[width=1.6in]{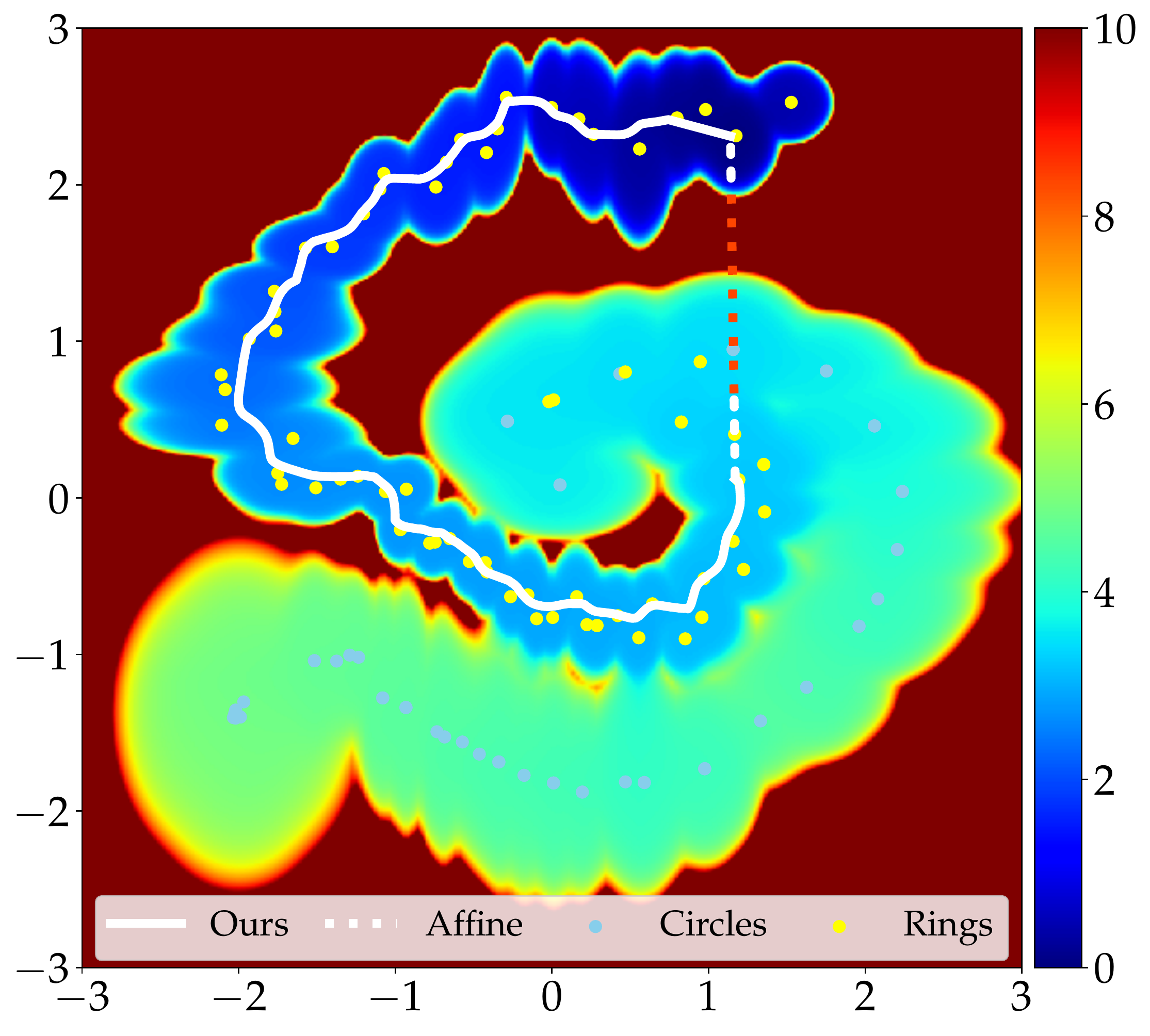}}
    \subfloat[(b)]{\includegraphics[width=1.6in]{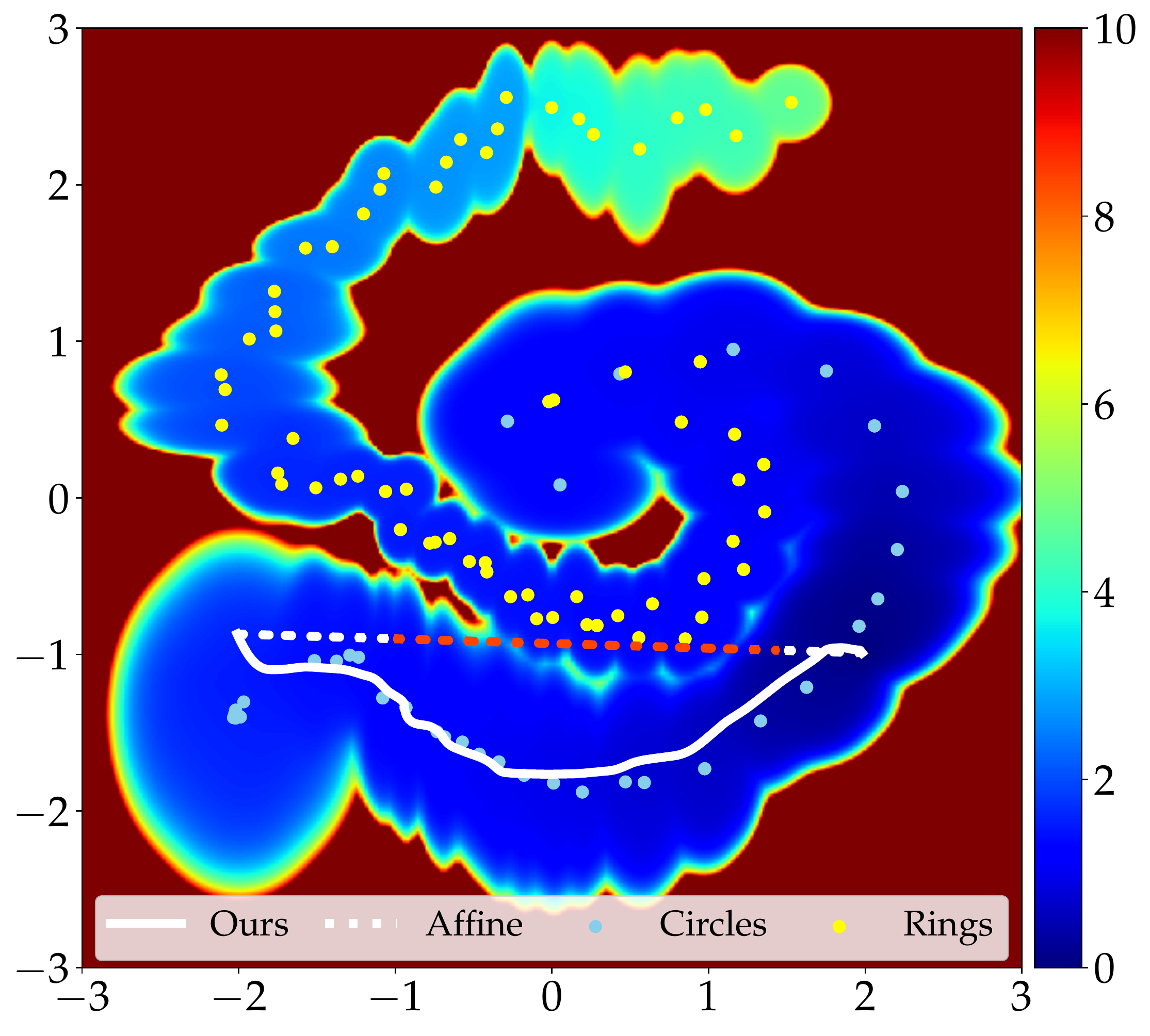}}
    \vspace{-1.3em}
    \subfloat{\includegraphics[width=4.8in]{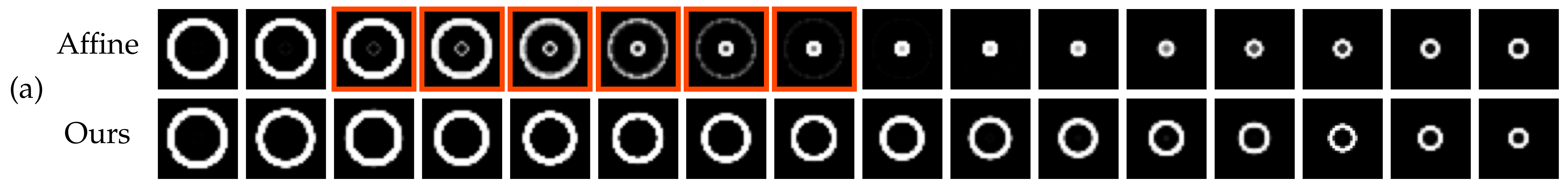}}
    \vspace{-1.3em}
    \subfloat{\includegraphics[width=4.8in]{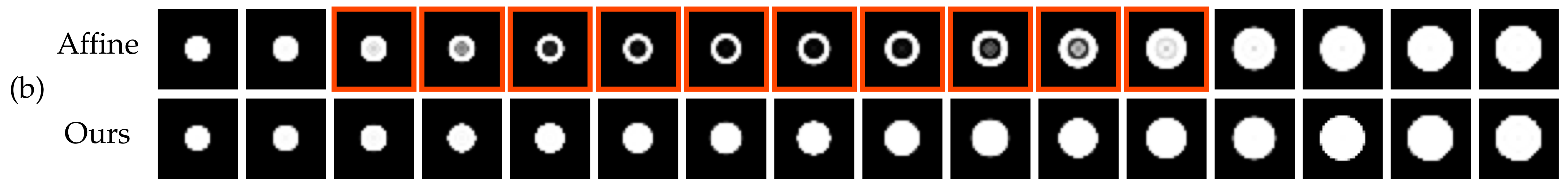}}
    \caption{\emph{Top left:} Visualization and interpolation in a 2D latent space learned by a VAE trained with binary images of rings and disks. The log of the metric volume element $\sqrt{\det \mathbf{G}(z)}$ (proportional to the log of the density we propose to sample from) is shown in gray scale. \emph{Top middle and right}: Riemannian distance from a starting point (color maps). The dashed lines are affine interpolations between two points in the latent space and the solid ones are obtained by solving Eq.~\eqref{eq: potential minimizer}. \emph{Bottom:} Decoded samples along the interpolation curves.}
    \label{fig: toy disks}
    \end{figure}

\subsection{Illustration on a toy dataset}

The usefulness of such sampling procedure can be observed in Figure~\ref{fig: toy disks} where a vanilla VAE was trained with a toy dataset composed of binary images of disks and rings of different size and thickness (example inspired by \cite{chadebec_data_2021}). On the left is presented the learned latent space along with the embedded training points given by the colored dots. The log of the metric volume element is given in gray scale. In this example, we clearly see a geometrical structure appearing since the disks and rings seem to wrap around each other. Obviously, sampling using the prior (taken as a $\mathcal{N}(0, I_d)$) in such a case is far from being optimal since the sampling will be performed regardless of the underlying distribution of the latent variables and so will create irrelevant samples. To further illustrate this, we propose to interpolate between points in the latent space using different cost functions. Dashed lines represent affine interpolations while the solid ones show interpolations aiming at minimizing the potential $V(z) = (\sqrt{\det \mathbf{G}(z)})^{-1}$ all along the curve \emph{i.e.} solving the minimization problem
\begin{equation}\label{eq: potential minimizer}
    \inf \limits_{\gamma} \int \limits _0 ^1 V(\gamma(t)) dt \hspace{5mm}\text{s.t.}\hspace{5mm}\gamma(0)=z_1,~ \gamma(1)=z_2\,.
\end{equation}
In Figure \ref{fig: toy disks} are presented the decoded samples all along the interpolation curves. Thanks to those interpolations we can see that i) the latent space seems to really have a specific geometrical structure since decoding all along the interpolation curves obtained by solving Eq.~\eqref{eq: potential minimizer} leads to qualitatively satisfying results, ii) certain locations of the latent space must be avoided since sampling there will produce irrelevant samples (see red frames and corresponding red dashes). Using the proposed sampling scheme will allow to sample in the light-colored areas and so ensure that the sampling remains close to the data \emph{i.e.} where information is available and so does not produce irrelevant images when decoded while still proposing relevant variations from the input data.

\section{Experiments}

In this section, we conduct a comparison with other VAE models using other regularization schemes, more complex priors, richer posteriors, ex-post density estimation or trying to take into account geometrical aspects. In the following, all the models share the same auto-encoding neural network architectures and we used the code and hyper-parameters provided by the authors if available\footnote{We also perform a wider hyper-parameter search on MNIST and CELEBA for each model in Appendix~\ref{appC}}. See Appendix~\ref{appD} for models descriptions and the comprehensive experimental set-up.

\subsection{Generation with benchmark datasets}
First, we compare the proposed sampling method to several VAE variants such as a Wasserstein Autoencoder (WAE) \cite{tolstikhin2018wasserstein}, Regularized Autoencoders (RAEs) \cite{ghosh_variational_2020}, a vamp-prior VAE (VAMP) \cite{tomczak_vae_2018}, a Hamiltonian VAE (HVAE) \cite{caterini_hamiltonian_2018}, a geometry-aware VAE (RHVAE) \cite{chadebec_geometry-aware_2020} and an Autoencoder (AE). We elect these models since they use different ways to generate the data using either the prior or ex-post density estimation. For the latter, we fit a 10-component mixture of Gaussian in the latent space after training like \cite{ghosh_variational_2020} . 

\begin{figure}[t]
    \centering
    \captionsetup[subfigure]{position=above, labelformat = empty}
    \adjustbox{minipage=5em,raise=\dimexpr -3.\height}{\small AE - $\mathcal{N}$}
    \subfloat[MNIST]{\includegraphics[width=2.2in]{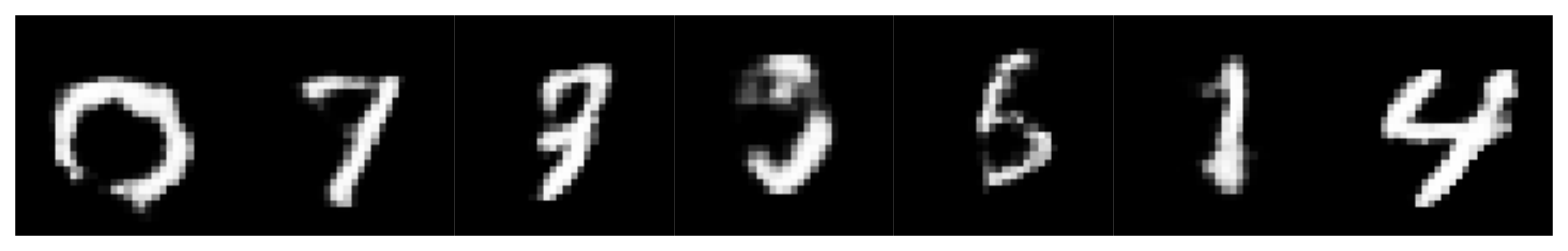}}
    \subfloat[CELEBA]{\includegraphics[width=2.2in]{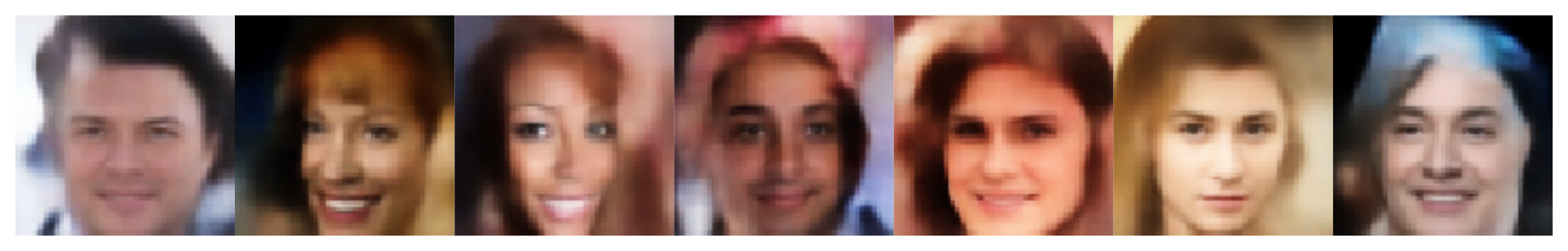}}\\
    \vspace{-5mm}
    \adjustbox{minipage=5em,raise=\dimexpr -3.\height}{\small VAE - $\mathcal{N}$}
    \subfloat{\includegraphics[width=2.2in]{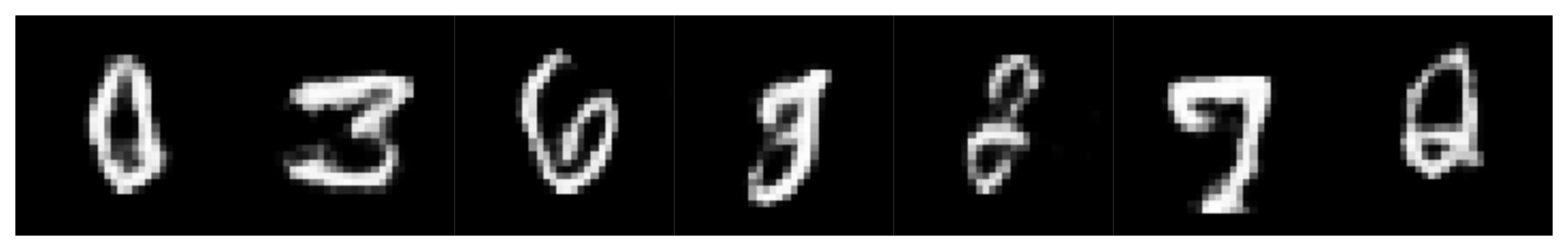}}
    \subfloat{\includegraphics[width=2.2in]{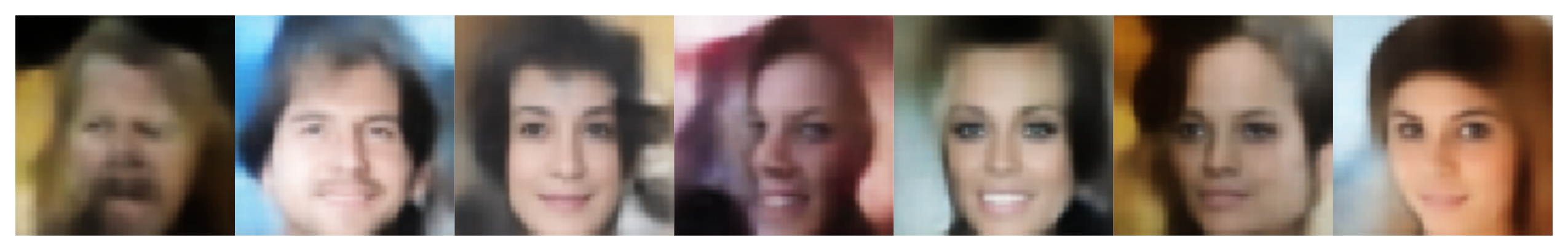}}\\
    \vspace{-5mm}
    \adjustbox{minipage=5em,raise=\dimexpr -3.\height}{\small WAE}
    \subfloat{\includegraphics[width=2.2in]{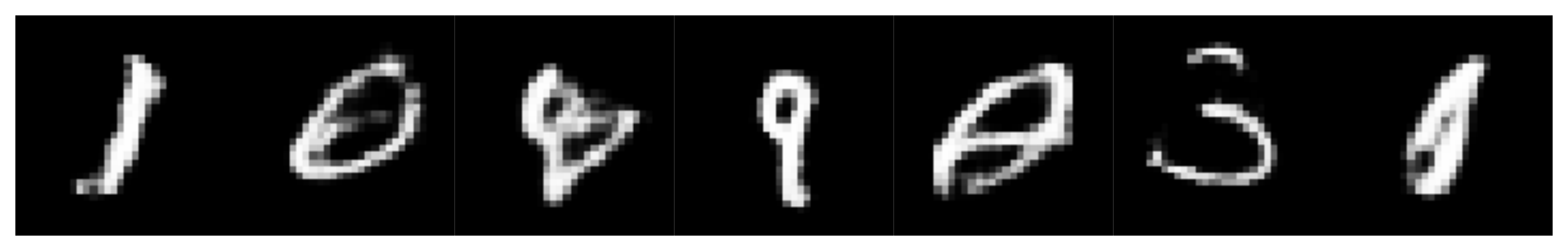}}
    \subfloat{\includegraphics[width=2.2in]{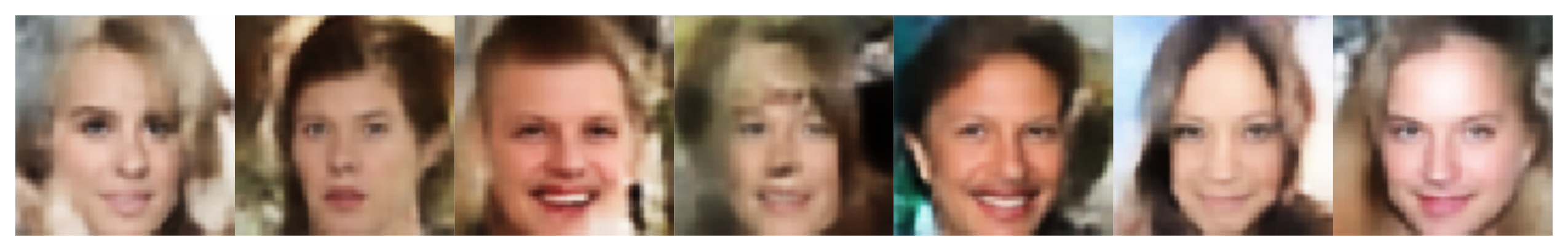}}\\
    \vspace{-5mm}
    \adjustbox{minipage=5em,raise=\dimexpr -3.\height}{\small VAMP}
    \subfloat{\includegraphics[width=2.2in]{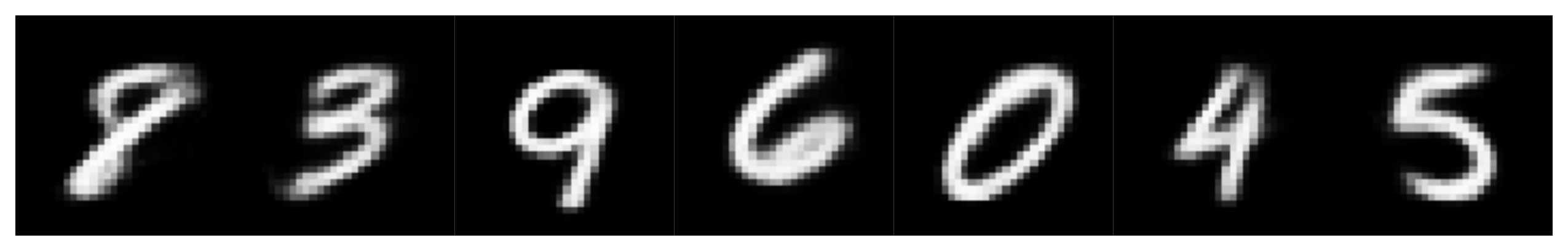}}
    \subfloat{\includegraphics[width=2.2in]{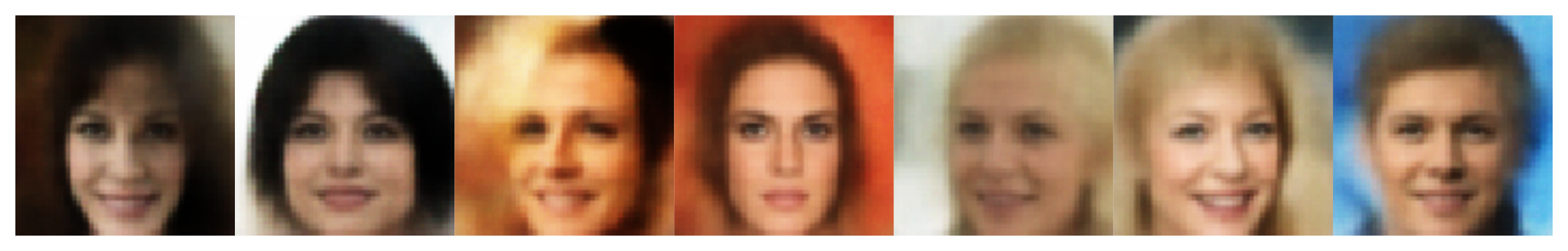}}\\
    \vspace{-5mm}
     \adjustbox{minipage=5em,raise=\dimexpr -3.\height}{\small HVAE}
    \subfloat{\includegraphics[width=2.2in]{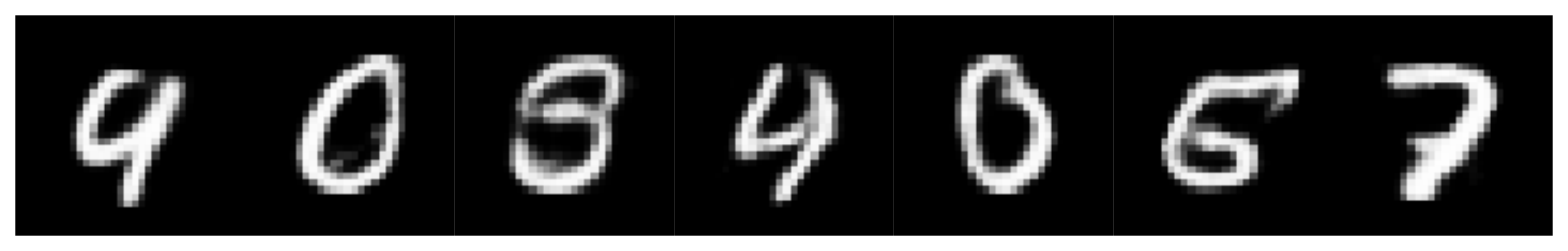}}
    \subfloat{\includegraphics[width=2.2in]{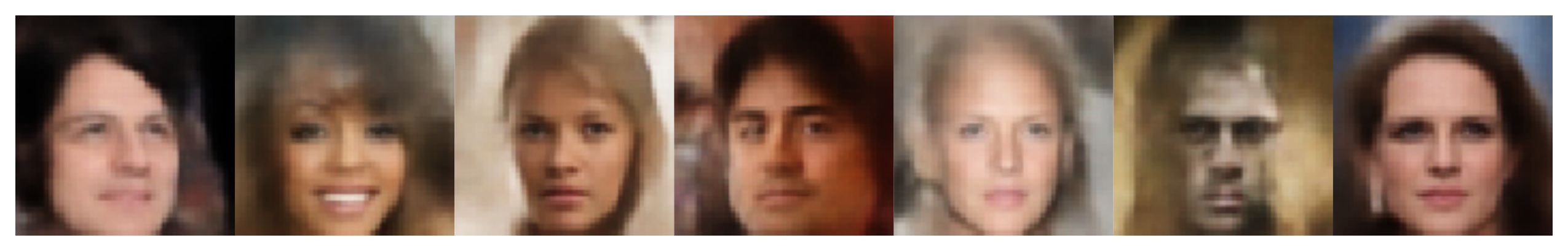}}\\
    \vspace{-5mm}
    \adjustbox{minipage=5em,raise=\dimexpr -3.\height}{\small RHVAE}
    \subfloat{\includegraphics[width=2.2in]{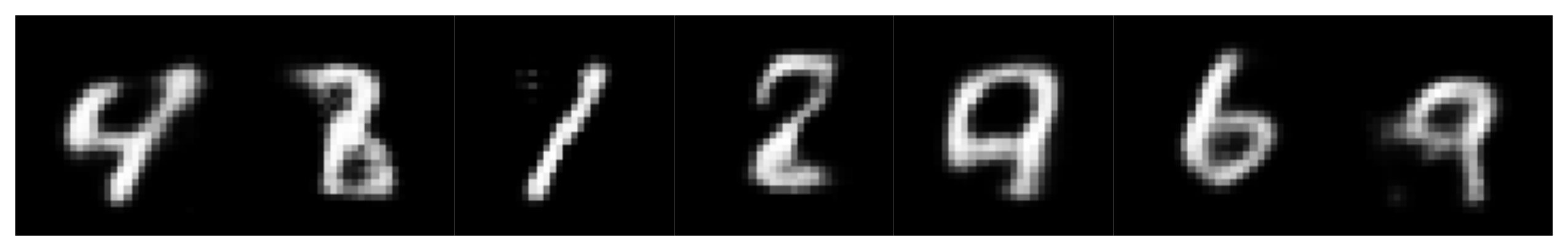}}
    \subfloat{\includegraphics[width=2.2in]{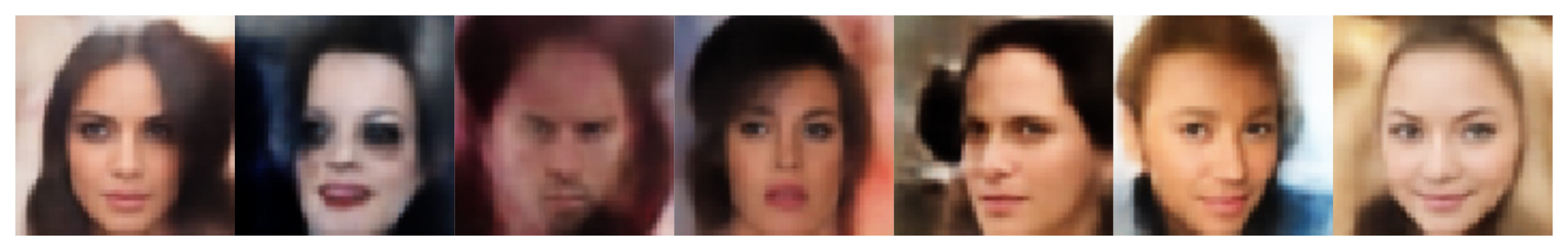}}\\
    \vspace{-5mm}
    \adjustbox{minipage=5em,raise=\dimexpr -3.\height}{\small AE - GMM}
    \subfloat{\includegraphics[width=2.2in]{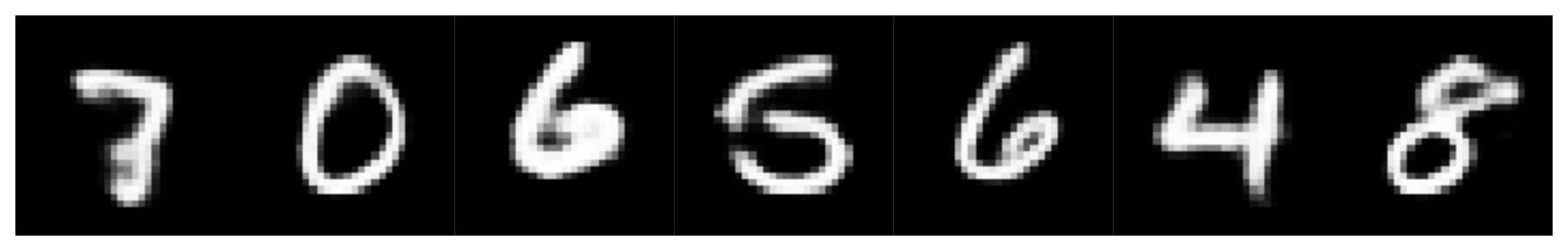}}
    \subfloat{\includegraphics[width=2.2in]{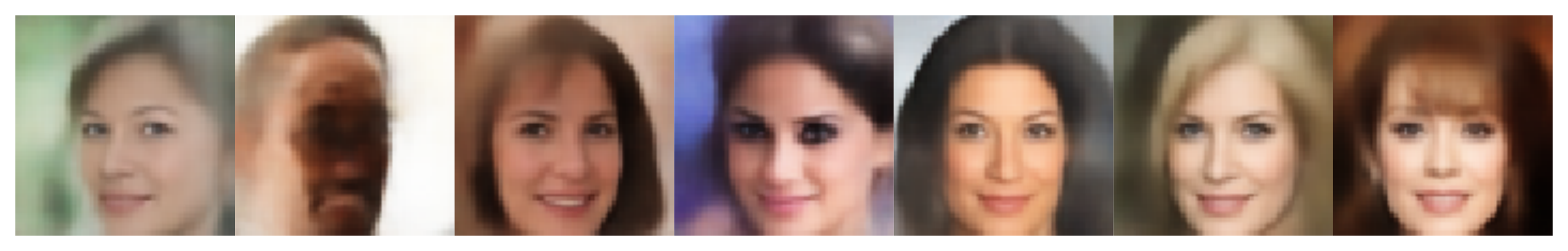}}\\
    \vspace{-5mm}
    \adjustbox{minipage=5em,raise=\dimexpr -3.\height}{\small VAE - GMM}
    \subfloat{\includegraphics[width=2.2in]{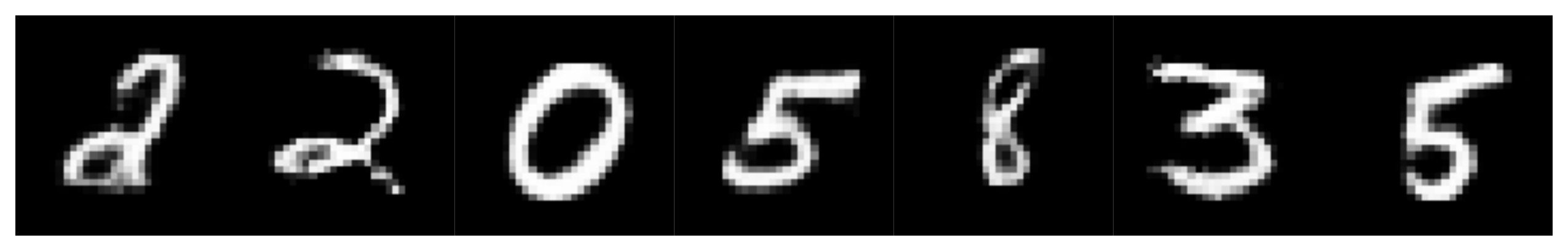}}
    \subfloat{\includegraphics[width=2.2in]{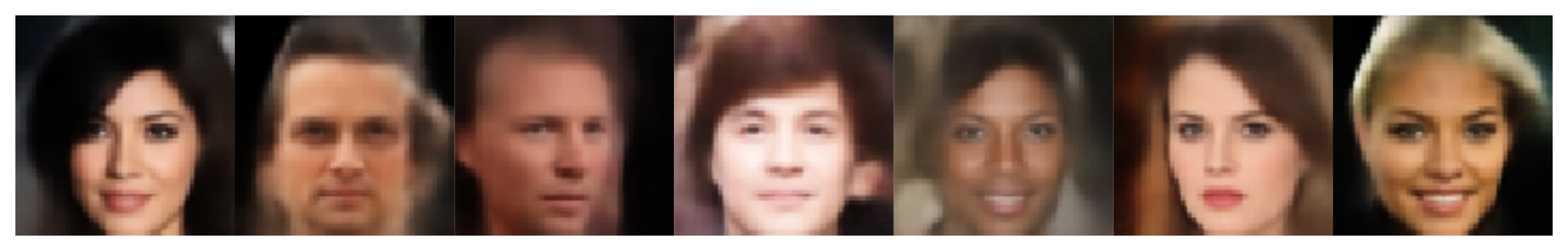}}\\
    \vspace{-5mm}
    \adjustbox{minipage=5em,raise=\dimexpr -3.\height}{\small RAE}
    \subfloat{\includegraphics[width=2.2in]{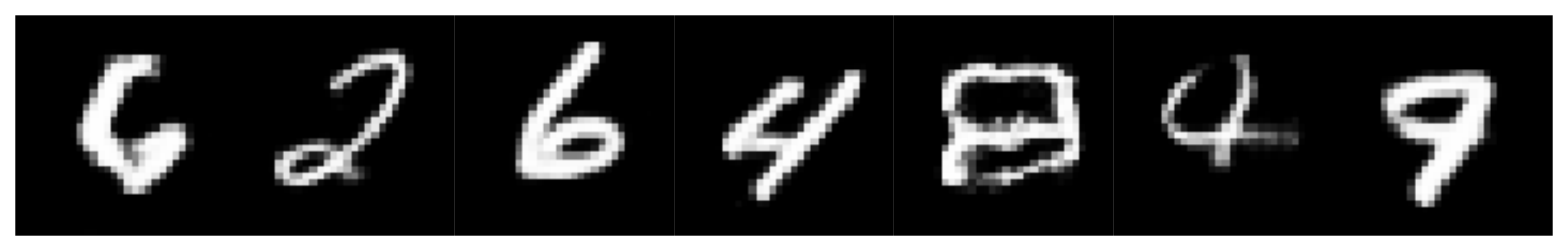}}
    \subfloat{\includegraphics[width=2.2in]{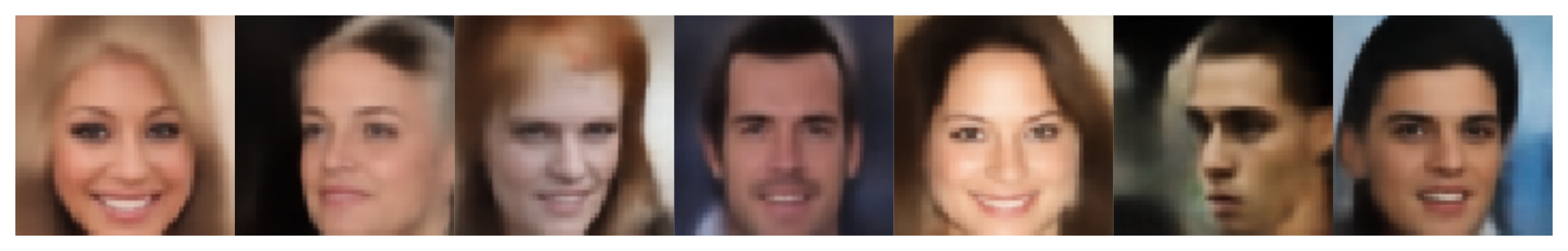}}\\
    \vspace{-5mm}
    \adjustbox{minipage=5em,raise=\dimexpr -3.\height}{\small VAE - Ours}
    \subfloat{\includegraphics[width=2.2in]{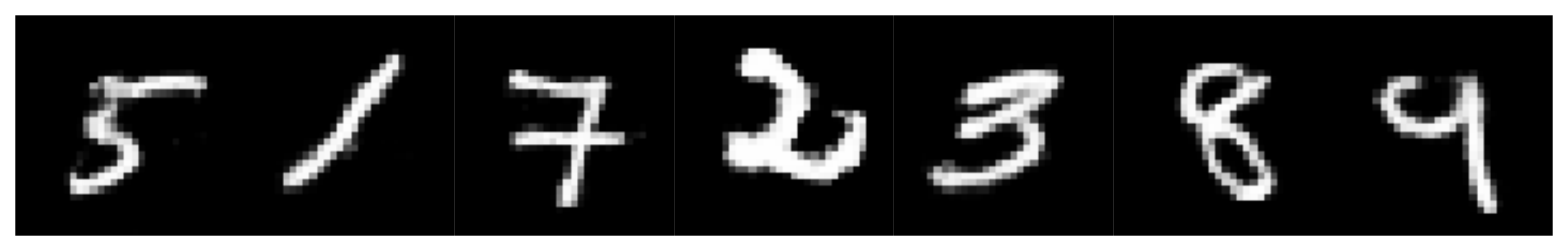}}
    \subfloat{\includegraphics[width=2.2in]{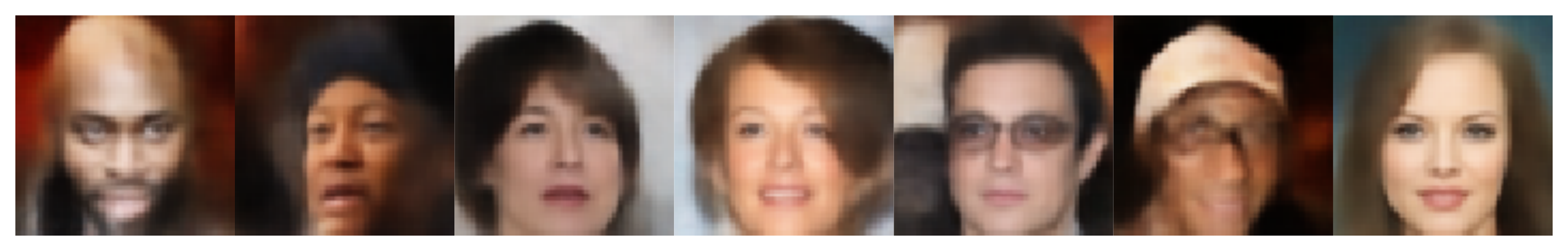}}

    \caption{Generated samples with different models and generation methods. Results with RAE variants are provided in Appendix~\ref{appC}.}
    \label{fig: generated samples}
    \end{figure}

Figure~\ref{fig: generated samples} shows a qualitative comparison between the resulting generated samples for MNIST \cite{lecun_mnist_1998} and CELEBA \cite{liu2015faceattributes}, see Appendix~\ref{appC} for SVHN \cite{netzer2011reading} and CIFAR 10 \cite{krizhevsky2009learning}. Interestingly, using the non-prior based methods seems to produce qualitatively better samples (rows 7 to end). Nonetheless, the resulting samples seem even sharper when the sampling takes into account geometrical aspects of the latent space as we propose (last row). Additionally, even though the exact same model is used, we clearly see that using the proposed method represents a strong improvement of the generation process from a vanilla VAE when compared to the samples coming from a normal prior (second row). This confirms that even the simplest VAE model actually contains a lot of information in its latent space but the limited expressiveness of the prior impedes to access to it. Hence, using more complex priors such as the VAMP may be a tempting idea. However, one must keep in mind that the ELBO objective in Eq.~\eqref{eq: ELBO} must remain tractable and so using more expressive priors may be impossible.

These observations are even more supported by Table~\ref{tab: fid-prd} where we report the Frechet Inception Distance (FID) and the precision and recall (PRD) score against the test set to assess the sampling quality and diversity. Again, fitting a mixture of Gaussian (GMM) in the latent space appears to be an interesting idea since it allows for a better expressiveness and latent space prospecting. For instance, on MNIST the FID falls from 40.7 with the prior to 13.1 when using a GMM. Nonetheless, with the proposed method we are able to make it even smaller (8.5) and PRD scores higher without changing the model and performing post processing. This can also be observed on the 3 other datasets. Impressively, in almost all cases, the proposed generation method can either compete or outperform peers both in terms of FID and PRD scores.

\begin{figure}[t]
\begin{minipage}{0.68\linewidth}
    \centering
    \captionsetup[subfigure]{position=above, labelformat = empty}
    \subfloat[\centering \scriptsize Gen.\hspace{5mm}]{\includegraphics[width=0.3in]{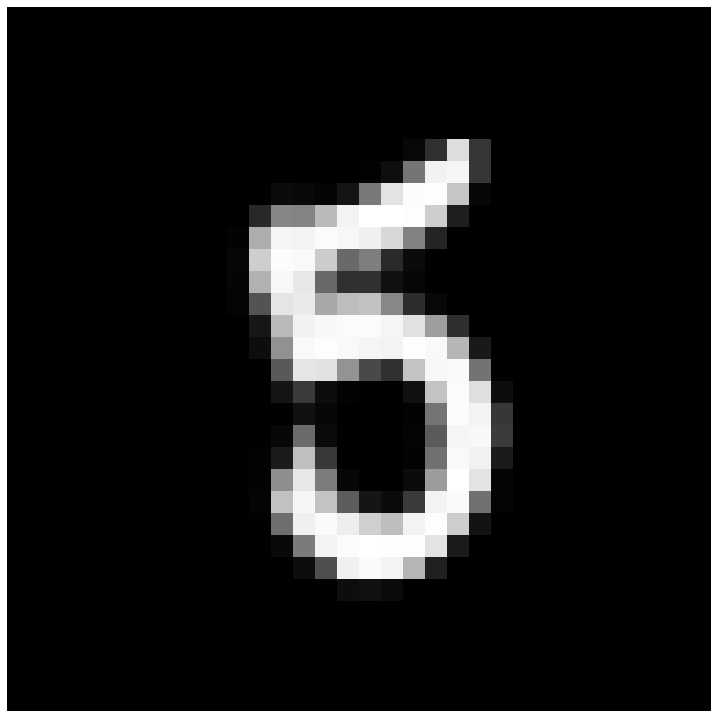}}
    \subfloat[\centering \scriptsize Near. train]{\includegraphics[width=0.3in]{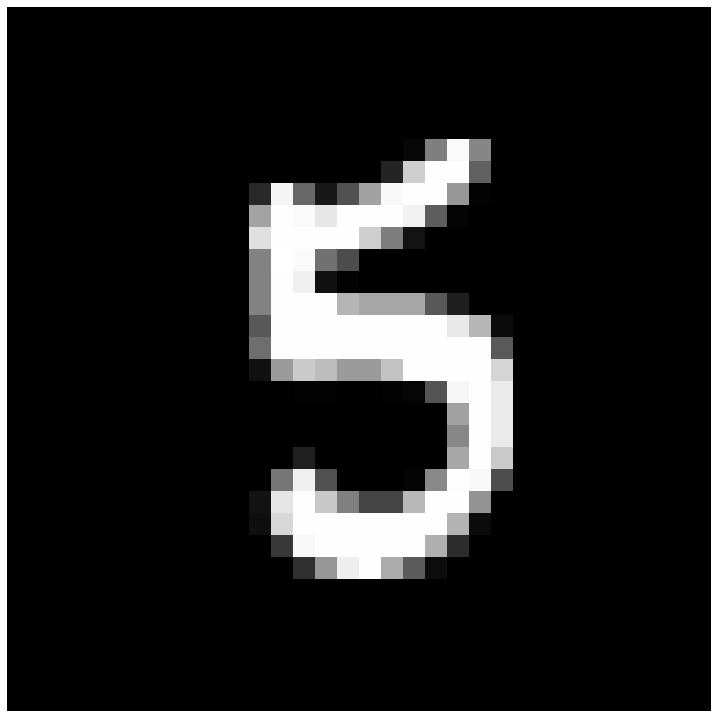}}
    \subfloat[\centering \scriptsize Near. rec.]{\includegraphics[width=0.3in]{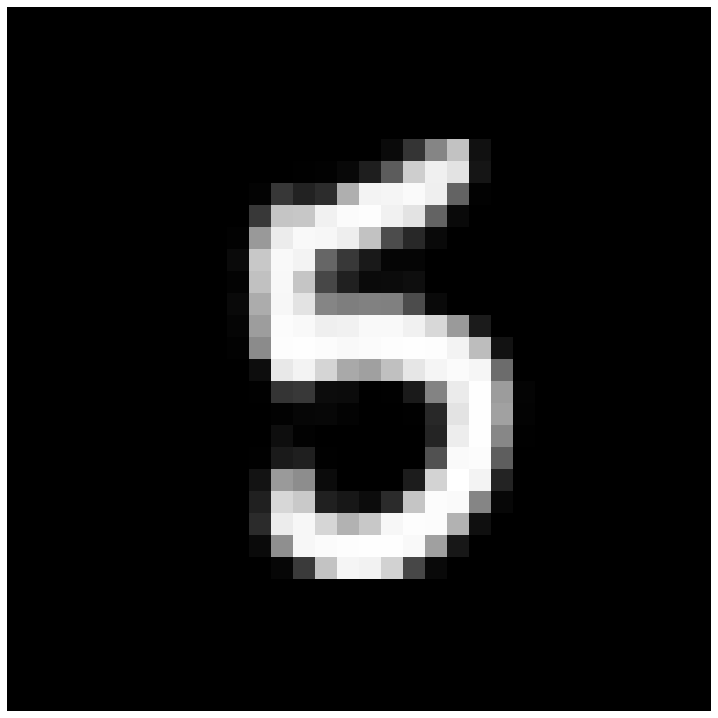}}
    \hspace{0.0001em}
     \subfloat[\centering \scriptsize Gen.\hspace{5mm}]{\includegraphics[width=0.3in]{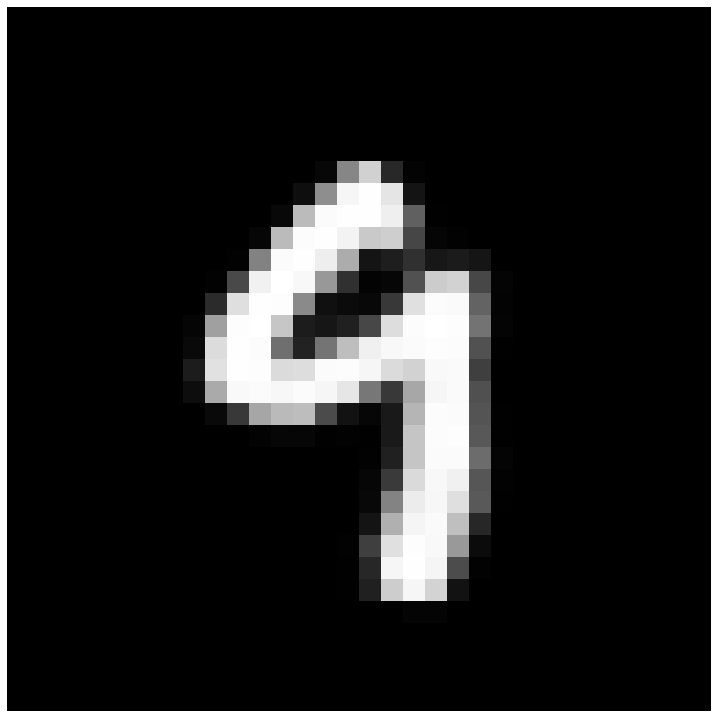}}
    \subfloat[\centering \scriptsize Near. train]{\includegraphics[width=0.3in]{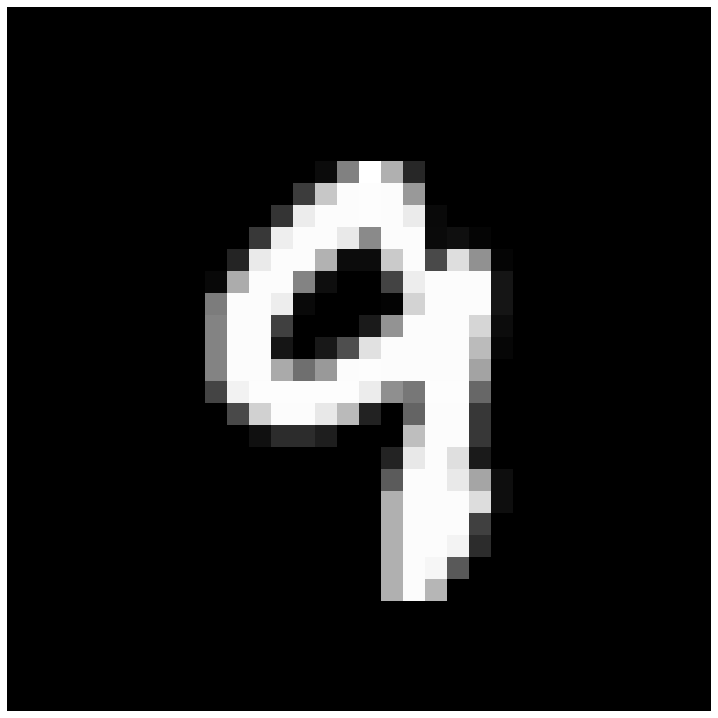}}
    \subfloat[\centering \scriptsize Near. rec.]{\includegraphics[width=0.3in]{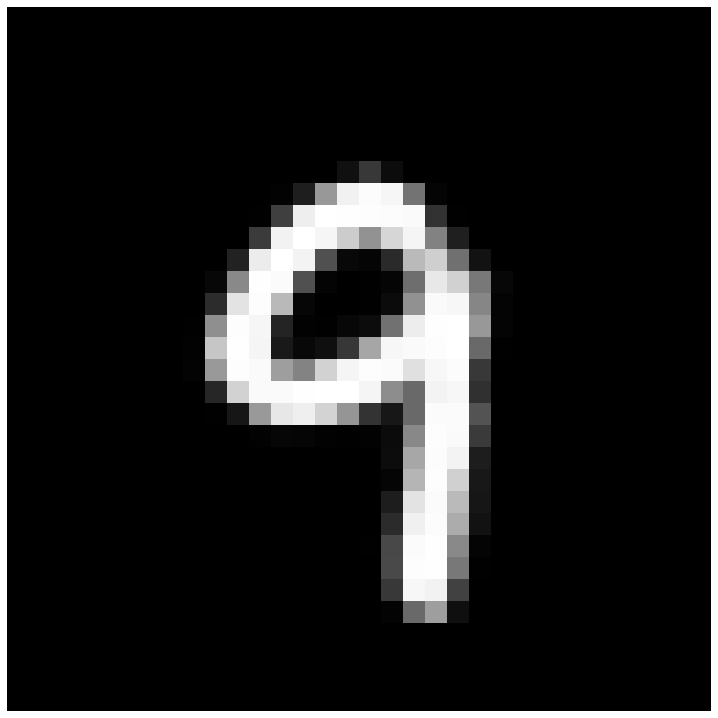}}
    \hspace{0.001em}
     \subfloat[\centering \scriptsize Gen.\hspace{5mm}]{\includegraphics[width=0.3in]{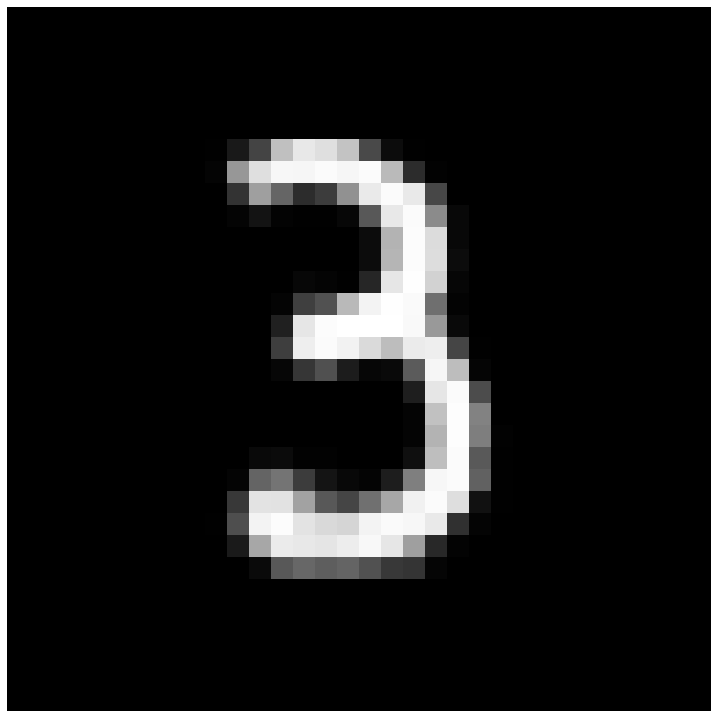}}
    \subfloat[\centering \scriptsize Near. train]{\includegraphics[width=0.3in]{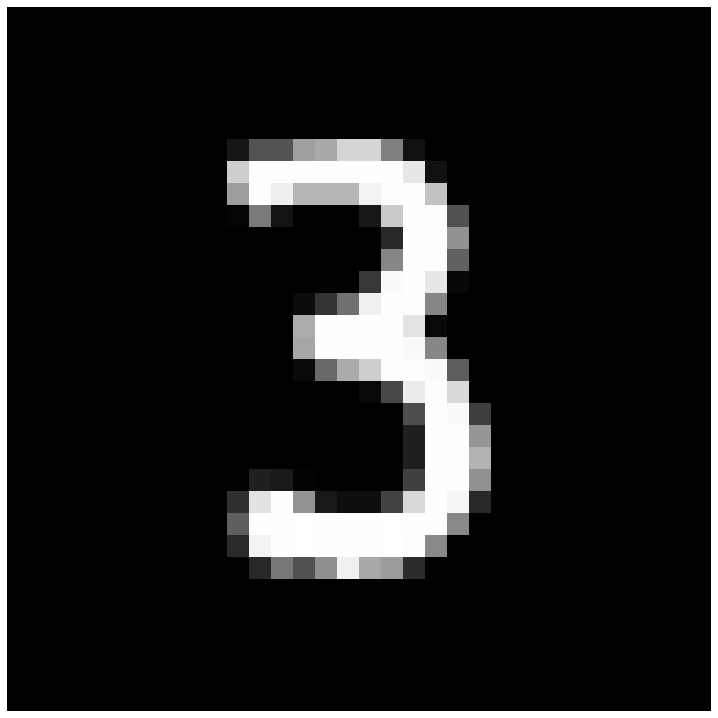}}
    \subfloat[\centering \scriptsize Near. rec.]{\includegraphics[width=0.3in]{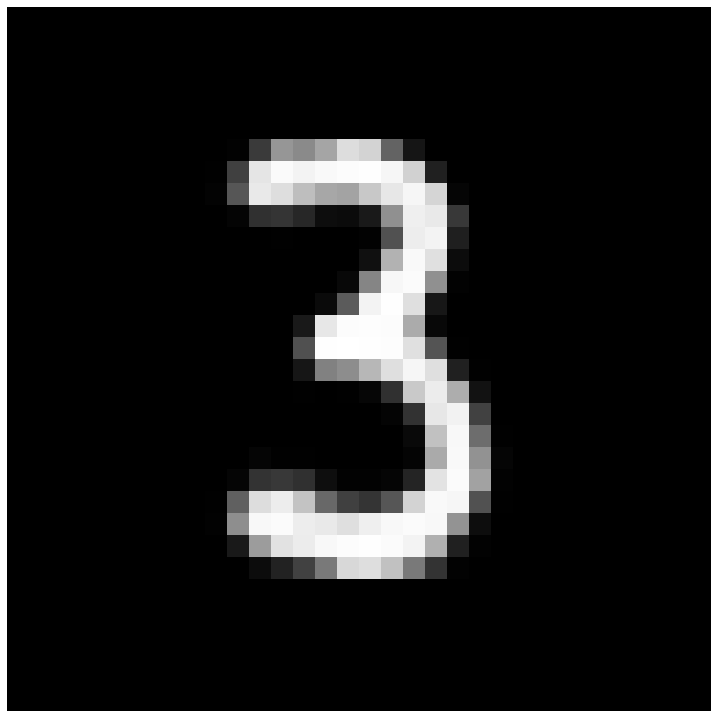}}
    \hspace{0.001em}
     \subfloat[\centering \scriptsize Gen.\hspace{5mm}]{\includegraphics[width=0.3in]{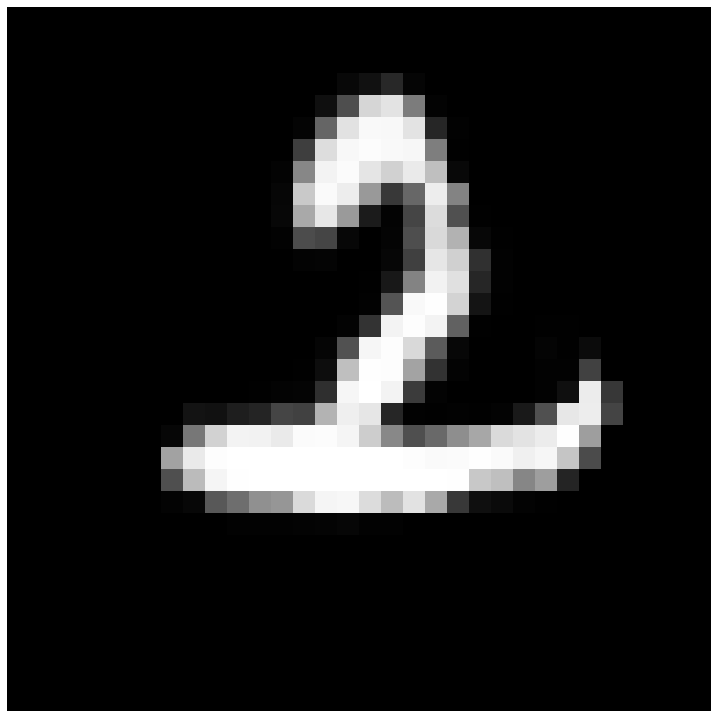}}
    \subfloat[\centering \scriptsize Near. train]{\includegraphics[width=0.3in]{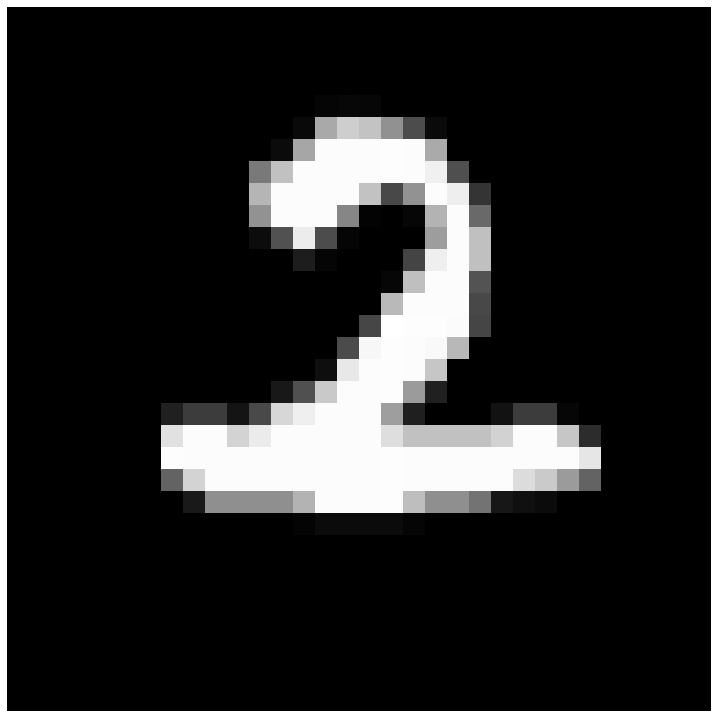}}
    \subfloat[\centering \scriptsize Near. rec.]{\includegraphics[width=0.3in]{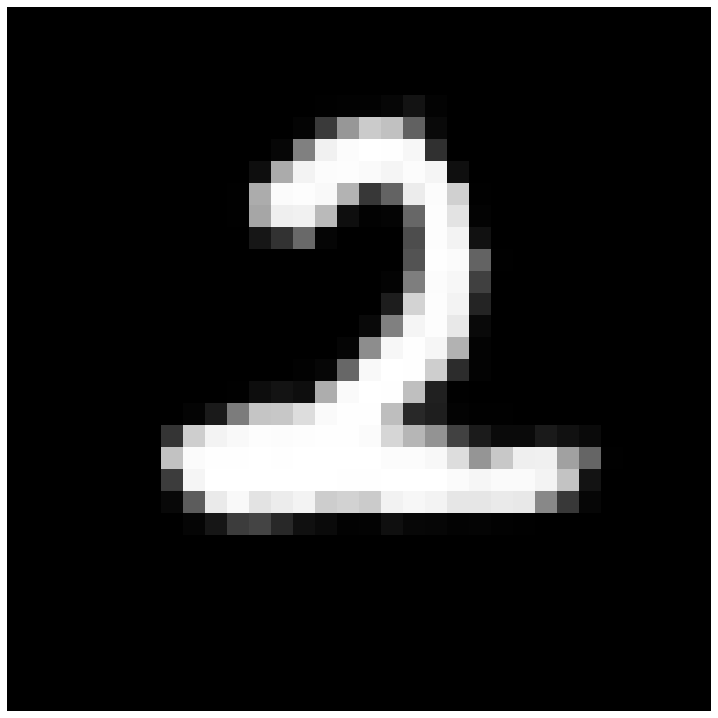}}
    \vfil
    \vspace{-4mm}
    \subfloat{\includegraphics[width=0.3in]{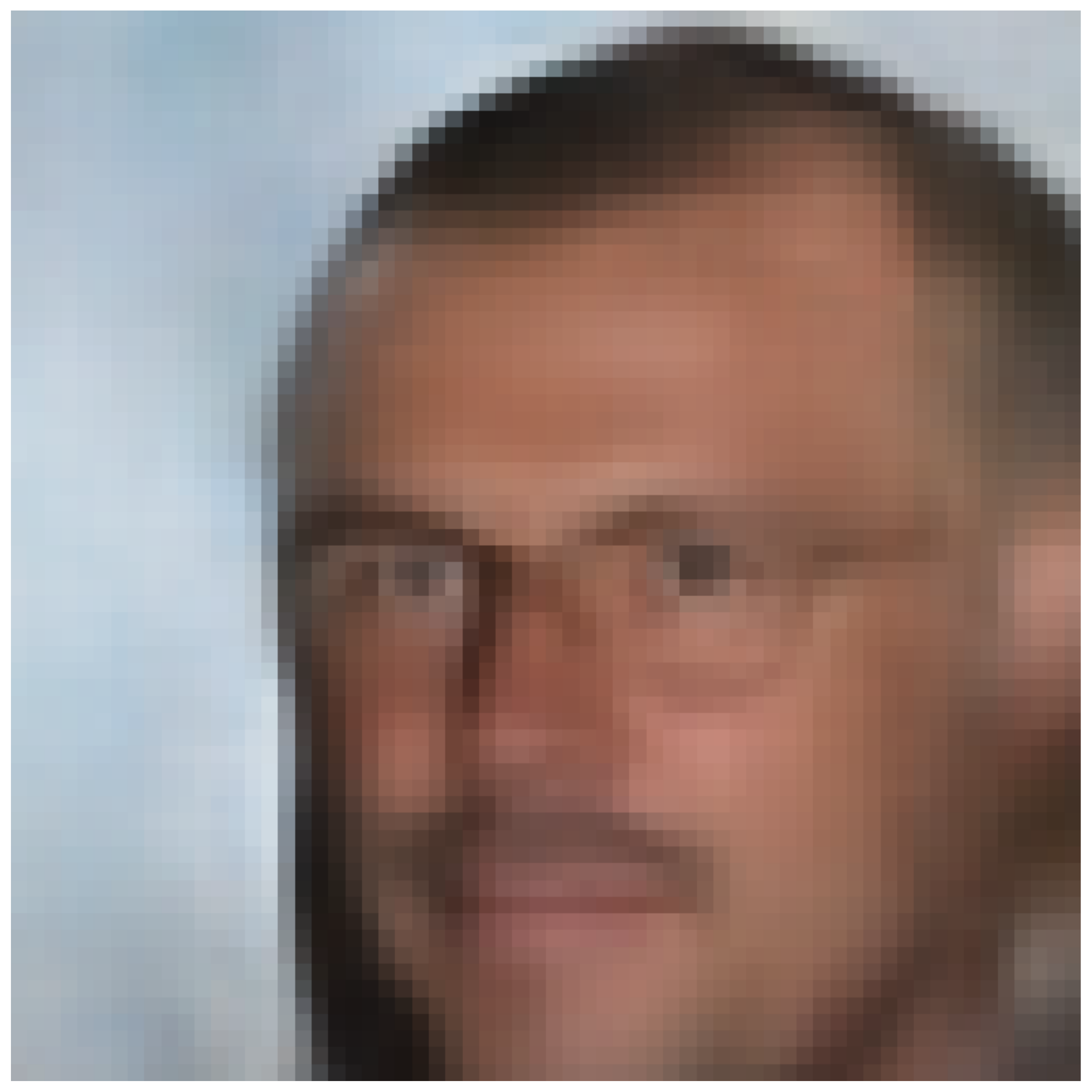}}
    \subfloat{\includegraphics[width=0.3in]{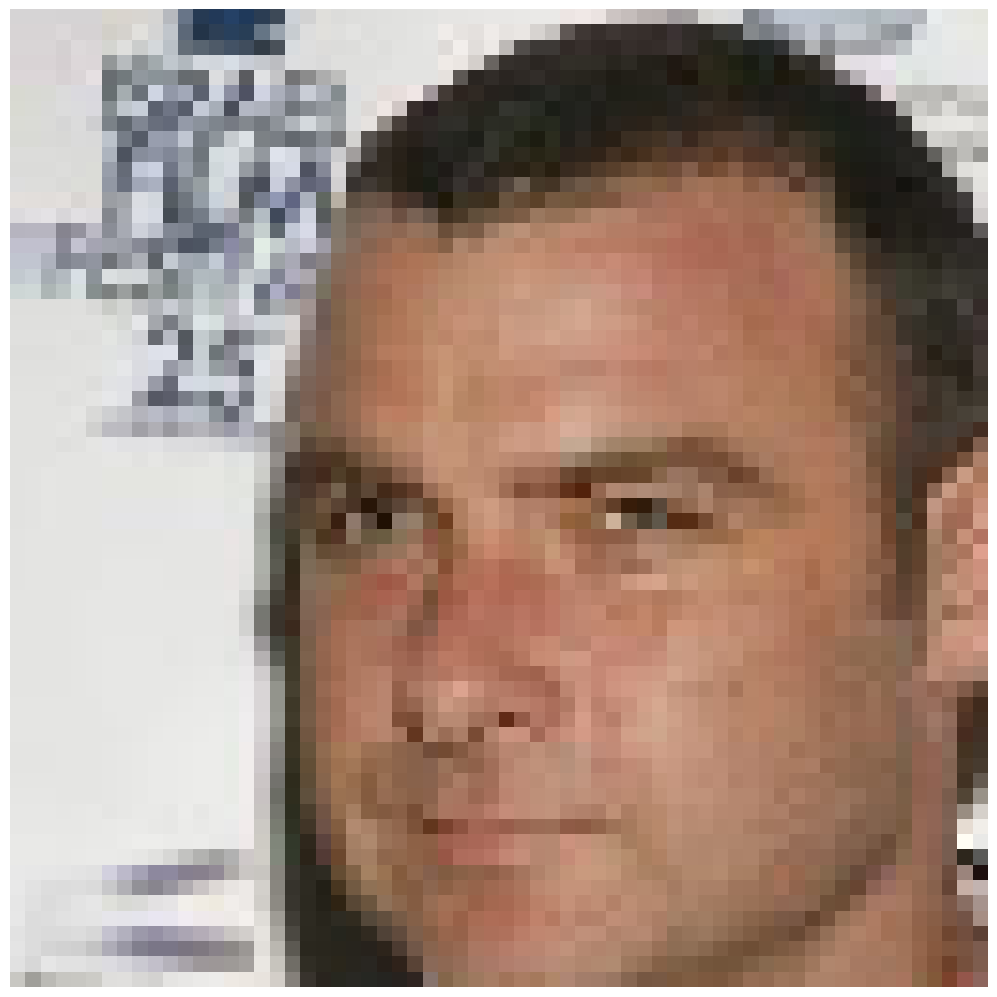}}
    \subfloat{\includegraphics[width=0.3in]{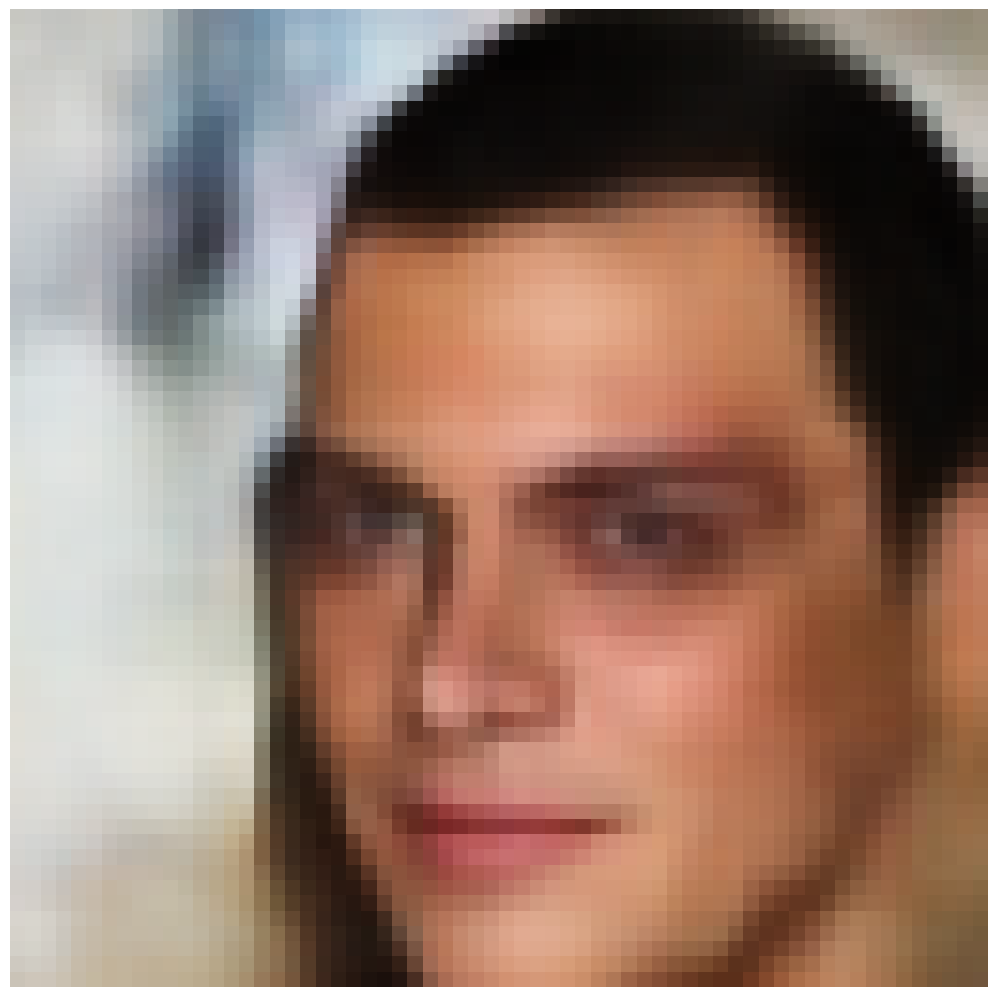}}
    \hspace{0.001em}
    \subfloat{\includegraphics[width=0.3in]{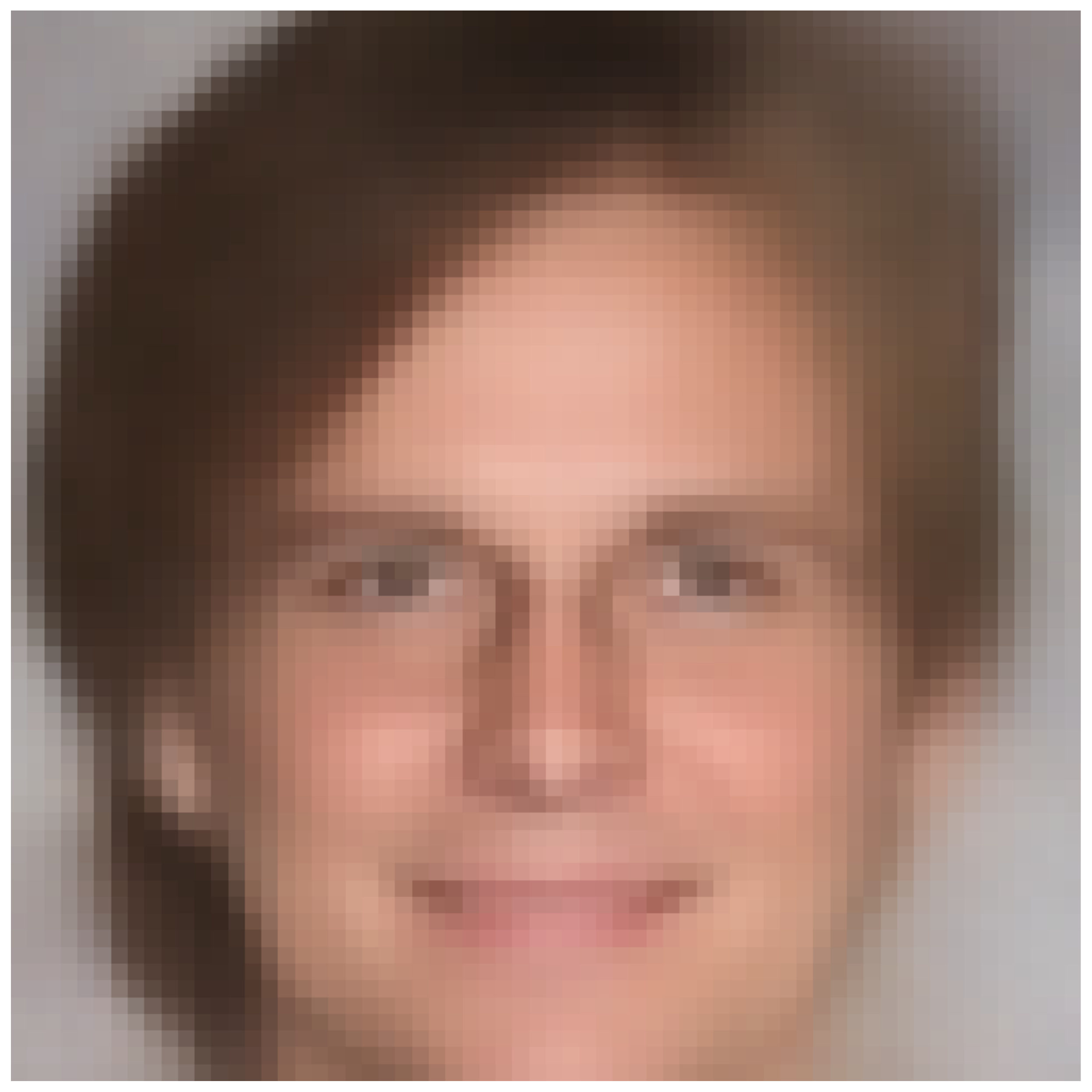}}
    \subfloat{\includegraphics[width=0.3in]{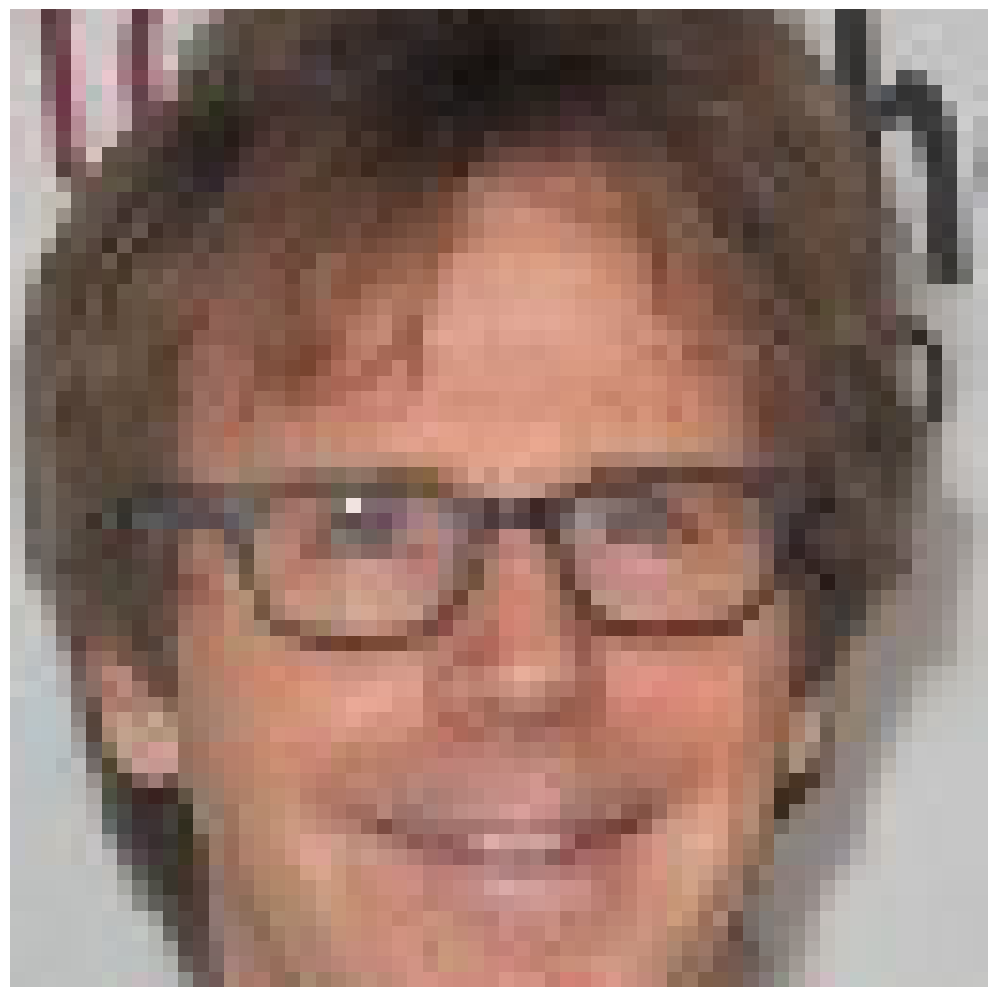}}
    \subfloat{\includegraphics[width=0.3in]{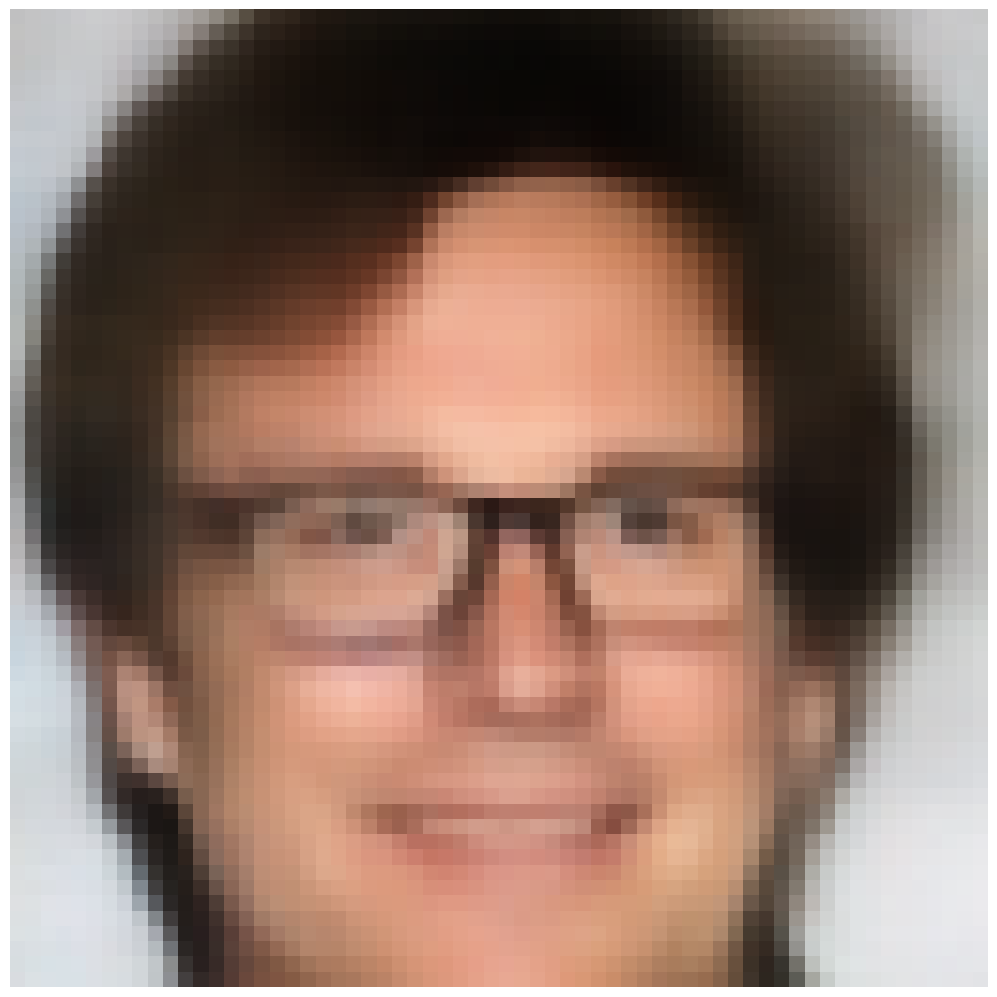}}
   \hspace{0.001em}
    \subfloat{\includegraphics[width=0.3in]{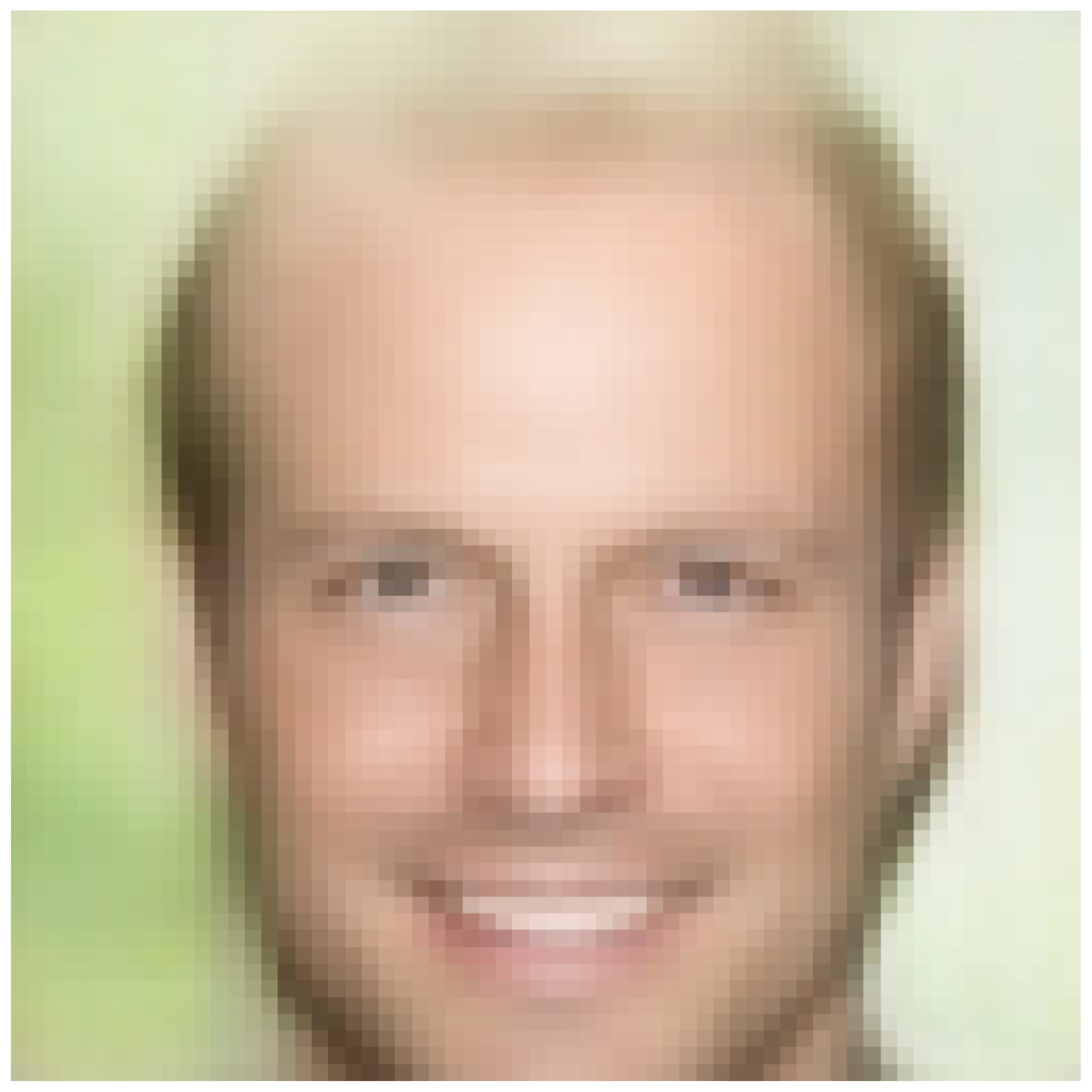}}
    \subfloat{\includegraphics[width=0.3in]{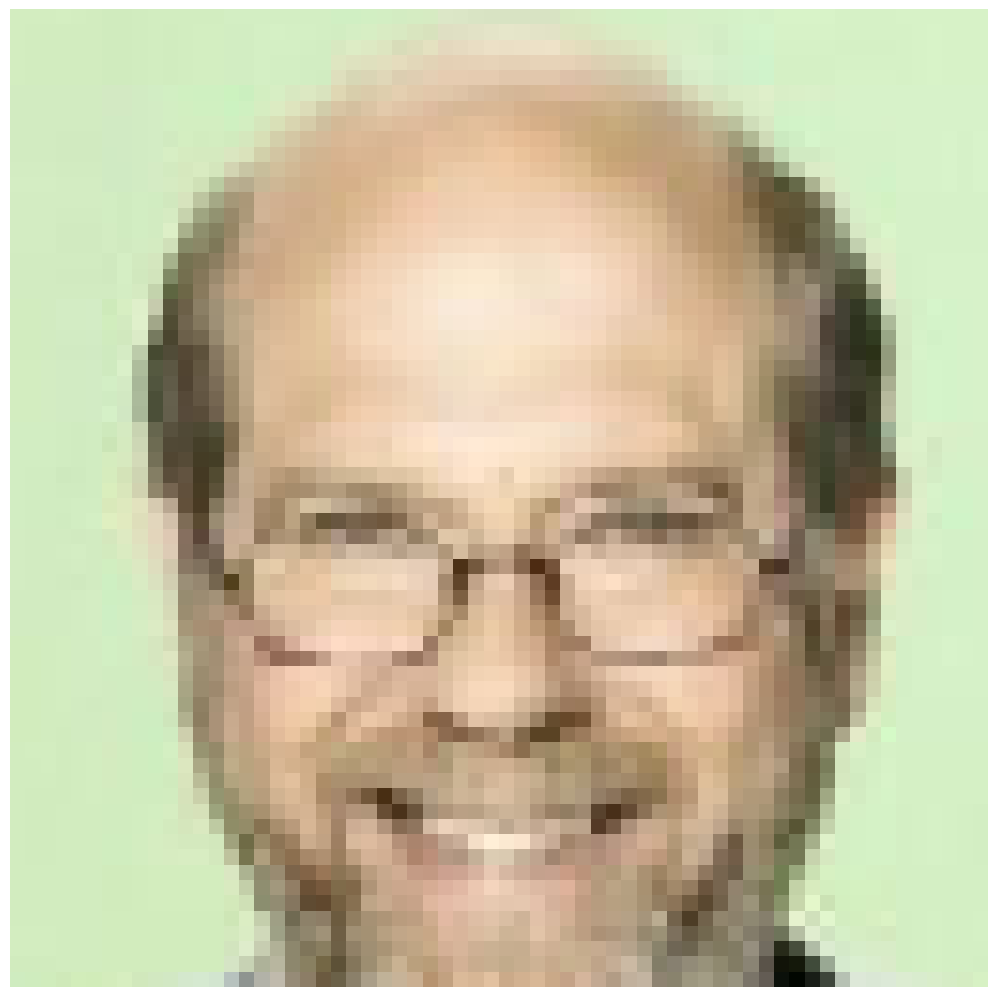}}
    \subfloat{\includegraphics[width=0.3in]{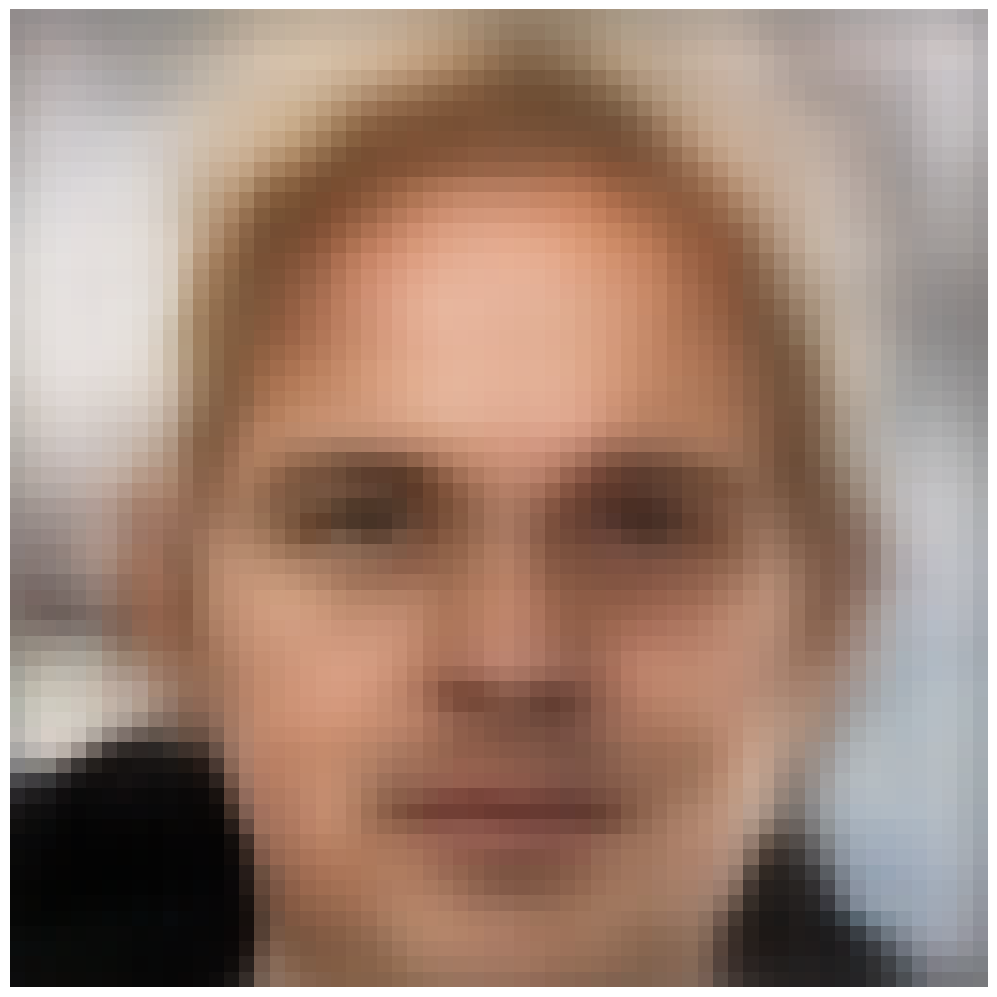}}
    \hspace{0.001em}
    \subfloat{\includegraphics[width=0.3in]{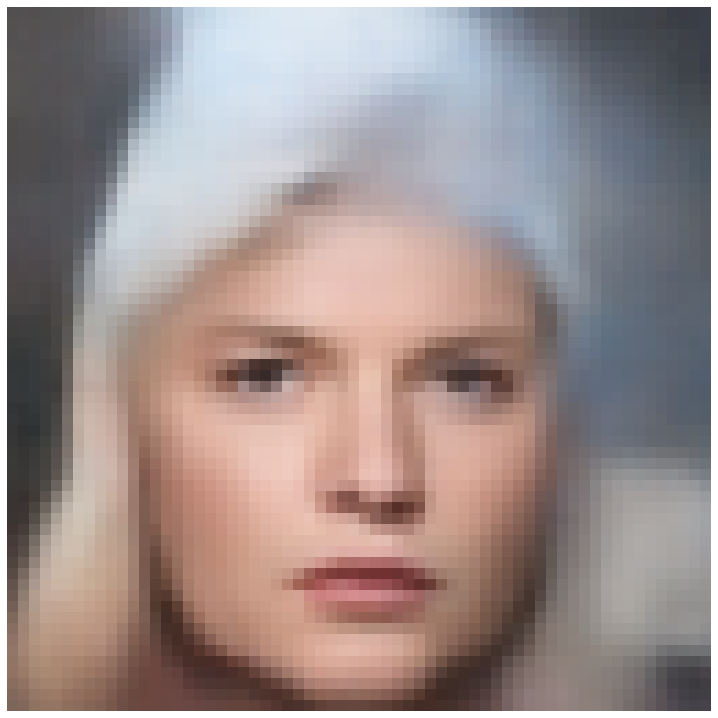}}
    \subfloat{\includegraphics[width=0.3in]{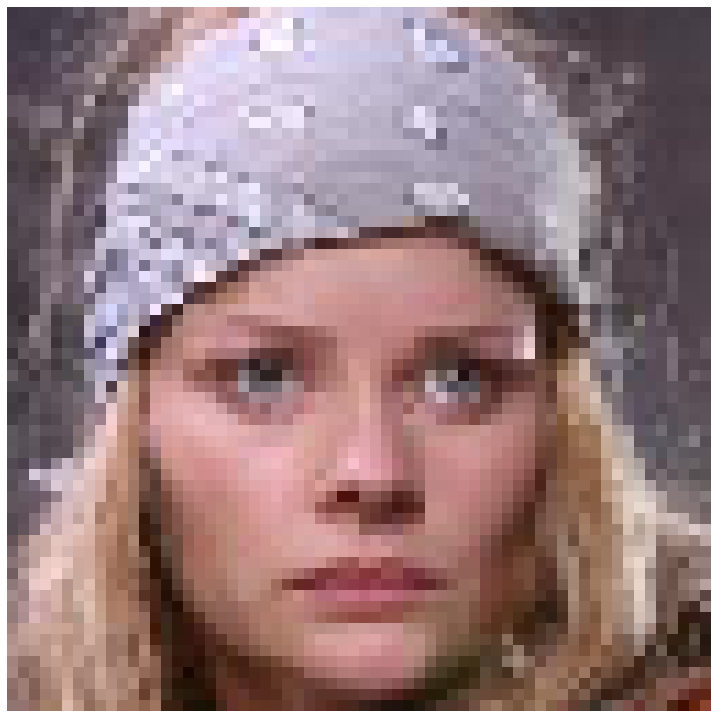}}
    \subfloat{\includegraphics[width=0.3in]{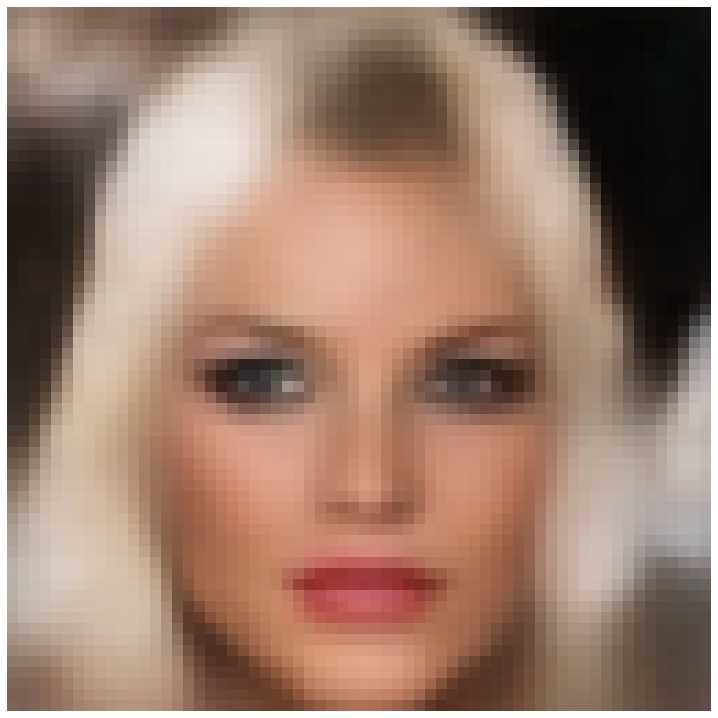}}
\end{minipage}
\hfil
\begin{minipage}{0.3\linewidth}
\vspace{3.1em}
\small
        \centering
        \captionsetup{type=table}
        \begin{tabular}{l|cc}
        \toprule
        \multicolumn{3}{c}{reconstruction vs. generation}\\
        \midrule
            & MNIST & CELEBA \\
        FID & 11.27 & 30.12  \\
        \bottomrule
        \end{tabular}
\end{minipage}
    \caption{\emph{Left:} Nearest train image (near. train) and nearest image in all reconstructions of train images (near. rec.) to the generated one (Gen.) with the proposed method. Note: the nearest reconstruction may be different from the reconstruction of the nearest train image. \emph{Right}: The FID score between 10k generated images and 10k reconstructed train samples.}
    \label{fig: overfitting}
    \end{figure}

Finally, we check if the proposed method does not overfit the training data and is able to produce diverse samples by showing the nearest neighbor in the train set and the nearest image in all the reconstructions of the train images to a generated image in Figure~\ref{fig: overfitting} (left). We also provide the FID score between 10k generated samples and 10k train reconstructions in Figure~\ref{fig: overfitting} (right). These experiments show that the generated samples are not only resampled train images and that the sampling prospects quite well the manifold. To support even more this claim we provide in Appendix~\ref{appF} an analysis in a case where only two centroids are selected in the metric. This also shows that the generated samples are not only an interpolation between the $k$ selected centroids since some generated images contain attributes that are not present in the images of the decoded centroids. 

The outcome of such an experiment is that using post  training latent space processing such as ex-post density estimation or adding some geometrical consideration to the model allows to strongly improve the sampling without adding more complexity to the model. Generating 1k samples on CELEBA takes approx.~5.5 min for our method vs. 4 min for a 10-component GMM on a GPU V100-16GB.

\begin{table}[ht]
    \caption{FID (lower is better) and PRD score (higher is better) for different models and datasets. For the mixture of Gaussian (GMM), we fit a 10-component mixture of Gaussian in the latent space.}
    \label{tab: fid-prd}
    \begin{center}
    \scriptsize
    \begin{sc}
    
    \begin{tabular}{l|cc|cc|cc|cc}
    \toprule
    \multirow{2}{*}{Model}  &\multicolumn{2}{c|}{MNIST (16)} & \multicolumn{2}{c|}{SVHN (16)} & \multicolumn{2}{c|}{CIFAR 10 (32)} & \multicolumn{2}{c}{CELEBA (64)}\\
                                  & FID $\downarrow$   & PRD $\uparrow$ & FID $\downarrow$   & PRD $\uparrow$ & FID $\downarrow$    & PRD $\uparrow$ & FID $\downarrow$   & PRD $\uparrow$ \\
        \midrule
        AE  - $\mathcal{N}(0, 1)$ & 46.41  & 0.86/0.77 & 119.65 & 0.54/0.37 & 196.50    & 0.05/0.17 & 64.64 & 0.29/0.42 \\
        WAE & 20.71  & 0.93/0.88 & 49.07  & 0.80/\textbf{0.85} & 132.99  & 0.24/0.52 & 54.56 & \textbf{0.57}/0.55 \\
        VAE - $\mathcal{N}(0, 1)$ & 40.70 & 0.83/0.75 & 83.55  & 0.69/0.55 & 162.58 & 0.10/0.32 & 64.13 & 0.27/0.39\\
        VAMP                      & 34.02 & 0.83/0.88 & 91.98  & 0.55/0.63 & 198.14 & 0.05/0.11 & 73.87 & 0.09/0.10 \\
        HVAE  & 15.54 & 0.97/0.95 & 98.05 & 0.64/0.68 & 201.70 & 0.13/0.21 & 52.00 & 0.38/0.58 \\
        RHVAE & 36.51 & 0.73/0.28 & 121.69 & 0.55/0.41    & 167.41 & 0.12/0.22 & 55.12 & 0.45/0.56 \\
        \midrule
        AE - GMM                  & 9.60   & 0.95/0.90 & 54.21  & 0.82/0.83 & 130.28 & 0.35/0.58    & 56.07 & 0.32/0.48 \\
        RAE (GP)                  & 9.44   & 0.97/\textbf{0.98} & 61.43  & 0.79/0.78 & 120.32  & 0.34/0.58 & 59.41 & 0.28/0.49 \\
        RAE (L2)                  & 9.89   & 0.97/\textbf{0.98} & 58.32  & 0.82/0.79 & 123.25  & 0.33/0.54 & 54.45 & 0.35/0.55 \\
        RAE (SN)                  & 11.22  & 0.97/\textbf{0.98} & 95.64  & 0.53/0.63 & 114.59  & 0.32/0.53 & 55.04 & 0.36/0.56 \\
        RAE                       & 11.23  & \textbf{0.98}/\textbf{0.98} & 66.20  & 0.76/0.80 & 118.25  & 0.35/0.57 & 53.29 & 0.36/0.58 \\
        VAE - GMM                 & 13.13  & 0.95/0.92 & 52.32  & 0.82/\textbf{0.85} & 138.25  & 0.29/0.53 & 55.50 & 0.37/0.49\\
        \midrule
        VAE - Ours                & \textbf{8.53}   & \textbf{0.98}/0.97 & \textbf{46.99}  & \textbf{0.84}/\textbf{0.85} & \textbf{93.53}   & \textbf{0.71}/\textbf{0.68} & \textbf{48.71} & 0.44/\textbf{0.62}\\
        \bottomrule
    \end{tabular}
    \end{sc}
    \end{center}
        \vskip -0.1in
\end{table}

\begin{figure}[ht]
    \centering
    \captionsetup[subfigure]{position=above, labelformat = empty}
    \vskip -0.1in
    \subfloat[MNIST]{\includegraphics[width=2.2in]{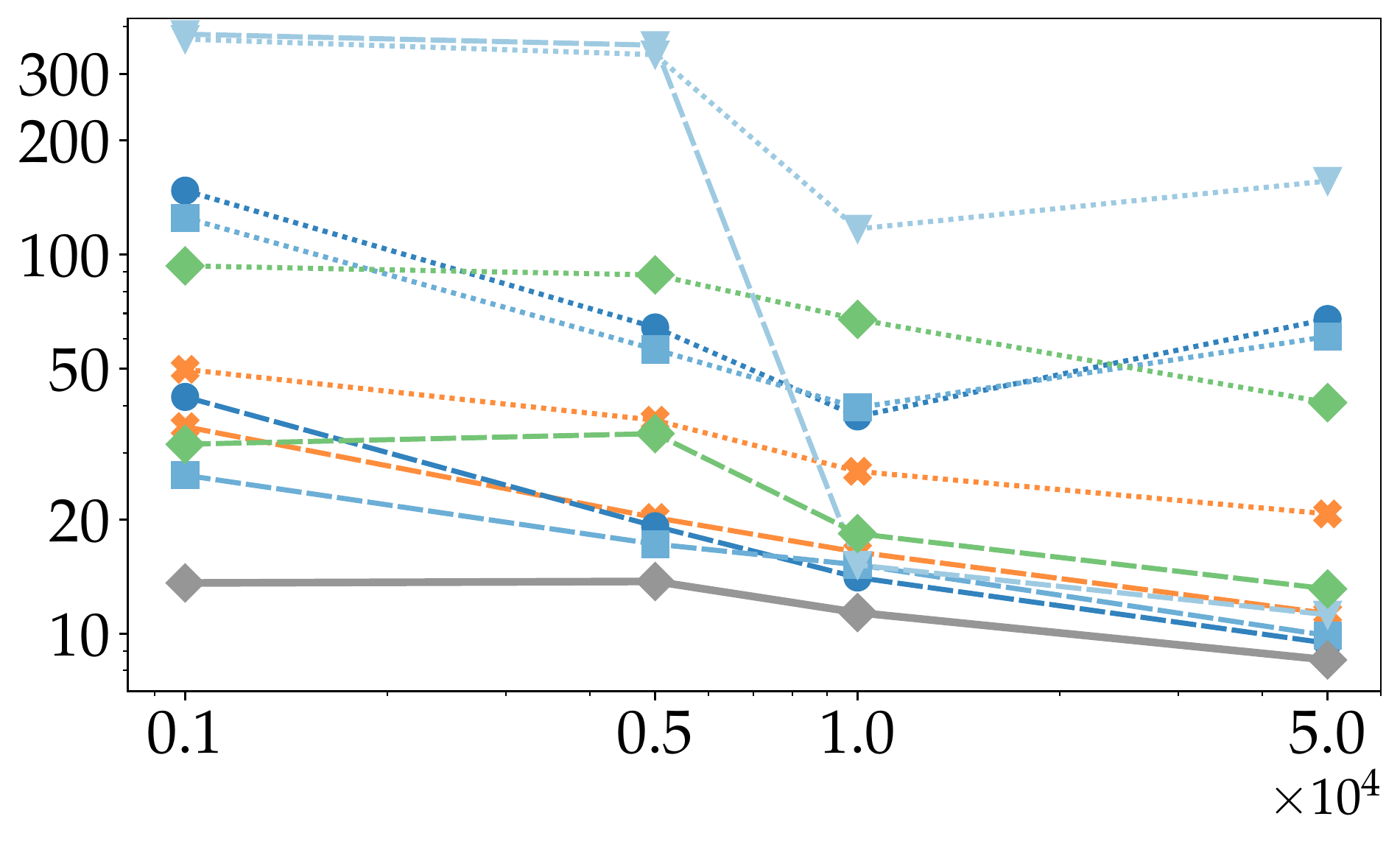}}
    \subfloat[\hspace{-17mm}CIFAR 10]{\includegraphics[width=2.8in]{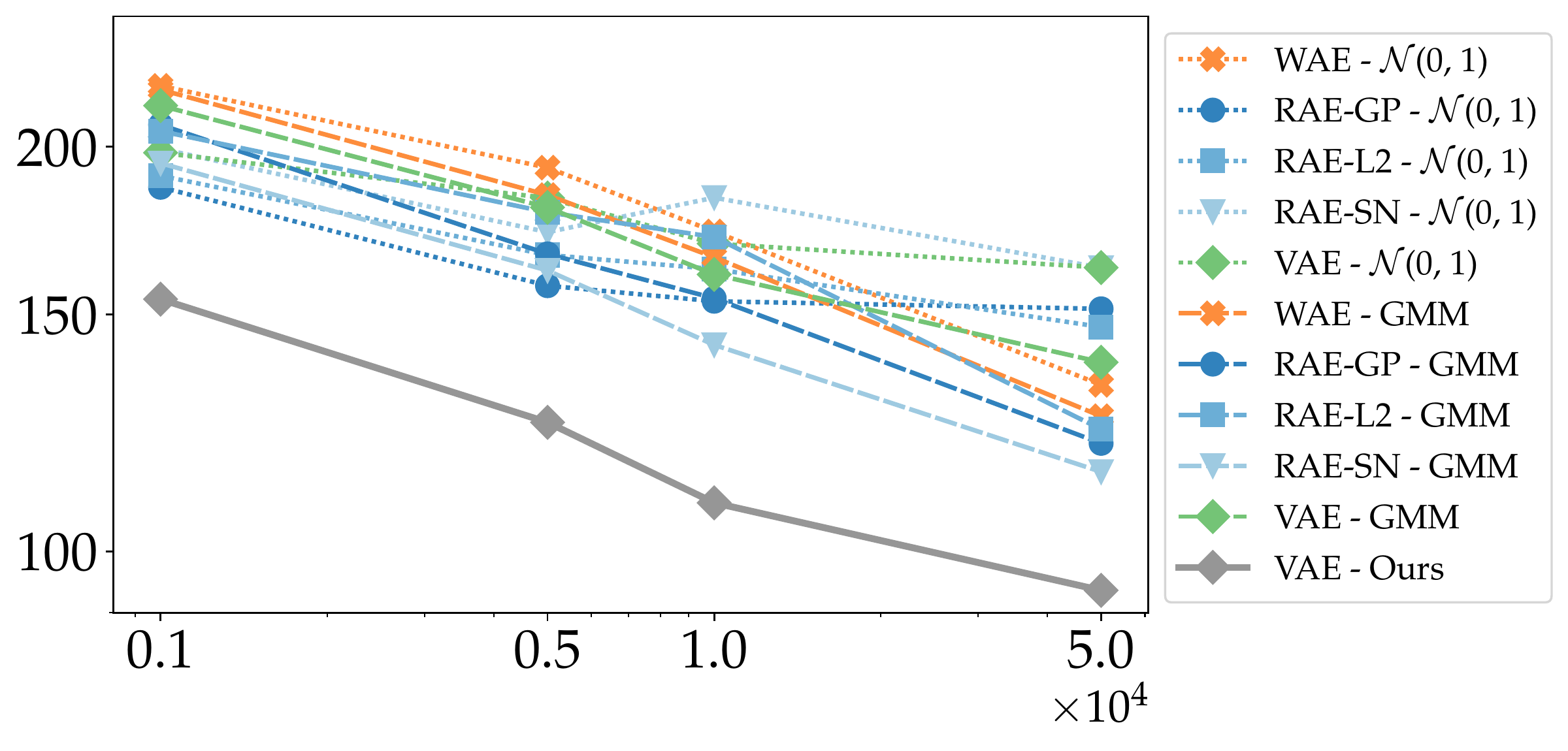}}
    \caption{Evolution of the FID score according to the number of training samples.}
    \label{fig: fid evolution}
    \end{figure}

\subsection{Investigating robustness in low data regime}
\label{Sec:Investigating Generation Robustness in Low Data Regime}

We perform a comparison using the same models and datasets as before but we progressively decrease the size of the training set to see the robustness of the methods according to the number of samples. Despite rarely performed in most generative models related papers, this setup appeared to us relevant since 1) it is well known that it may be challenging for these models, 2) in day-to-day applications collecting large databases may reveal costly if not impossible (\emph{e.g.} medicine). Hence, we consider MNIST, CIFAR10 and SVHN and use either the full dataset size, 10k, 5k or 1k training samples. For each experiment, the best retained model is again the one achieving the best ELBO on the validation set the size of which is set as 20\% of the train size. See Appendix~\ref{appD} for further details about experiments set-up. Then, we report the evolution of the FID against the test set in Figure~\ref{fig: fid evolution}. Results obtained on SVHN are presented in Appendix~\ref{appE}. Again, the proposed sampling method appears quite robust to the dataset size since it outperforms the other models' FID even when the number of training samples is smaller. This is made possible thanks to  the proposed metric that allows to avoid regions of the latent space having poor information. Finally, our study shows that although using more complex generation procedures such as ex-post density estimation seems to still enhance the generation capability of the model when the number of training samples remains quite high ($\geq$5k), this gain seems to worsen when the dataset size reduces as illustrated on CIFAR. In addition, we also evaluate the model on a data augmentation task with neuroimaging data from OASIS \citep{marcus_open_2007} mimicking a day-to-day scenario where the limited data regime is very common in Appendix~\ref{appC3}.

\section{Conclusion}

In this paper, we provided a geometric understanding of the latent space learned by a VAE and showed that it can actually be seen as a Riemannian manifold. We proposed a new natural generation process consisting in sampling from the intrinsic uniform distribution defined on this learned manifold. The proposed method was empirically shown to be competitive with more advanced versions of the VAEs using either more complex priors, ex-post density estimation, normalizing flows or other regularization schemes. Interestingly, the proposed method revealed good robustness properties in complex settings such as high dimensional data or low sample sizes and appeared to benefit more recent VAE models as well. Future work would consist in trying to use this method to perform data augmentation in those challenging contexts and compare its reliability for such a task with state of the art methods or trying to use this metric to perform clustering in the latent space.

\clearpage

\begin{ack}
The research leading to these results has received funding from the French government under management of Agence Nationale de la Recherche as part of the ``Investissements d'avenir'' program, reference ANR-19-P3IA-0001 (PRAIRIE 3IA Institute) and reference ANR-10-IAIHU-06 (Agence Nationale de la Recherche-10-IA Institut Hospitalo-Universitaire-6). 
This work was granted access to the HPC
resources of IDRIS under the allocation AD011013517 made by
GENCI (Grand Equipement National de Calcul Intensif).
\end{ack}

\clearpage
\medskip

\small
\bibliographystyle{plainnat}
\bibliography{references}

\section*{Checklist}


\begin{enumerate}

\item For all authors...
\begin{enumerate}
  \item Do the main claims made in the abstract and introduction accurately reflect the paper's contributions and scope?
    \answerYes{}
  \item Did you describe the limitations of your work?
    \answerYes{}
  \item Did you discuss any potential negative societal impacts of your work?
    \answerYes{} See Appendix~C.
  \item Have you read the ethics review guidelines and ensured that your paper conforms to them?
    \answerYes{}
\end{enumerate}

\item If you are including theoretical results...
\begin{enumerate}
  \item Did you state the full set of assumptions of all theoretical results?
    \answerYes{}
        \item Did you include complete proofs of all theoretical results?
    \answerYes{}{}
\end{enumerate}

\item If you ran experiments...
\begin{enumerate}
  \item Did you include the code, data, and instructions needed to reproduce the main experimental results (either in the supplemental material or as a URL)?
    \answerYes{}{} See supplementary materials.
  \item Did you specify all the training details (e.g., data splits, hyperparameters, how they were chosen)?
    \answerYes{}{} See Appendix~D.
        \item Did you report error bars (e.g., with respect to the random seed after running experiments multiple times)?
    \answerYes{}{}
        \item Did you include the total amount of compute and the type of resources used (e.g., type of GPUs, internal cluster, or cloud provider)?
    \answerYes{} See Sec.~5.1.
\end{enumerate}

\item If you are using existing assets (e.g., code, data, models) or curating/releasing new assets...
\begin{enumerate}
  \item If your work uses existing assets, did you cite the creators?
    \answerYes{}{}
  \item Did you mention the license of the assets?
    \answerYes{} See Appendix~D. We mention that we use code and data only when allowed by the license. 
  \item Did you include any new assets either in the supplemental material or as a URL?
    \answerYes{}{}
  \item Did you discuss whether and how consent was obtained from people whose data you're using/curating?
    \answerNo{} We used well-known and publicly available datasets.
  \item Did you discuss whether the data you are using/curating contains personally identifiable information or offensive content?
    \answerNo{} We used well-known and publicly available datasets.
\end{enumerate}

\item If you used crowdsourcing or conducted research with human subjects...
\begin{enumerate}
  \item Did you include the full text of instructions given to participants and screenshots, if applicable?
    \answerNA{}{}
  \item Did you describe any potential participant risks, with links to Institutional Review Board (IRB) approvals, if applicable?
    \answerNA{}{}
  \item Did you include the estimated hourly wage paid to participants and the total amount spent on participant compensation?
    \answerNA{}
\end{enumerate}

\end{enumerate}

\clearpage
\appendix

\newpage
\appendix
\onecolumn

\section{Further elements on Riemannian geometry}\label{appA}

A $d$-dimensional Riemannian manifold $\mathcal{M}$ can be defined as a $d$-dimensional differentiable manifold equipped with is a smooth inner product $g: z \to \langle \cdot | \cdot \rangle_z$ defined on each tangent space $T_z\mathcal{M}$ of the manifold with $z \in \mathcal{M}$. A chart (or coordinate system) $(U, \phi)$ is a homeomorphism mapping an open set $U$ of the manifold to an open set $V$ of an Euclidean space.  Given $z \in U$, a chart $\phi: (z^1, \dots, z^d)$ induces a basis $\Big (\frac{\partial}{\partial z^1}, \dots, \frac{\partial}{\partial z^d} \Big)_z $ on the tangent space $T_z\mathcal{M}$. Hence, the metric of a Riemannian manifold can be locally represented in the chart $\phi$ as a positive definite matrix as mentionned in Sec.~4.1.
\begin{equation}\label{eq. App A Riemannian metric}
    \mathbf{G}(z) = ( g_{i, j})_{z, 0 \leq i, j \leq d} = \Big(\Big\langle \frac{\partial}{\partial z^i} | \frac{\partial}{\partial z^j} \Big\rangle_z\Big)_{0 \leq i, j \leq d}\,,
\end{equation} for each point $z$ of the manifold. That is for $v, w \in T_z\mathcal{M}$ and $z \in \mathcal{M}$, the inner product writes $\langle u | w \rangle_z = u^{\top} \mathbf{G}(z) w$. Assuming that the manifold is also connected, for any $z_1, z_2 \in \mathcal{M}$, two points of the manifold, we can consider a curve $\gamma$ traveling in $\mathcal{M}$ and parametrized by $ t \in [a, b]$ such that $\gamma(a) = z_1$ and $\gamma(b) = z_2$. Then, the length of $\gamma$ is given by
\[
    L(\gamma) = \int \limits _a ^b \lVert \dot{\gamma}(t) \rVert_{\gamma(t)} dt = \int \limits _a ^b \sqrt{\langle \dot{\gamma}(t) | \dot{\gamma}(t) \rangle_{\gamma(t)}} dt
\]  
Curves $\gamma$ that minimize $L$ and are parameterized proportionally to the arc length are called \textit{geodesic} curves. A distance $\mathrm{dist}_{\mathbf{G}}$ on the manifold $\mathcal{M}$ can then be derived and writes
\begin{equation}\label{Eq: geodesic distance}
    \mathrm{dist}_{\mathbf{G}}(z_1, z_2) = \inf_{\gamma} L(\gamma) \hspace{5mm} \mathrm{s.t.} \hspace{5mm} \gamma(a) = z_1, \gamma(b) = z_2
\end{equation}
The manifold $\mathcal{M}$ is said to be \textit{geodesically complete} if all geodesic curves can be extended to $\mathbb{R}$. Given the Riemannian manifold $\mathcal{M}$ endowed with the Riemannian metric $\mathbf{G}$ and a chart $z$, an infinitesimal volume element may also be defined on each tangent space $T_{z}$ of the manifold $\mathcal{M}$ as follows
\begin{equation}
    d \mathcal{M}_z = \sqrt{\det \mathbf{G}(z)} dz\,,
\end{equation}
with $dz$ being the Lebesgue measure. This defines a canonical measure on the manifold and allows to extend the notion of probability distributions to Riemannian manifolds. In particular, such a property allows to refer to random variables with a density defined with respect to the measure on the manifold. We recall such definition from \cite{pennec_intrinsic_2006} below

\begin{definition}
Let $\mathcal{B}(\mathcal{M})$ be the Borel $\sigma$-algebra of $\mathcal{M}$. The random point $\mathbf{z}$ has a probability density function $\rho_{\mathbf{z}}$ if:
\[
    \forall \mathcal{Z} \in \mathcal{B}(\mathcal{M}), \hspace{2mm} \mathbb{P}(\mathbf{z} \in \mathcal{Z}) = \int \limits _{\mathcal{Z}} \rho(z) d\mathcal{M}(z) \hspace{2mm}
    \text{and} \hspace{2mm} \int \limits _{\mathcal{M}} \rho(z) d\mathcal{M}(z) = 1
\]
\end{definition}

Finally, given a chart $\phi$ defined on the whole manifold $\mathcal{M}$ and a random point $\mathbf{z}$ on $\mathcal{M}$, the point $\mathbf{p} = \phi(\mathbf{z})$ is a random point whose density $\rho'_{\mathbf{p}}$ may be written with respect to the Lebesgue measure as such \cite{pennec_intrinsic_2006}:

\begin{equation}\label{Eq: Riemann to Lebesgue}
    \rho'_{\mathbf{p}}(p) = \rho_{\mathbf{z}}(\phi^{-1}(p)) \sqrt{\det g(\phi^{-1}(p))} 
\end{equation}

\clearpage
\section{The generation process algorithm - Implementation details}\label{appB}

In this appendix, we provide pseudo-code algorithms explaining how to build the metric from a trained VAE and how to use the proposed sampling process. Noteworthy is the fact that we do not amend the training process of the vanilla VAE which remains pretty simple and stable.  

\subsection{Building the metric}
In this section, we explain how to build the proposed Riemannian metric. For the sake of clarity, we recall the expression of the metric below \begin{equation}\label{eg: Riemannian metric app}
     \mathbf{G}(z) = \sum \limits_{i=1}^N \mathbf{\Sigma}^{-1}(x_i) \cdot \omega_i(z) + \lambda \cdot e^{-\tau \lVert z\rVert_2^2} \cdot I_d\,,
\end{equation}
where
\[
    \omega_i(z) = \exp \Bigg ( -\frac{\mathrm{dist}_{\mathbf{\Sigma}^{-1}(x_i)}(z, \mu(x_i))^2}{\rho^2}\Bigg) = \exp \Bigg ( -\frac{(z - \mu(x_i))^{\top} \mathbf{\Sigma}^{-1}(x_i)(z - \mu(x_i))}{\rho^2}\Bigg)\,,
\]

\begin{algorithm}[ht]
     \caption{Building the metric from a trained model}
     \label{Alg:build metric}
    \begin{algorithmic}
    \State{\bfseries Input:} A trained VAE model $m$, the training dataset $\mathcal{X}$, $\lambda$, $\tau$\; \algorithmiccomment{In practice $\tau$ $\approx 0$}
    \For{$x_i \in \mathcal{X}$ }
    \State{$\mu_i, \mathbf{\Sigma}_i = m(x_i)$}\;  \algorithmiccomment{Retrieve training embeddings and covariance matrices}
    \EndFor
    \State{Select $k$ centroids $c_i$ in the $\mu_i$}\; \algorithmiccomment{e.g. with $k$-medoids}
    \State{Get corresponding covariance matrices $\mathbf{\Sigma}_i$}\;
    \State{$\rho \leftarrow \max \limits _i \min \limits _{j\neq i} \Vert c_i -c_j\Vert_2$}\; \algorithmiccomment{Set $\rho$ to the max distance between two closest neighbors}
    \State{Build the metric using Eq.~\eqref{eg: Riemannian metric app}}\;
    \[
    \mathbf{G}(z) = \sum \limits_{i=1}^N \mathbf{\Sigma}^{-1}_i \cdot \omega_i(z) + \lambda \cdot e^{-\tau \lVert z\rVert_2^2} \cdot I_d
    \]
    \State{\bfseries Return $\mathbf{G}$} \algorithmiccomment{Return $\mathbf{G}$ as a function}
    \end{algorithmic}
\end{algorithm}

As is standard in VAE implementations, we assume that the covariance matrices $\mathbf{\Sigma}_i$ given by the VAE are diagonal and that the encoder outputs a mean vector and the log of the diagonal coefficients. In the implementation, the exponential is then applied to recover the $\mathbf{\Sigma}_i$ so that no singular matrix arises.   

\subsection{Sampling process}

Further to the description performed in the paper, we provide here a detailed algorithm stating the main steps of the generation process. 

\subsubsection{The HMC sampler}
In the sampling process we propose to rely on the Hamiltonian Monte Carlo sampler to sample from the Riemanian uniform distribution. In a nutshell, the HMC sampler aims at sampling from a target distribution $p_{\mathrm{target}}(z)$ with $z \in \mathbb{R}^d$ using Hamiltonian dynamics. The main idea behind such a sampler is to introduce an auxiliary random variable $v \sim \mathcal{N}(0, I_d)$ independent from $z$ and mimic the behavior of a particle having $z$ (resp. $v$) as location (resp. velocity). The Hamiltonian of the particle then writes 
\[
    H(z, v) = U(z) + K(v)\,,
\]
where $U(z)$ is the potential energy and $K(v)$ is its kinetic energy both given by
\[
    U(z) = - \log p_{\mathrm{target}}(z), \hspace{5mm} K(v) = \frac{1}{2}v^{\top} v
\]
The following Hamilton's equations govern the evolution in time of the particle.
\begin{equation}\label{eq: Hamilton}\left\{
    \begin{array}{cll}
        \frac{\partial H(z, v)}{\partial v} =& v \,,\\
        \frac{\partial H(z, v)}{\partial z}   =& - \nabla_{z} \log p_{\mathrm{target}}(z)\,.
    \end{array}\right.
\end{equation}
In order to integrate these equations, recourse to the leapfrog integrator is needed and consists in applying $n_{\mathrm{lf}}$ times the following equations.

\begin{equation}\label{Eq: leapfrog}\left\{
    \begin{array}{cl}
        v(t + \frac{\varepsilon_{\mathrm{lf}}}{2})  &= v(t) + \frac{\varepsilon_{\mathrm{lf}}}{2} \cdot \nabla_{z}\log p_{\mathrm{target}}(z(t))\,, \\
        z(t+ \varepsilon_{\mathrm{lf}})              &= z(t) + \varepsilon_{\mathrm{lf}}\cdot v(t + \frac{\varepsilon_{\mathrm{lf}}}{2})\,, \\
        v(t + \varepsilon_{\mathrm{lf}})             &= v(t + \frac{\varepsilon_{\mathrm{lf}}}{2}) + \frac{\varepsilon_{\mathrm{lf}}}{2} \cdot \nabla_{z}\log p_{\mathrm{target}}(z(t + \varepsilon_{\mathrm{lf}}))\,,
    \end{array}\right.
\end{equation}
where $\varepsilon_{\mathrm{lf}}$ is called the leapfrog step size. This algorithm produces a proposal $(\widetilde{z}, \widetilde{v})$ that is accepted with probability $\alpha$ where
\[
    \alpha = \min\Big(1, \exp \Big({H(z, v) - H(\widetilde{z}, \widetilde{v}} )\Big)\Big )\,.
\]
This procedure is then repeated to create an ergodic Markov chain $(z^n)$ converging to the distribution $p_{\mathrm{target}}$ \cite{duane_hybrid_1987, liu_monte_2008, neal_mcmc_2011, girolami_riemann_2011}.

\subsection{The proposed algorithm}

In our setting the target density is given by the density of the Riemannian uniform distribution which writes with respect to Lebesgue measure as follows
\begin{equation} \label{Eq: Uniform Riemann}
    p(z) = \mathcal{U}_{\text{Riem}}(z) = \frac{1}{C}\sqrt{\det \mathbf{G}(z)} \hspace{2em} C = \int_{\mathbb{R}^d}\sqrt{\det \mathbf{G}(z)}dz \,.
\end{equation}
Note that thanks to the shape of the metric, this distribution is well defined since $C < +\infty$. The log density follows
\[
\log p(z) = \frac{1}{2} \log \det \mathbf{G}(z) - \log C\,,
\]
Hence, the Hamiltonian writes
\[
    H(z, v) = -\log p(z) + \frac{1}{2} v^{\top}v\,,
\]
and Hamilton's equations become
\begin{equation*}\left\{
    \begin{array}{cl}
        \frac{\partial H(z, v)}{\partial v} =& v \,,\\
         \frac{\partial H(z, v)}{\partial z_i} =&- \frac{\partial \log p(z)}{\partial z_i} = -\frac{1}{2} \mathrm{tr}\Big(\mathbf{G}^{-1}(z)\frac{\partial \mathbf{G}(z)}{\partial z_i}\Big )
    \end{array}\right.
\end{equation*}

Since the covariance matrices are supposed to be diagonal as is standard in VAE implementations, the computation of the inverse metric is straightforward. Moreover, since $\mathbf{G}(z)$ is smooth and has a closed form, it can be differentiated with respect to $z$ pretty easily. Now, the leapfrog integrator given in Eq.~\eqref{Eq: leapfrog} can be used and the acceptance ratio $\alpha$ is easy to compute. Noteworthy is the fact that the normalizing constant $C$ is never needed since it vanishes in the gradient computation and simplifies in the acceptance ratio $\alpha$. We provide a pseudo-code of the proposed sampling procedure in Alg.~\ref{Alg:proposed algo}. A typical choice in the sampler's hyper-parameters used in the paper is $N=100$, $n_{\mathrm{lf}}=10$ and $\varepsilon_{\mathrm{lf}}=0.01$. The initialization of the chain can be done either randomly or on points that belong to the manifold (i.e. the centroids $c_i$ or $\mu(x_i)$).

\renewcommand\footnoterule{} 
\begin{algorithm}[ht]
     \caption{Proposed sampling process}
     \label{Alg:proposed algo}
    \begin{algorithmic}
    \State{\bfseries Input:} The metric function $\mathbf{G}$, hyper-parameters of the HMC sampler (chain length $N$, number of leapfrog steps $n_{\mathrm{lf}}$, leapfrog step size $\varepsilon_{\mathrm{lf}}$)
    \State{\bfseries Initialization: $z$}\; \algorithmiccomment{Initialize the chain}
    \For{$i=1 \rightarrow N$}
    \State{$v \sim \mathcal{N}(0, I_d)$}\; \algorithmiccomment{Draw a velocity}
    \State{$H_0 \leftarrow H(z, v)$}\; \algorithmiccomment{Compute the starting Hamiltonian}
    \State{$z_0 \leftarrow z$}\;
    \For{$k=1 \leftarrow n_{\mathrm{lf}}$}
    \State{$\bar{v} \leftarrow v  - \frac{\varepsilon_{\mathrm{lf}}}{2} \cdot \nabla_{z} H(z, v)$}\;
    \State{$\widetilde{z} \leftarrow z + \varepsilon_{\mathrm{lf}} \cdot \bar{v}$} \; \algorithmiccomment{Leapfrog step Eq.~\eqref{Eq: leapfrog}}
    \State{$\widetilde{v} \leftarrow \bar{v}  - \frac{\varepsilon_{\mathrm{lf}}}{2} \cdot \nabla_{z} H(\widetilde{z}, \bar{v})$}\;
    \State{$v \leftarrow \widetilde{v}$}\;
    \State{$z \leftarrow \widetilde{z}$}\;
    \EndFor
    \State{$H \leftarrow H(\widetilde{z}, \widetilde{v})$}\; \algorithmiccomment{Compute the ending Hamiltonian}
    \State{Accept $\widetilde{z}$ with probability $\alpha = \min \Big ( 1, \exp(H_0-H)\Big)$}
    \If{Accepted}
        \State{$z \leftarrow \widetilde{z}$}\;
    \Else
        \State{$z \leftarrow z_0$}
    \EndIf
    \EndFor
    \State{{\bfseries Return} $z$}
    \end{algorithmic}
\end{algorithm}

\clearpage
\section{Other generation results}\label{appC}

\subsection{Some further samples on CELEBA and MNIST}
\begin{figure}[ht]
    \centering
    \captionsetup[subfigure]{position=above, labelformat = empty}
    \subfloat{\includegraphics[width=3.in]{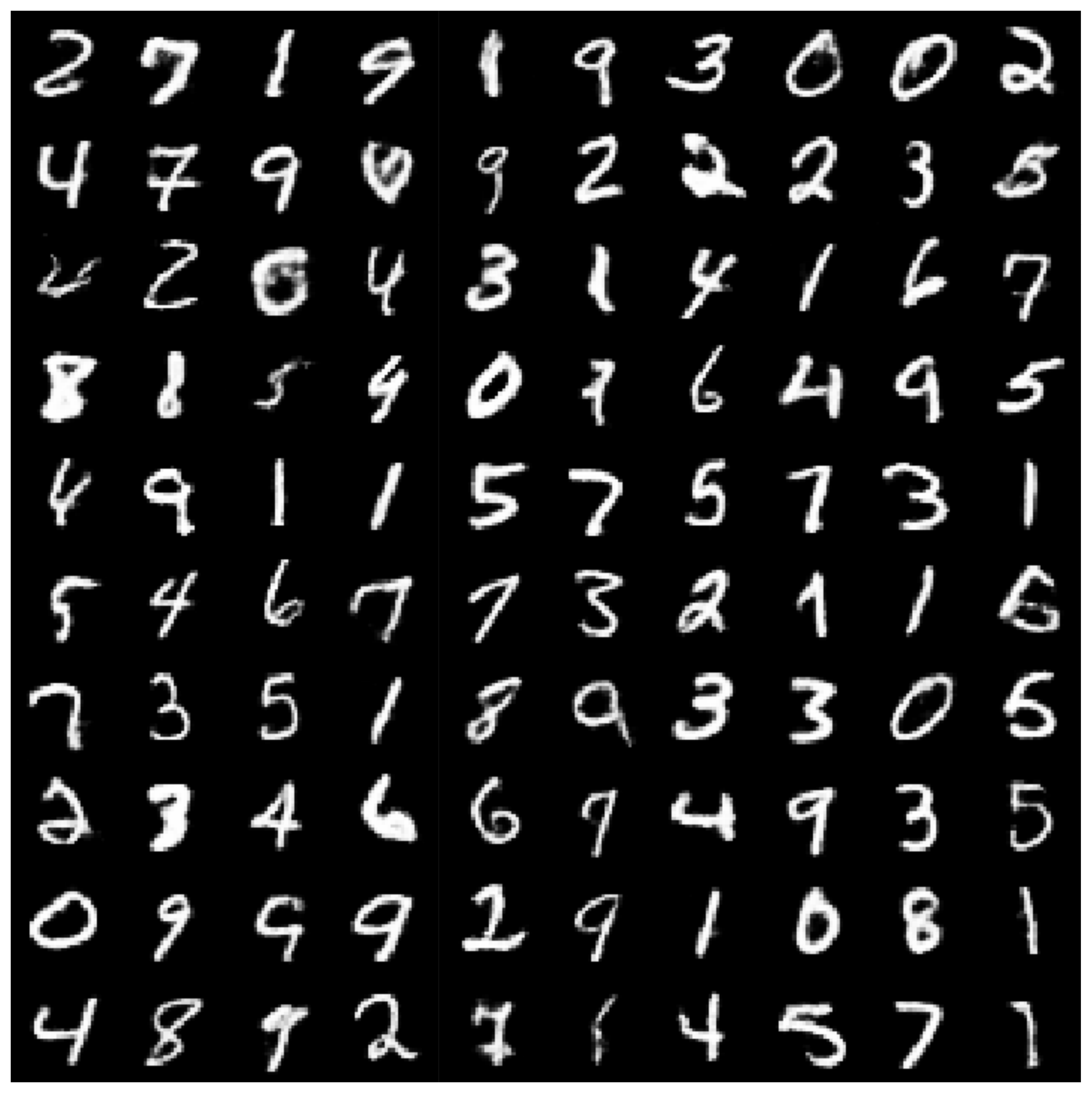}}
    \caption{100 samples with the proposed method on MNIST dataset.}
    \label{fig:100 mnist}
\end{figure}

\begin{figure}[ht]
    \centering
    \captionsetup[subfigure]{position=above, labelformat = empty}
    \subfloat{\includegraphics[width=3.in]{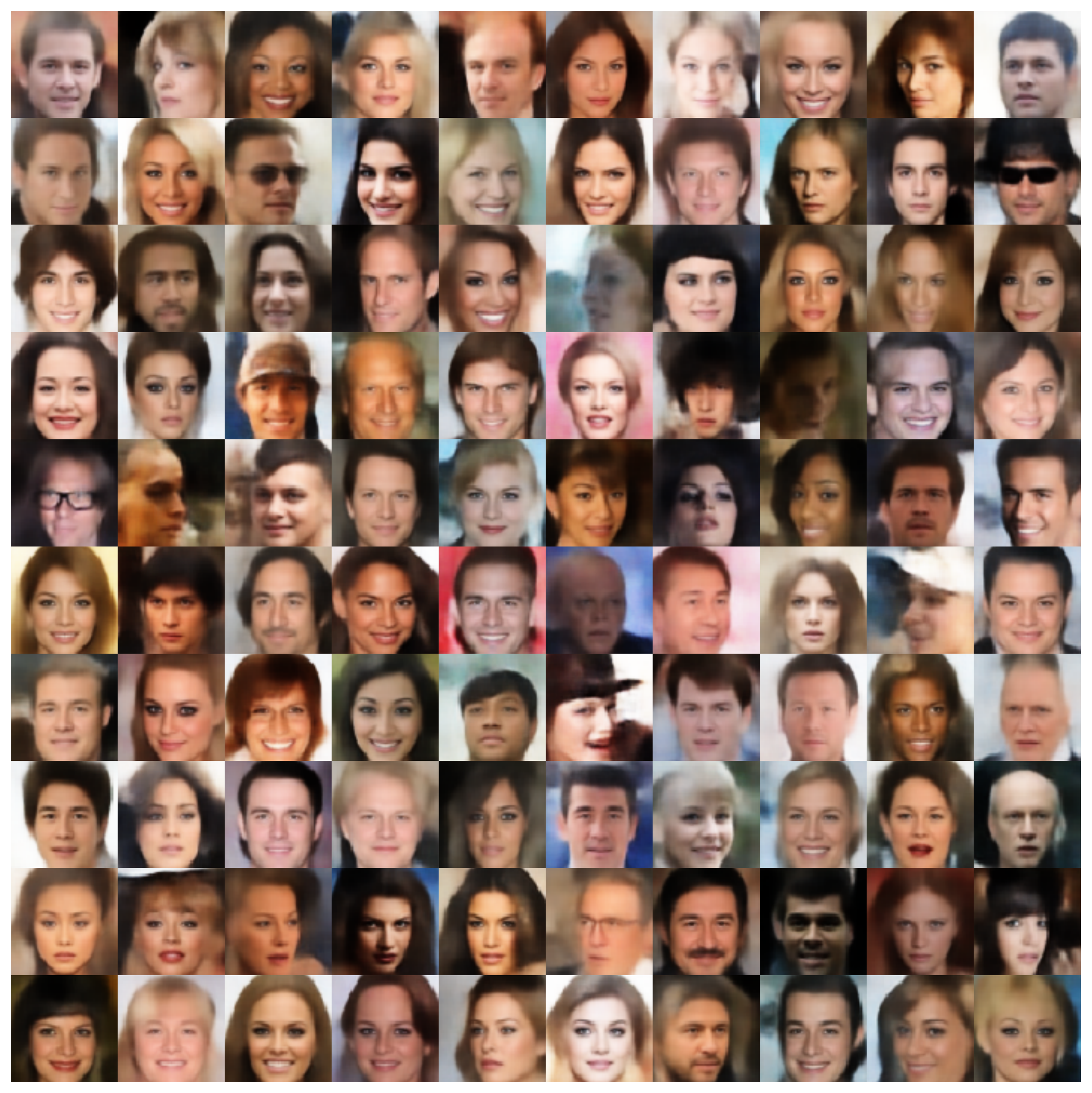}}
    \caption{100 samples with the proposed method on CELEBA dataset.}
    \label{fig:100 celeba}
\end{figure}

\begin{figure}[ht]
    \centering
    \captionsetup[subfigure]{position=above, labelformat = empty}
    \adjustbox{minipage=5em,raise=\dimexpr -3.\height}{\small AE - $\mathcal{N}$}
    \subfloat[MNIST]{\includegraphics[width=2.3in]{plots/sampling/RAE/AE_generated_mnist_gaussian.pdf}}
    \subfloat[CELEBA]{\includegraphics[width=2.3in]{plots/sampling/RAE/AE_generated_celeba_gaussian.pdf}}\\
    \vspace{-5mm}
    \adjustbox{minipage=5em,raise=\dimexpr -3.\height}{\small VAE - $\mathcal{N}$}
    \subfloat{\includegraphics[width=2.3in]{plots/sampling/generated_mnist_gaussian.pdf}}
    \subfloat{\includegraphics[width=2.3in]{plots/sampling/generated_celeba_gaussian.pdf}}\\
    \vspace{-5mm}
    \adjustbox{minipage=5em,raise=\dimexpr -3.\height}{\small WAE}
    \subfloat{\includegraphics[width=2.3in]{plots/sampling/WAE/WAE_generated_mnist.pdf}}
    \subfloat{\includegraphics[width=2.3in]{plots/sampling/WAE/WAE_generated_celeba.pdf}}\\
    \vspace{-5mm}
    \adjustbox{minipage=5em,raise=\dimexpr -3.\height}{\small VAMP}
    \subfloat{\includegraphics[width=2.3in]{plots/sampling/VAMP/VAMP_generated_mnist.pdf}}
    \subfloat{\includegraphics[width=2.3in]{plots/sampling/VAMP/VAMP_generated_celeba.pdf}}\\
    \vspace{-5mm}
     \adjustbox{minipage=5em,raise=\dimexpr -3.\height}{\small HVAE}
    \subfloat{\includegraphics[width=2.3in]{plots/sampling/HVAE/HVAE_mnist_gaussian_prior.pdf}}
    \subfloat{\includegraphics[width=2.3in]{plots/sampling/HVAE/HVAE_celeba_gaussian_prior.pdf}}\\
    \vspace{-5mm}
    \adjustbox{minipage=5em,raise=\dimexpr -3.\height}{\small RHVAE}
    \subfloat{\includegraphics[width=2.3in]{plots/sampling/RHVAE/RHVAE_mnist_gaussian_prior.pdf}}
    \subfloat{\includegraphics[width=2.3in]{plots/sampling/RHVAE/RHVAE_celeba_gaussian_prior.pdf}}\\
    \vspace{-5mm}
    \adjustbox{minipage=5em,raise=\dimexpr -3.\height}{\small AE - GMM}
    \subfloat{\includegraphics[width=2.3in]{plots/sampling/RAE/AE_generated_mnist_gmm.pdf}}
    \subfloat{\includegraphics[width=2.3in]{plots/sampling/RAE/AE_generated_celeba_gmm.pdf}}\\
    \vspace{-5mm}
    \adjustbox{minipage=5em,raise=\dimexpr -3.\height}{\small VAE - GMM}
    \subfloat{\includegraphics[width=2.3in]{plots/sampling/generated_mnist_gmm.pdf}}
    \subfloat{\includegraphics[width=2.3in]{plots/sampling/generated_celeba_gmm.pdf}}\\
    \vspace{-5mm}
    \adjustbox{minipage=5em,raise=\dimexpr -3.\height}{\small RAE (GP)}
    \subfloat{\includegraphics[width=2.3in]{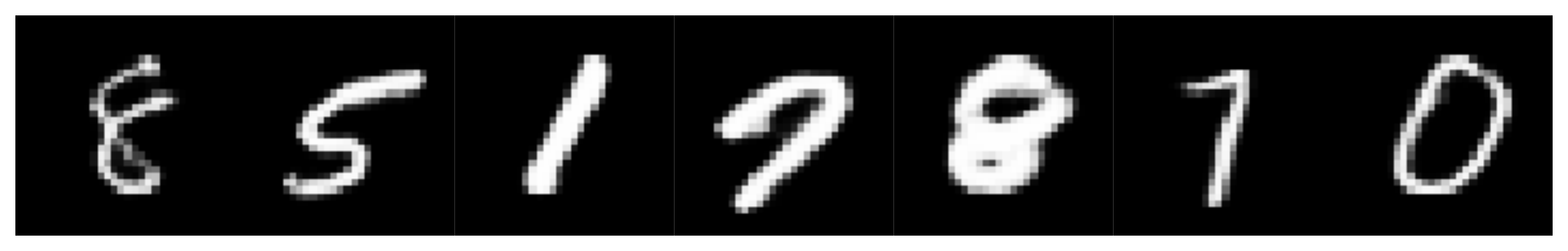}}
    \subfloat{\includegraphics[width=2.3in]{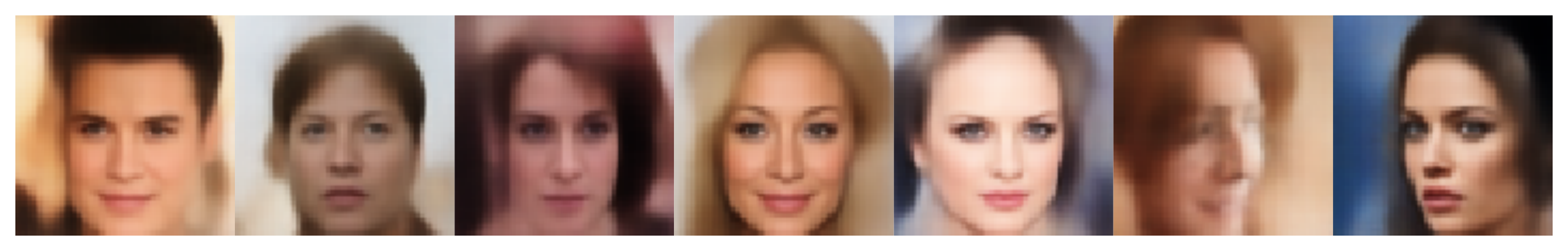}}\\
    \vspace{-5mm}
    \adjustbox{minipage=5em,raise=\dimexpr -3.\height}{\small RAE (L2)}
    \subfloat{\includegraphics[width=2.3in]{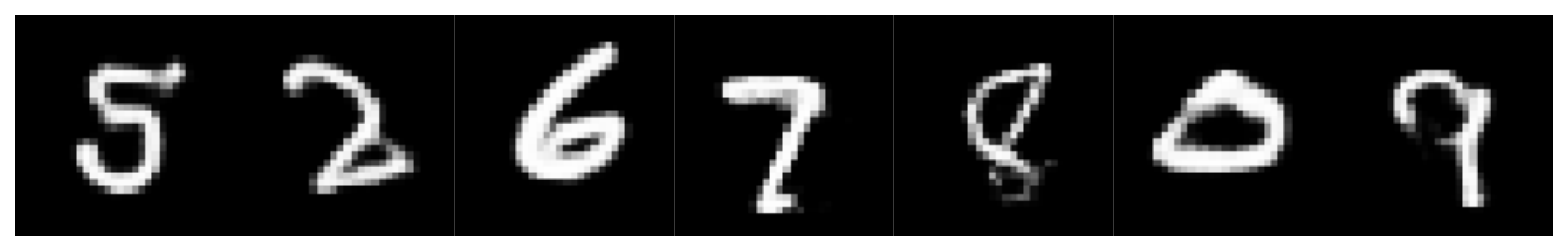}}
    \subfloat{\includegraphics[width=2.3in]{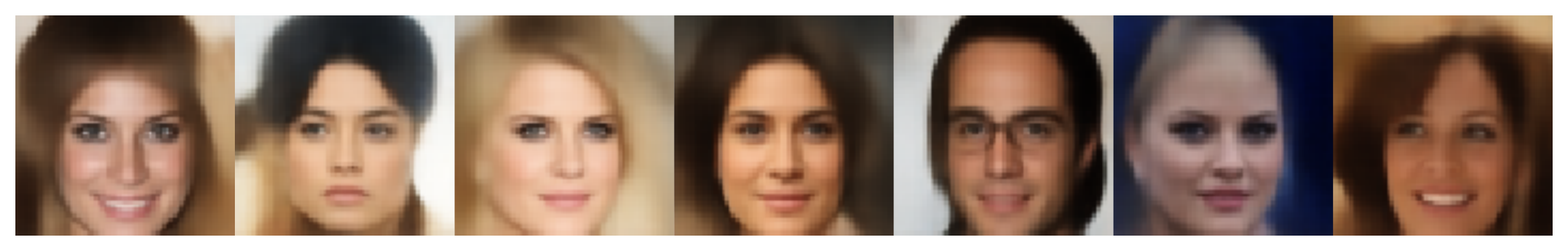}}\\
    \vspace{-5mm}
    \adjustbox{minipage=5em,raise=\dimexpr -3.\height}{\small RAE (SN)}
    \subfloat{\includegraphics[width=2.3in]{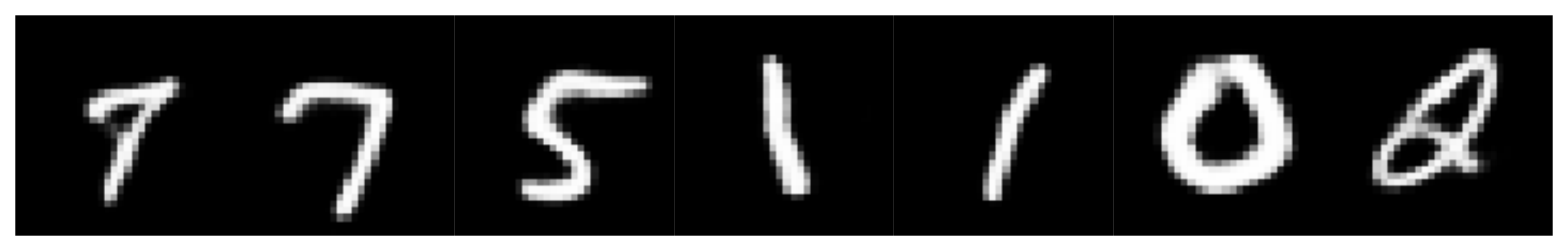}}
    \subfloat{\includegraphics[width=2.3in]{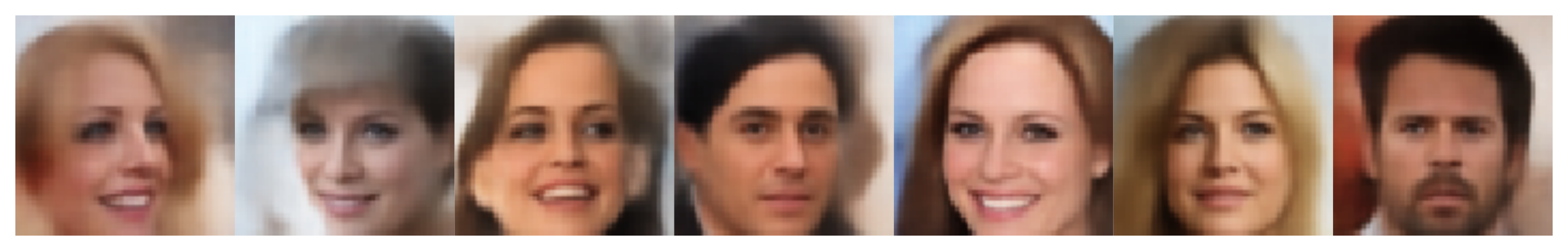}}\\
    \vspace{-5mm}
    \adjustbox{minipage=5em,raise=\dimexpr -3.\height}{\small RAE}
    \subfloat{\includegraphics[width=2.3in]{plots/sampling/RAE/RAE_generated_mnist_gmm.pdf}}
    \subfloat{\includegraphics[width=2.3in]{plots/sampling/RAE/RAE_generated_celeba_gmm.pdf}}\\
    \vspace{-5mm}
    \adjustbox{minipage=5em,raise=\dimexpr -3.\height}{\small VAE - Ours}
    \subfloat{\includegraphics[width=2.3in]{plots/sampling/generated_mnist.pdf}}
    \subfloat{\includegraphics[width=2.3in]{plots/sampling/generated_celeba.pdf}}

    \caption{Generated samples with different models and generation processes.}
    \label{fig: appendix generated samples}
    \end{figure}

In this section, we provide some further generated samples using the proposed method. Figure~\ref{fig:100 mnist} and Figure~\ref{fig:100 celeba} again support the fact that the method is able to generate sharp and diverse samples. We also add the other variants of the RAE model in Figure~\ref{fig: appendix generated samples}.

\clearpage
\subsection{CIFAR and SVHN}
In this appendix, we gather the resulting samplings from the different models considered for SVHN and CIFAR 10.

\begin{figure}[ht]
    \centering
    \captionsetup[subfigure]{position=above, labelformat = empty}
    \adjustbox{minipage=5em,raise=\dimexpr -3.\height}{\small AE - $\mathcal{N}$}
    \subfloat[SVHN]{\includegraphics[width=2.3in]{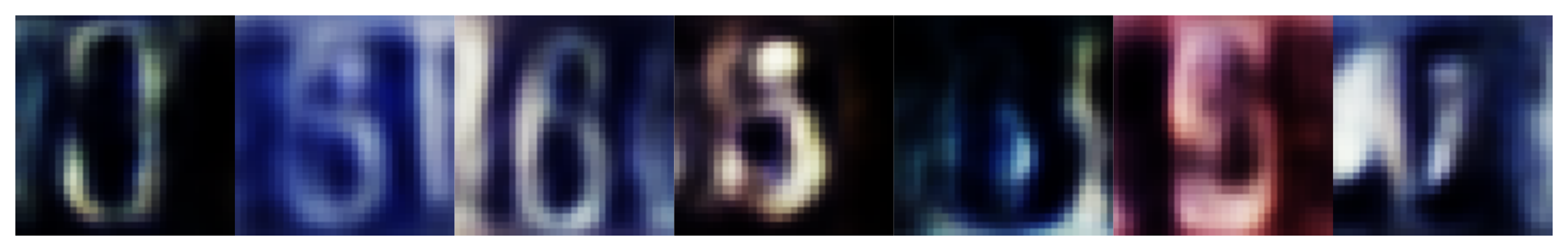}}
    \subfloat[CIFAR 10]{\includegraphics[width=2.3in]{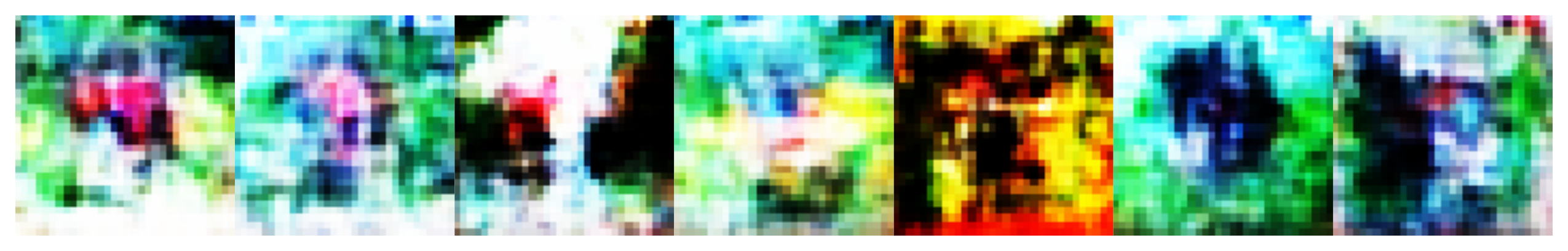}}\\
    \vspace{-5mm}
    \adjustbox{minipage=5em,raise=\dimexpr -3.\height}{\small VAE - $\mathcal{N}$}
    \subfloat{\includegraphics[width=2.3in]{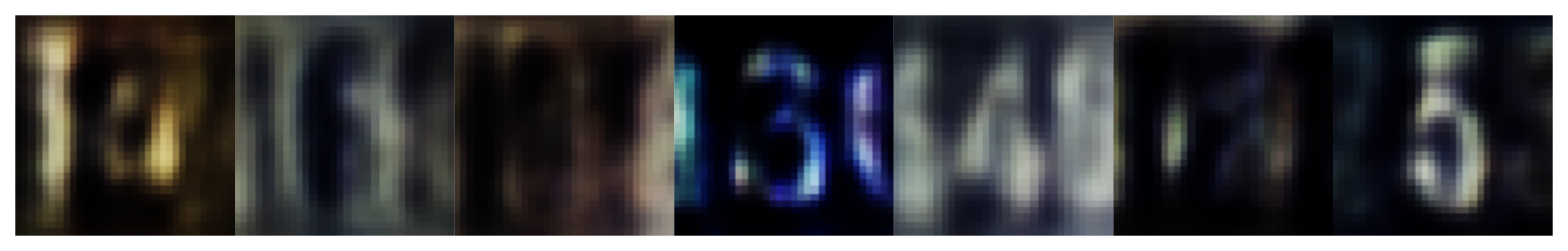}}
    \subfloat{\includegraphics[width=2.3in]{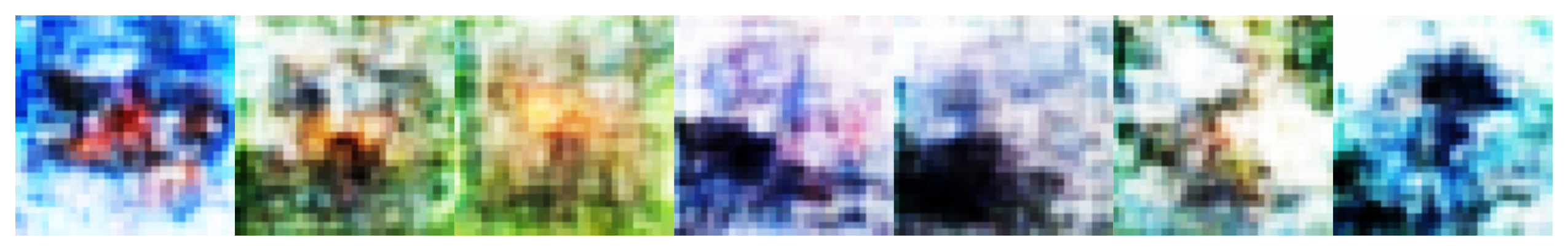}}\\
    \vspace{-5mm}
    \adjustbox{minipage=5em,raise=\dimexpr -3.\height}{\small WAE}
    \subfloat{\includegraphics[width=2.3in]{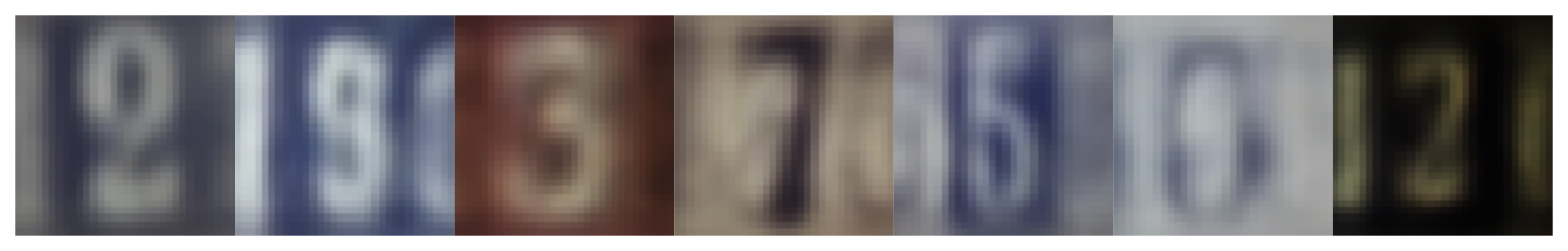}}
    \subfloat{\includegraphics[width=2.3in]{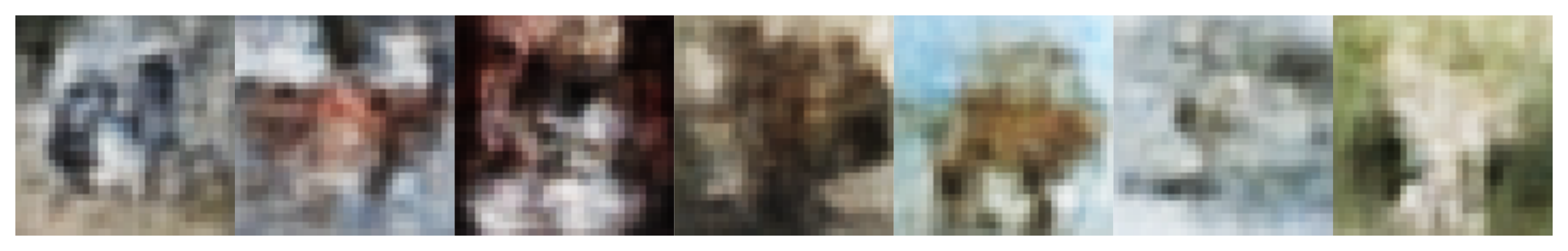}}\\
    \vspace{-5mm}
    \adjustbox{minipage=5em,raise=\dimexpr -3.\height}{\small VAMP}
    \subfloat{\includegraphics[width=2.3in]{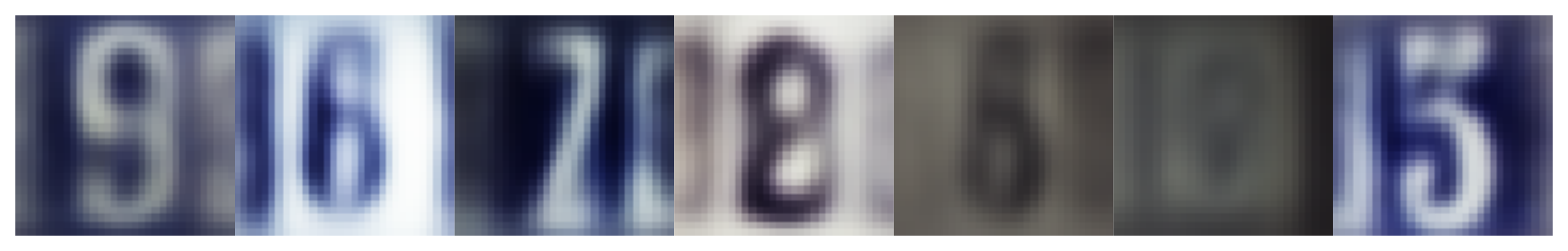}}
    \subfloat{\includegraphics[width=2.3in]{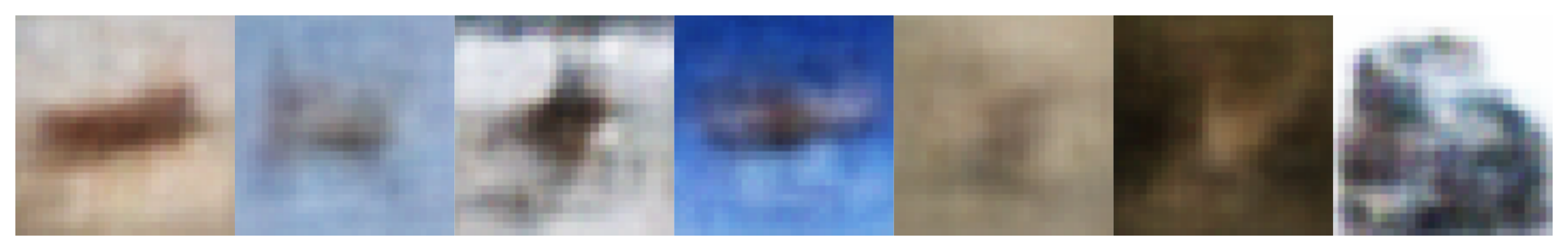}}\\
    \vspace{-5mm}
    \adjustbox{minipage=5em,raise=\dimexpr -3.\height}{\small HVAE}
    \subfloat{\includegraphics[width=2.3in]{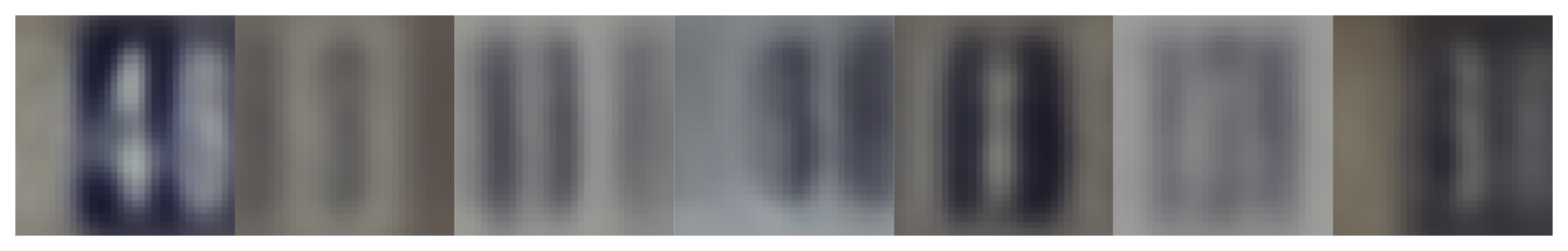}}
    \subfloat{\includegraphics[width=2.3in]{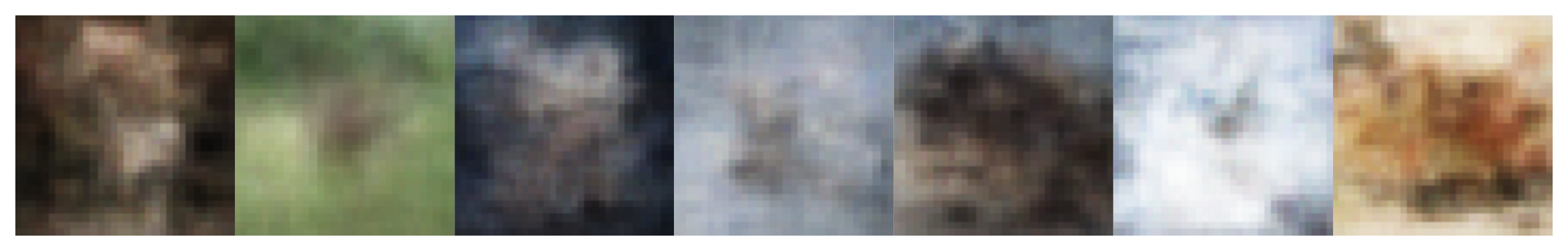}}\\
    \vspace{-5mm}
    \adjustbox{minipage=5em,raise=\dimexpr -3.\height}{\small RHVAE}
    \subfloat{\includegraphics[width=2.3in]{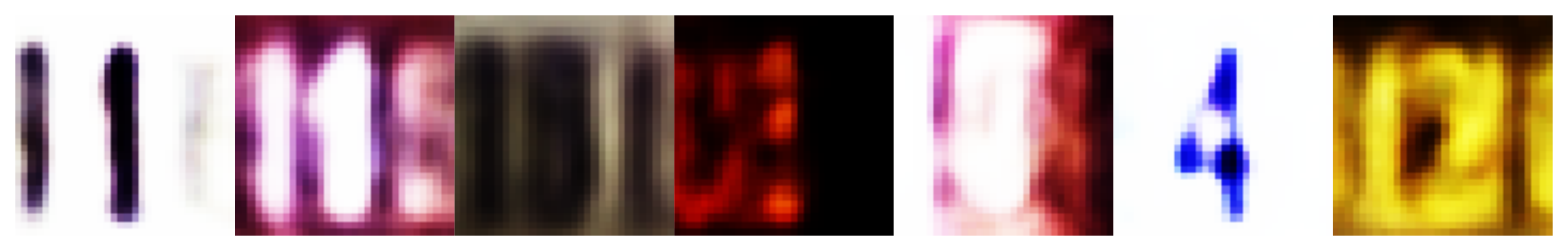}}
    \subfloat{\includegraphics[width=2.3in]{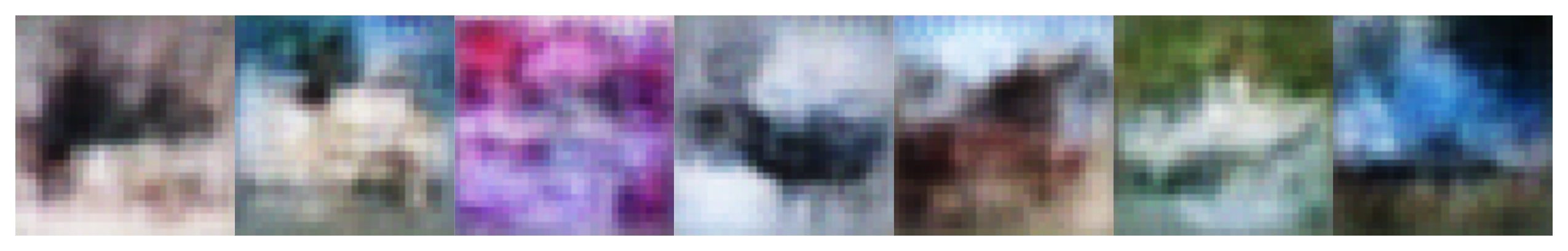}}\\
    \vspace{-5mm}
    \adjustbox{minipage=5em,raise=\dimexpr -3.\height}{\small AE - GMM}
    \subfloat{\includegraphics[width=2.3in]{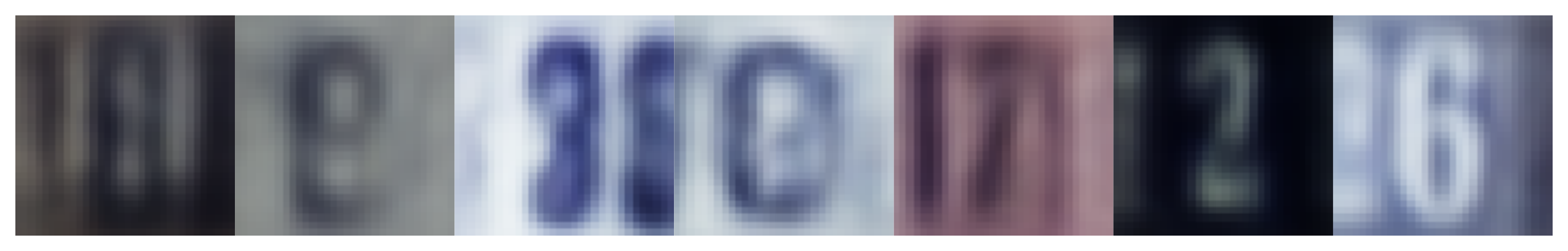}}
    \subfloat{\includegraphics[width=2.3in]{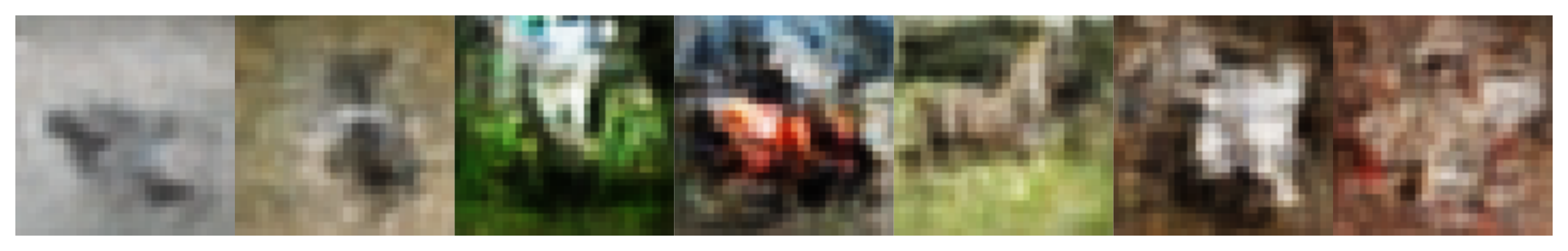}}\\
    \vspace{-5mm}
    \adjustbox{minipage=5em,raise=\dimexpr -3.\height}{\small VAE - GMM}
    \subfloat{\includegraphics[width=2.3in]{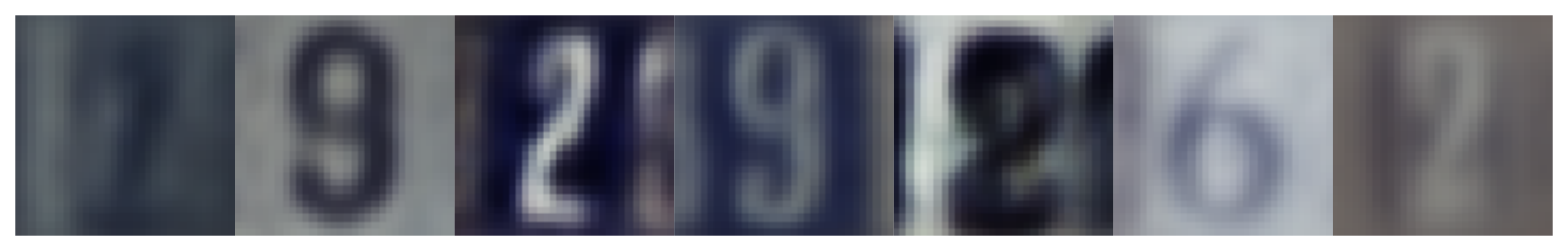}}
    \subfloat{\includegraphics[width=2.3in]{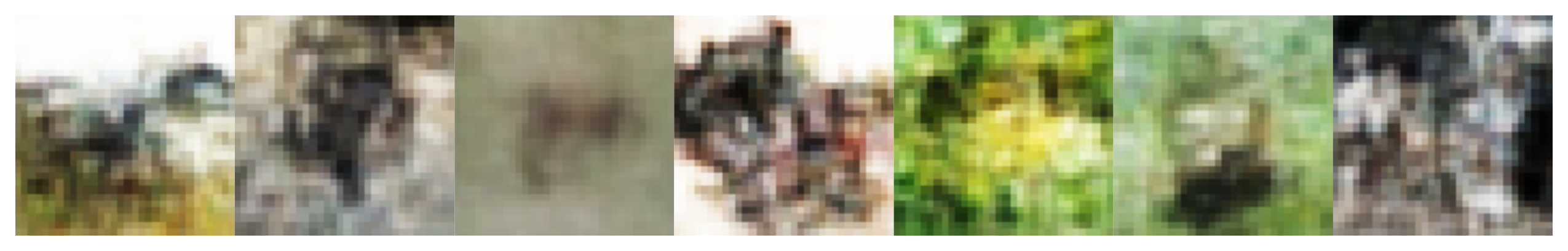}}\\
    \vspace{-5mm}
    \adjustbox{minipage=5em,raise=\dimexpr -3.\height}{\small RAE (GP)}
    \subfloat{\includegraphics[width=2.3in]{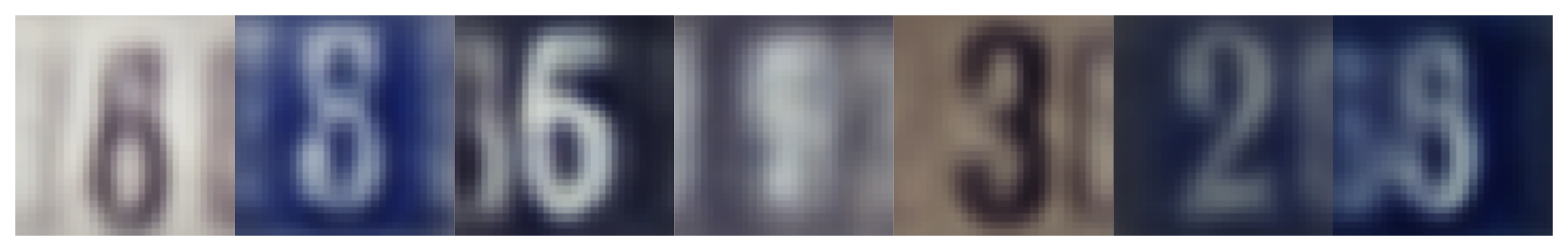}}
    \subfloat{\includegraphics[width=2.3in]{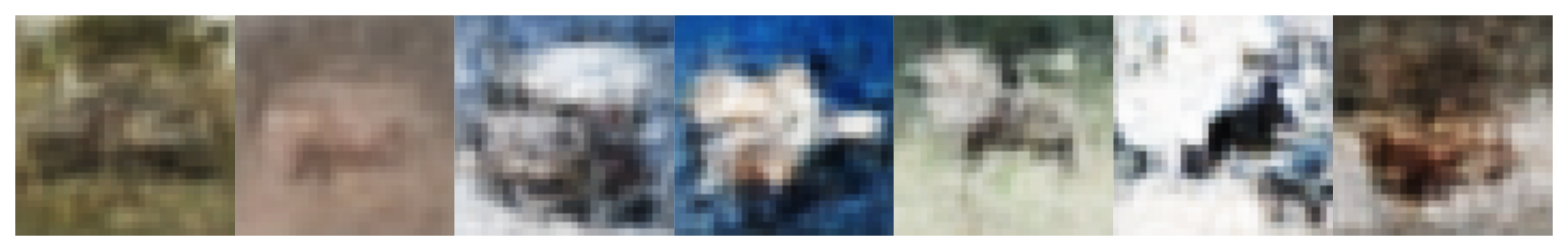}}\\
    \vspace{-5mm}
    \adjustbox{minipage=5em,raise=\dimexpr -3.\height}{\small RAE (L2)}
    \subfloat{\includegraphics[width=2.3in]{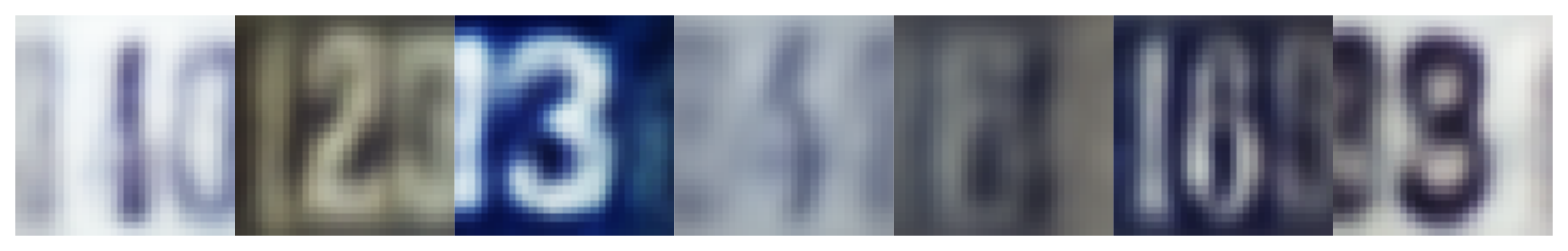}}
    \subfloat{\includegraphics[width=2.3in]{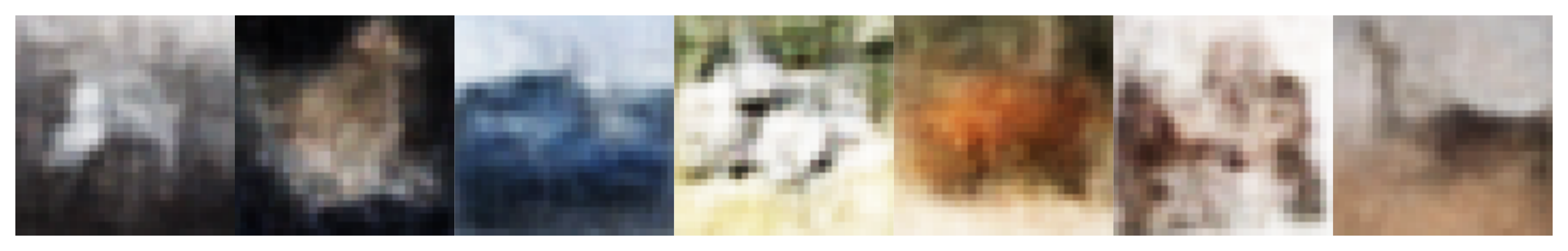}}\\
    \vspace{-5mm}
    \adjustbox{minipage=5em,raise=\dimexpr -3.\height}{\small RAE (SN)}
    \subfloat{\includegraphics[width=2.3in]{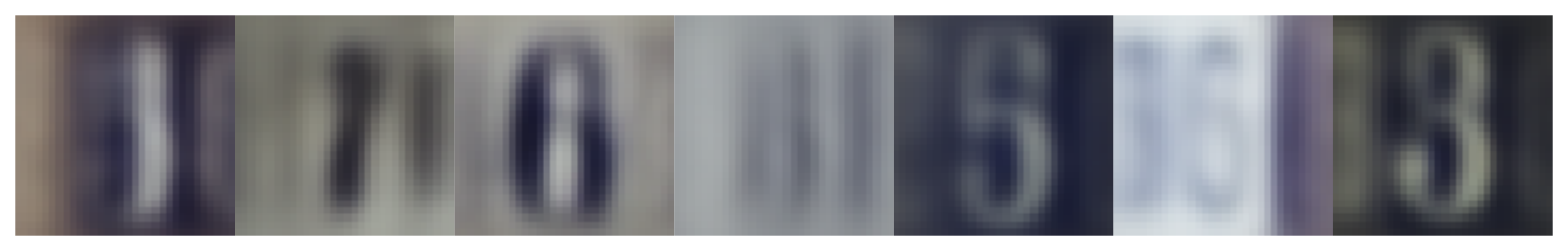}}
    \subfloat{\includegraphics[width=2.3in]{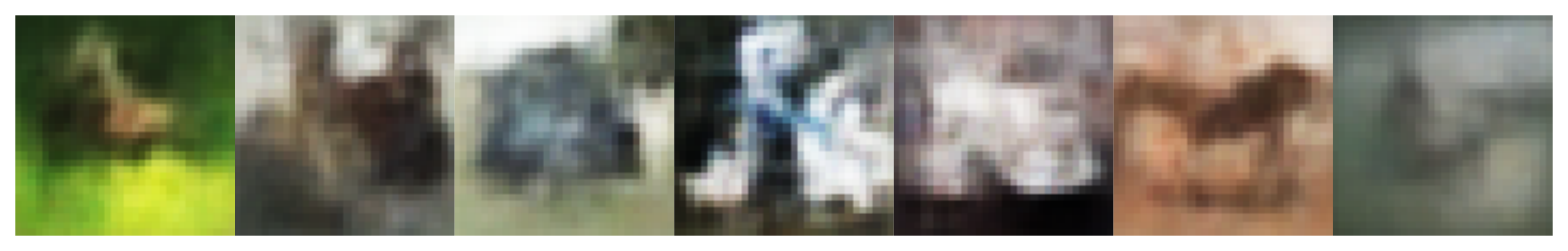}}\\
    \vspace{-5mm}
    \adjustbox{minipage=5em,raise=\dimexpr -3.\height}{\small RAE}
    \subfloat{\includegraphics[width=2.3in]{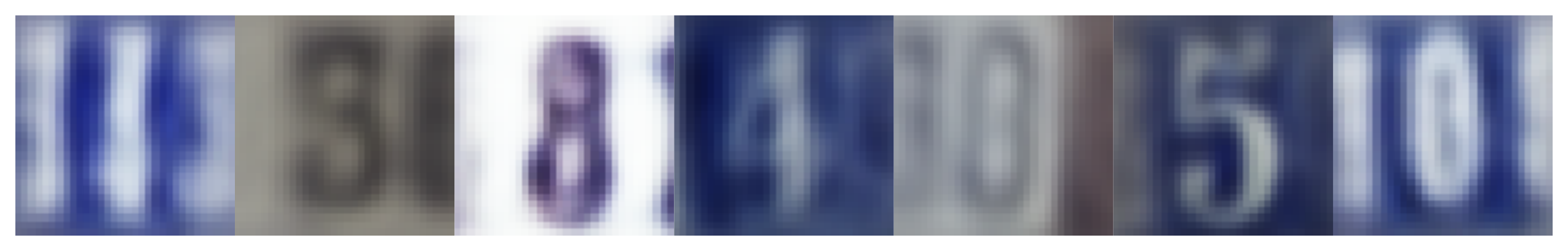}}
    \subfloat{\includegraphics[width=2.3in]{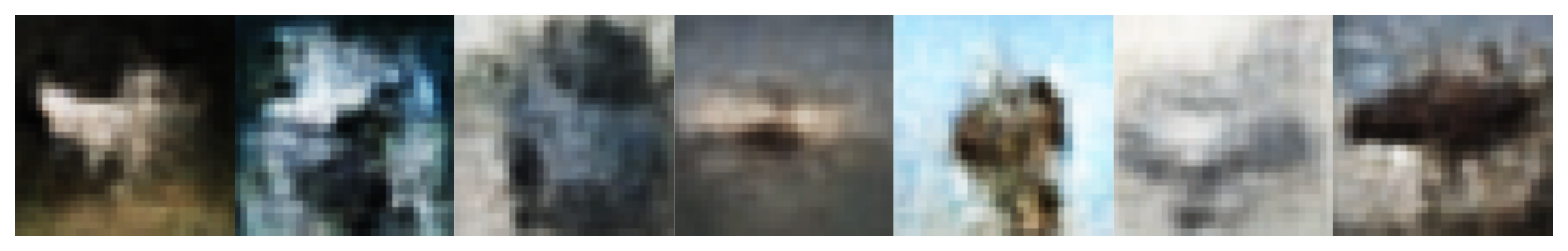}}\\
    \vspace{-5mm}
    \adjustbox{minipage=5em,raise=\dimexpr -3.\height}{\small VAE - Ours}
    \subfloat{\includegraphics[width=2.3in]{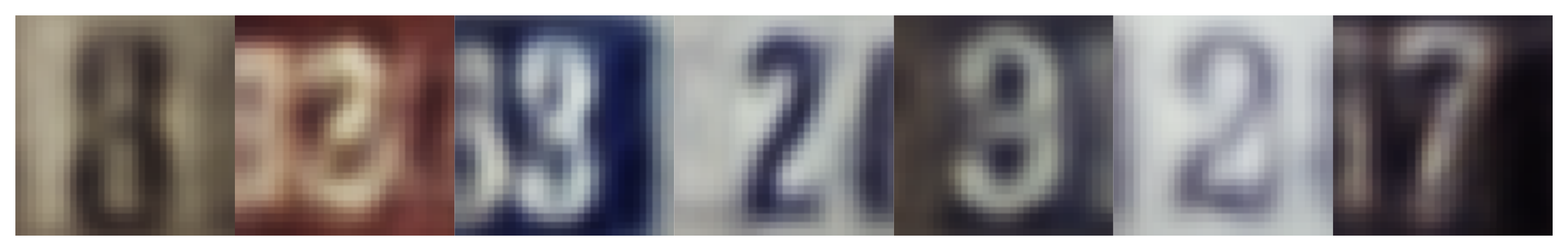}}
    \subfloat{\includegraphics[width=2.3in]{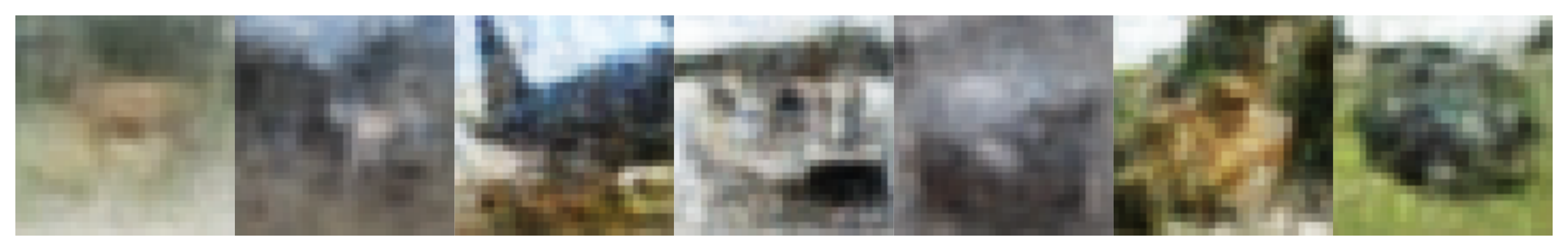}}

    \caption{Generated samples with different models and generation processes. }
    \label{Fig: Generated samples}
    \end{figure}

\begin{figure}[ht]
    \centering
    \captionsetup[subfigure]{position=above, labelformat = empty}
    \subfloat[Gen.\hspace{5.5mm}Near.]{\includegraphics[width=1in]{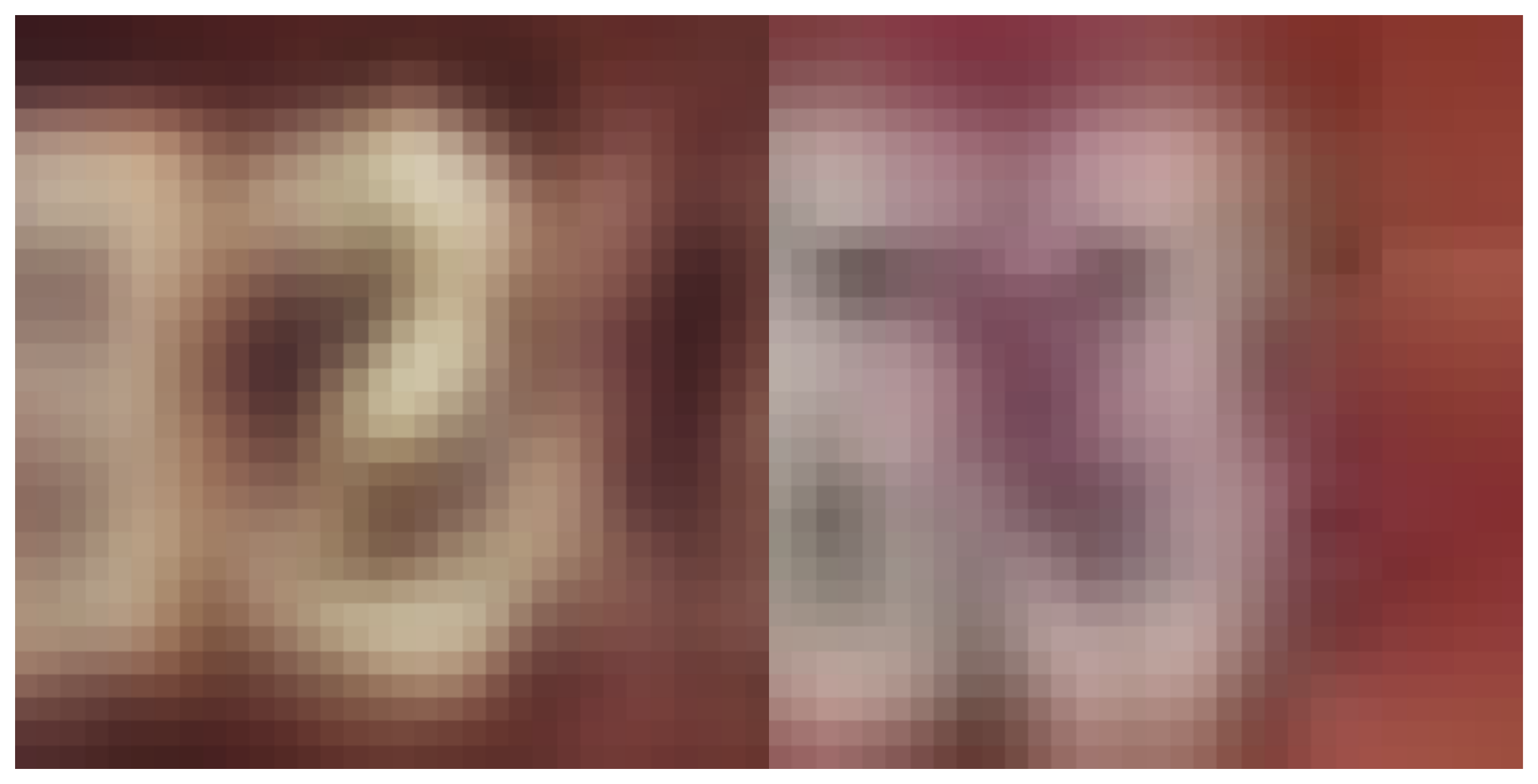}}
    \subfloat[Gen.\hspace{5.5mm}Near.]{\includegraphics[width=1in]{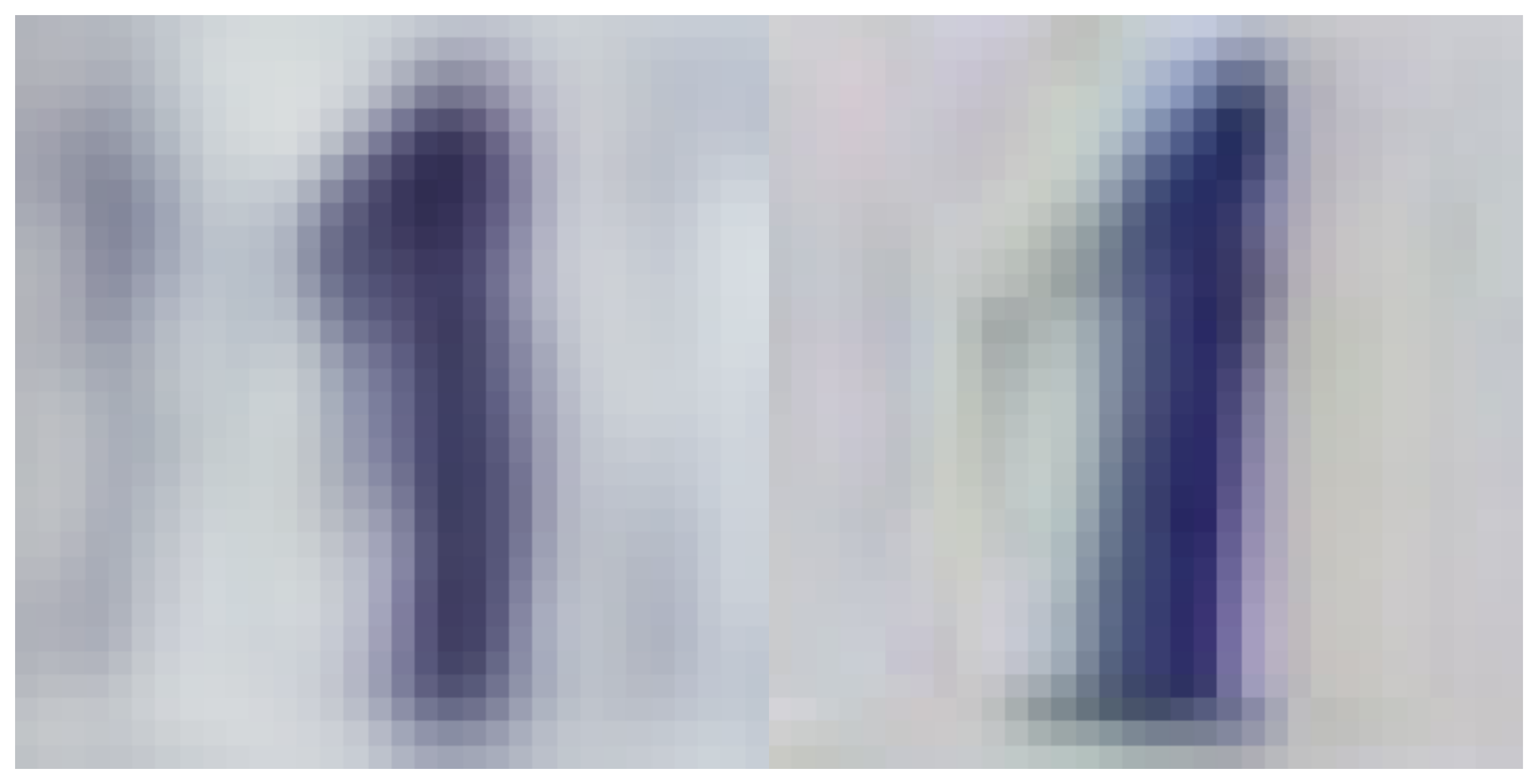}}
    \subfloat[Gen.\hspace{5.5mm}Near.]{\includegraphics[width=1in]{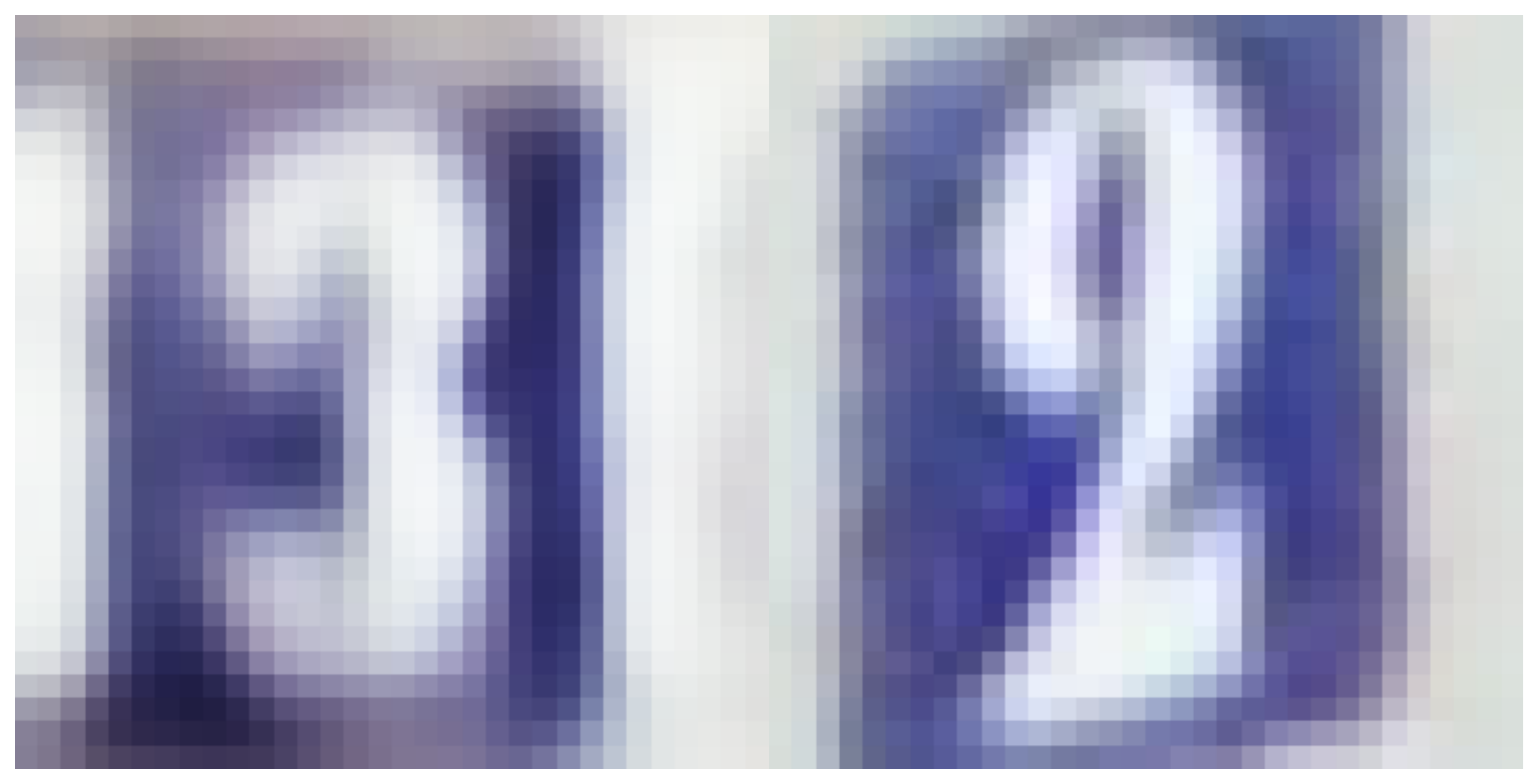}}
    \subfloat[Gen.\hspace{5.5mm}Near.]{\includegraphics[width=1in]{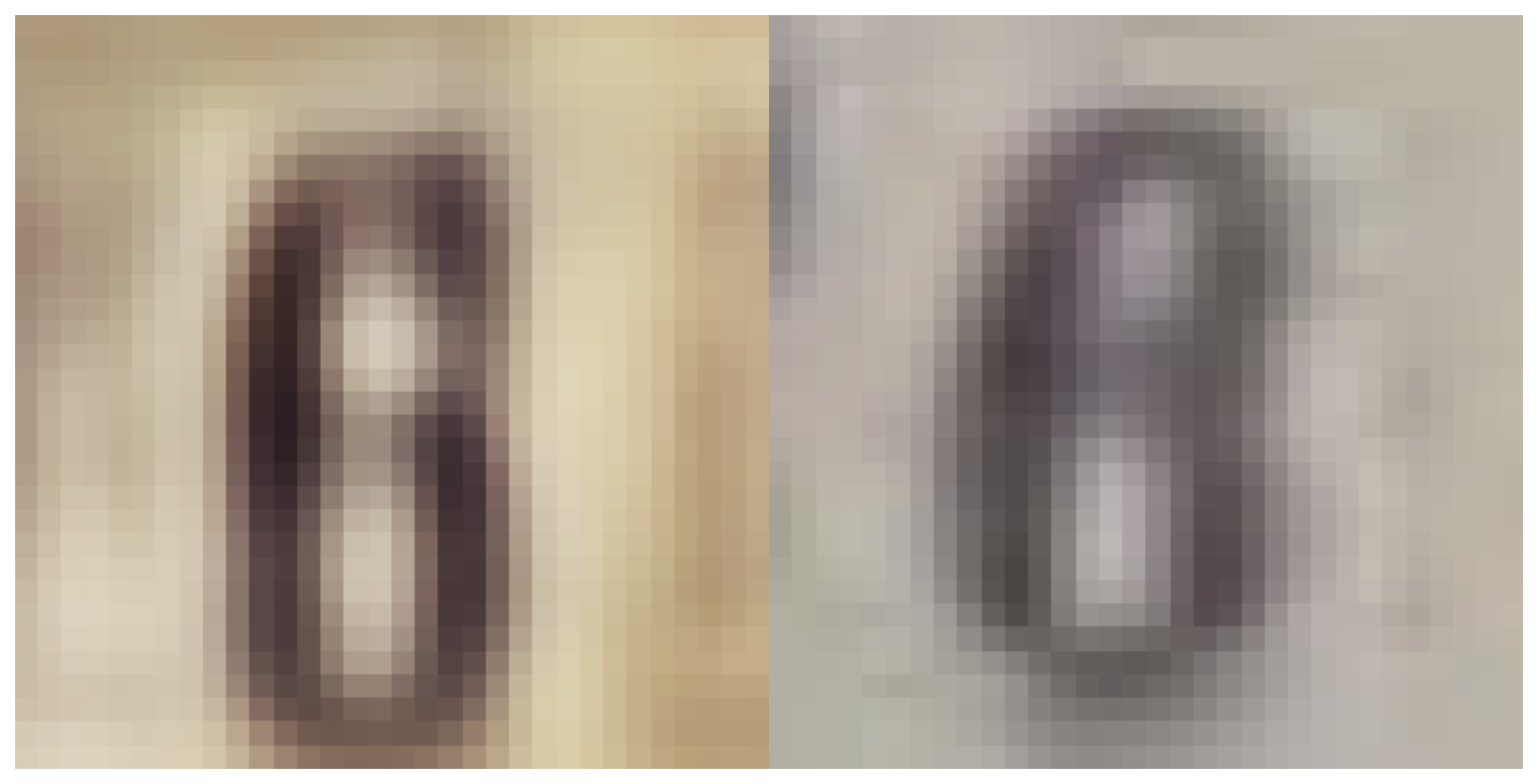}}
    \subfloat[Gen.\hspace{5.5mm}Near.]{\includegraphics[width=1in]{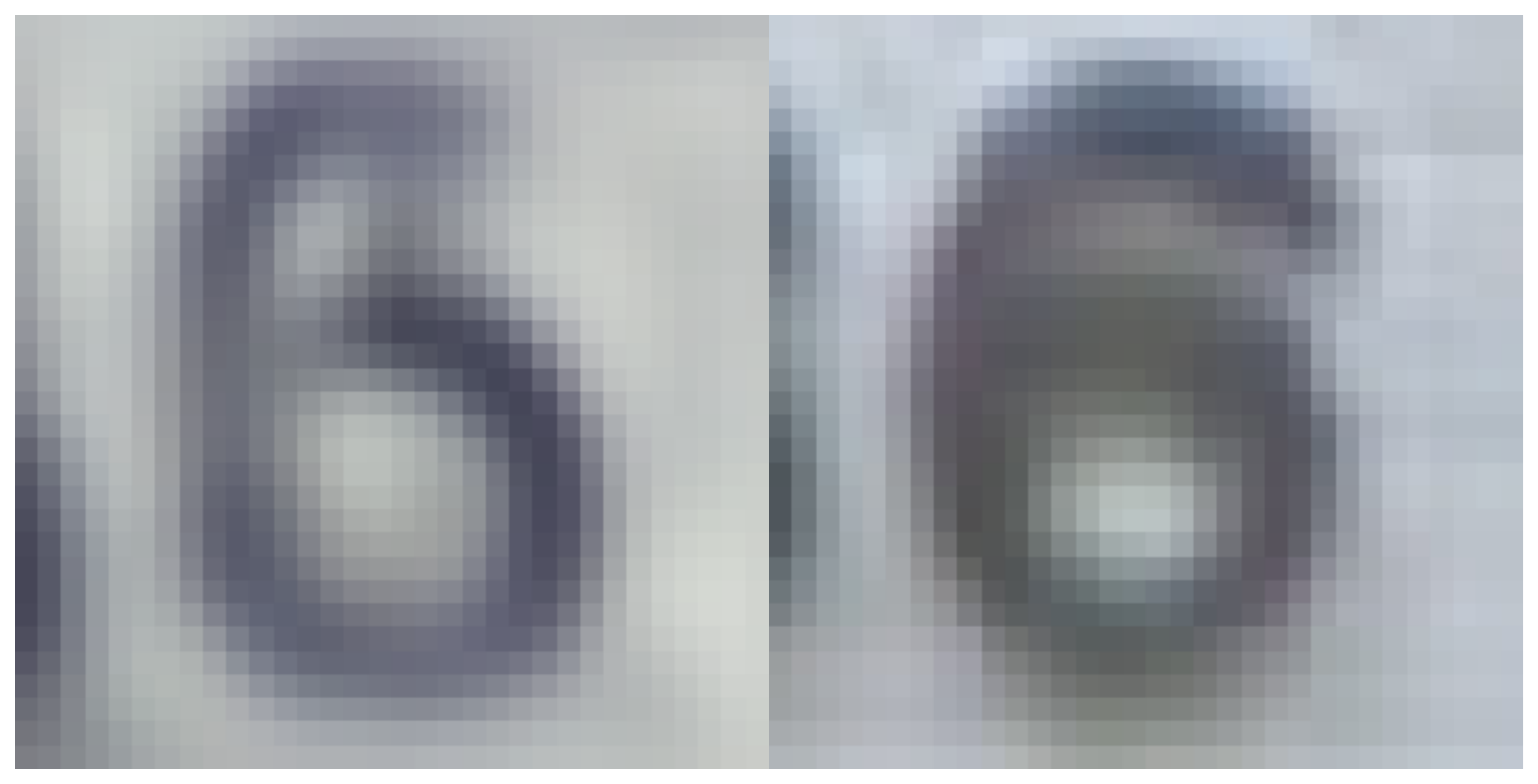}}
    \vfil
    \vspace{-4mm}
    \subfloat{\includegraphics[width=1in]{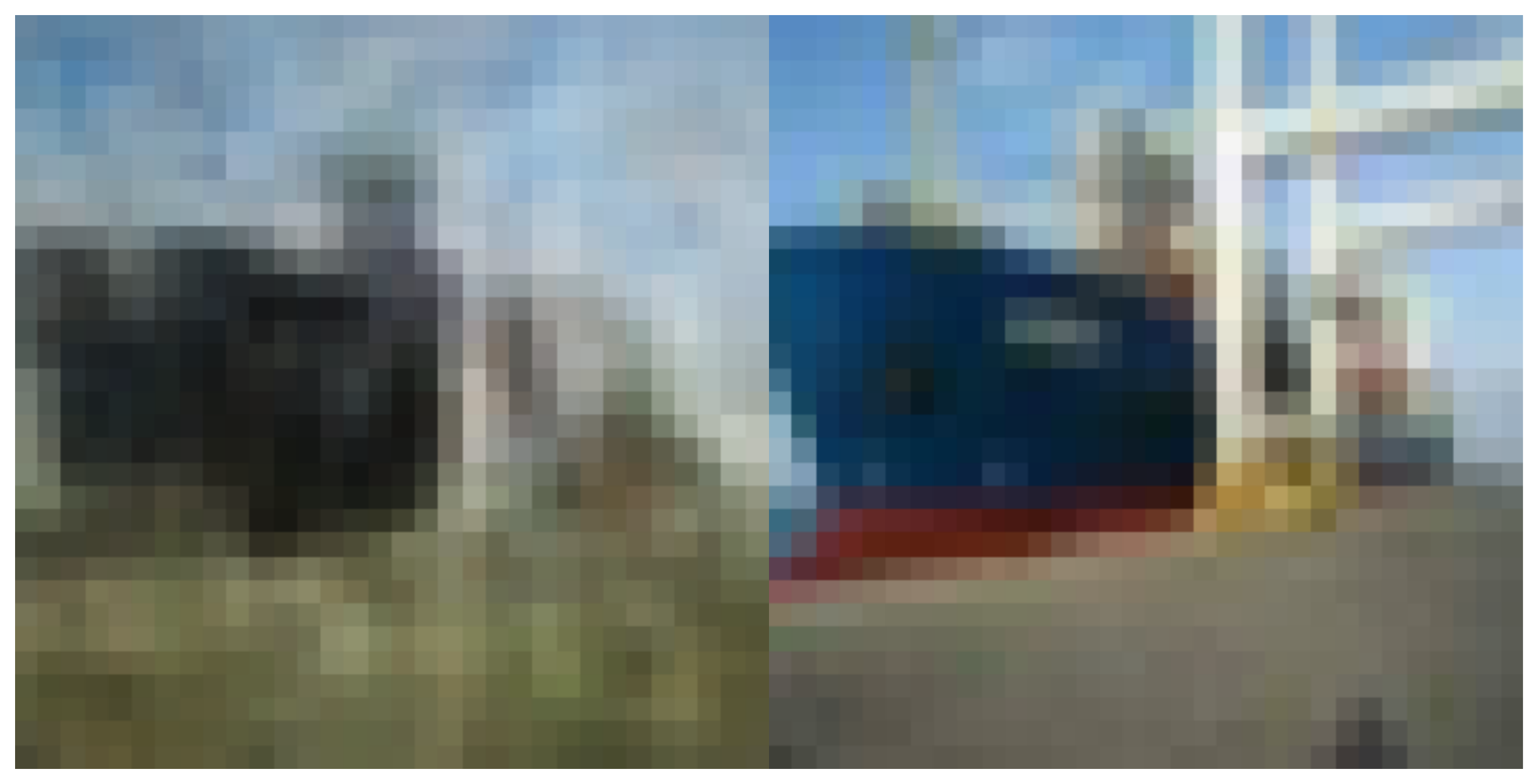}}
    \subfloat{\includegraphics[width=1in]{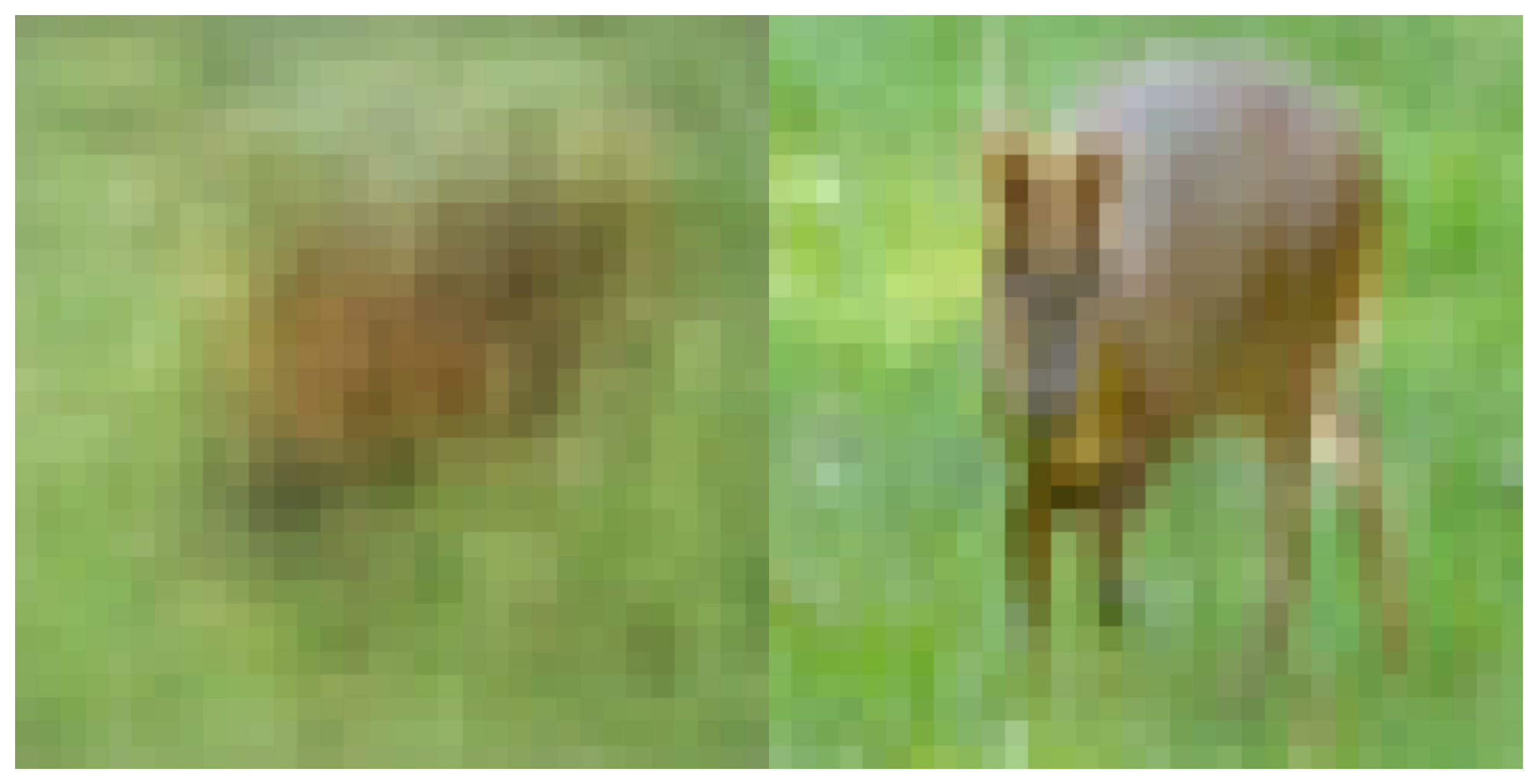}}
    \subfloat{\includegraphics[width=1in]{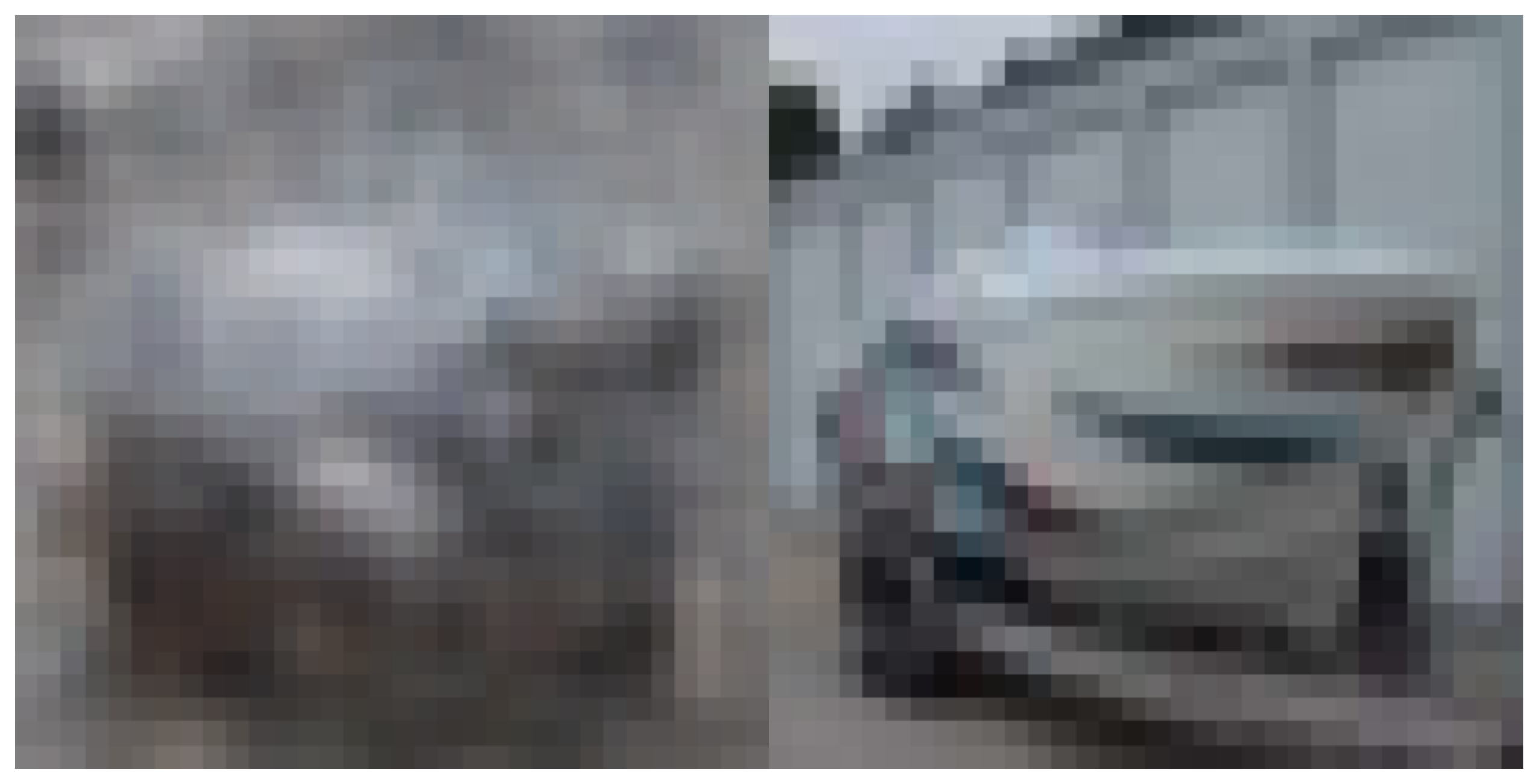}}
    \subfloat{\includegraphics[width=1in]{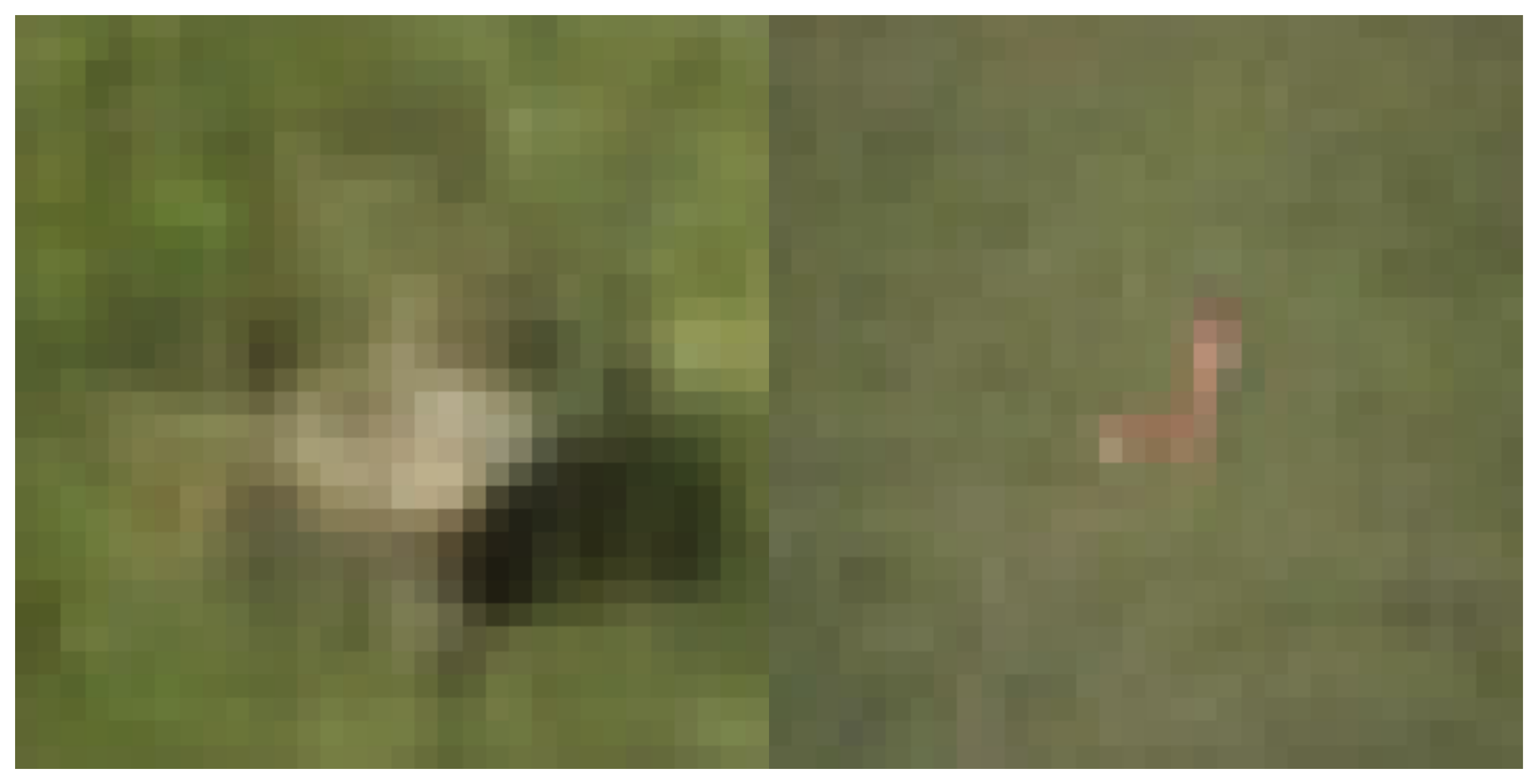}}
    \subfloat{\includegraphics[width=1in]{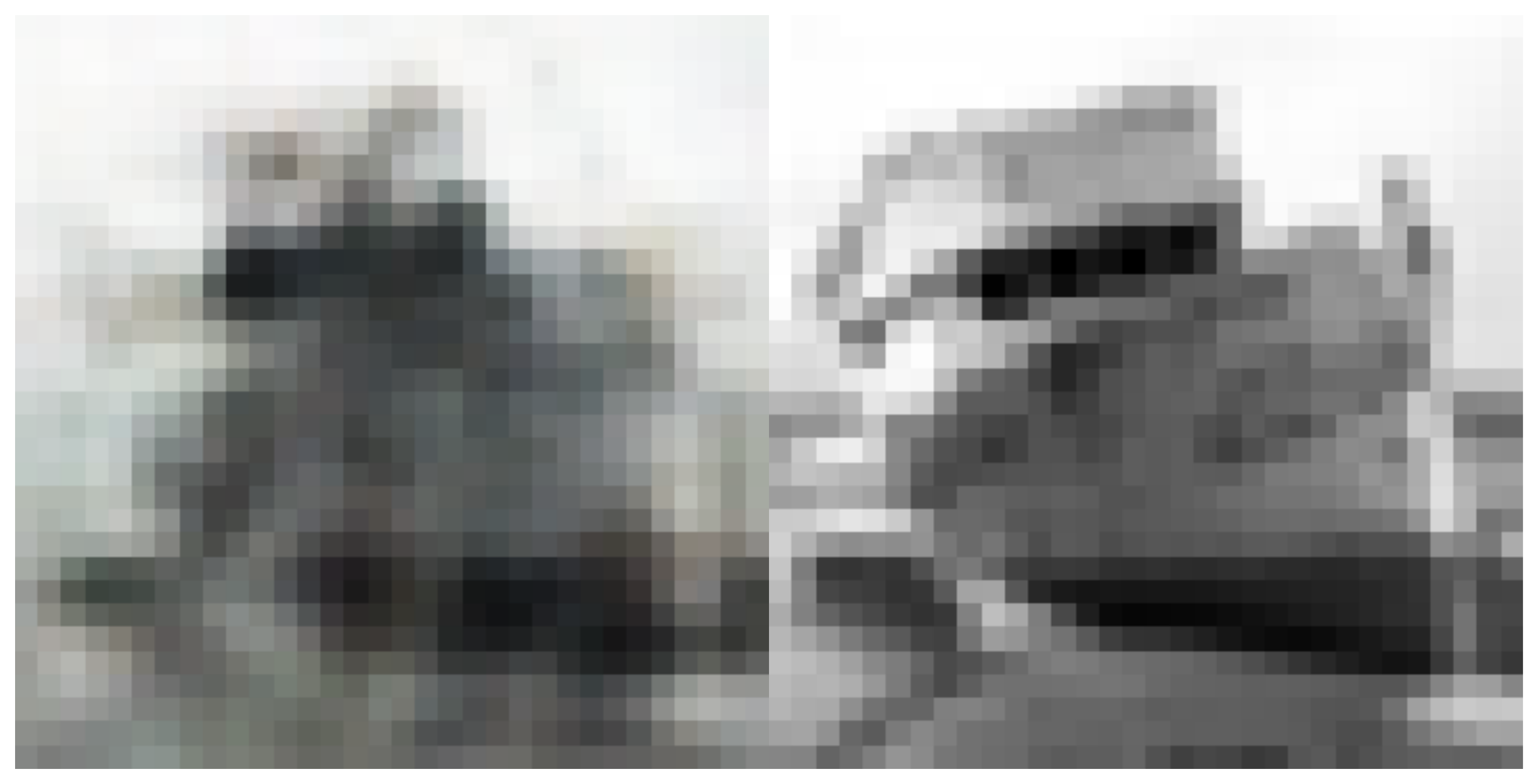}}
    \caption{Closest element in the training set (Near.) to the generated one (Gen.) with the proposed method.}
    \label{fig: overfitting bis}
    \end{figure}

\clearpage

\subsection{Generation with complex data}\label{appC3}
\label{Sec: genration with complex data}

\begin{figure}[ht]
    \centering
    \captionsetup[subfigure]{position=above, labelformat = empty}
    \adjustbox{minipage=5em,raise=\dimexpr -5.\height}{\small Train}
    \subfloat[OASIS]{\includegraphics[width=3.8in]{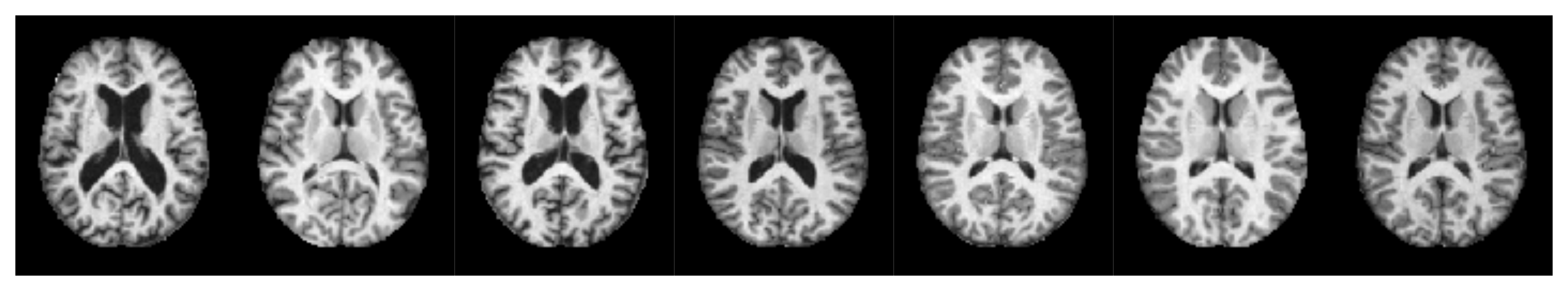}}\\
    \vspace{-5mm}
    \adjustbox{minipage=5em,raise=\dimexpr -5.\height}{\small  VAE - $\mathcal{N}$}
    \subfloat{\includegraphics[width=3.8in]{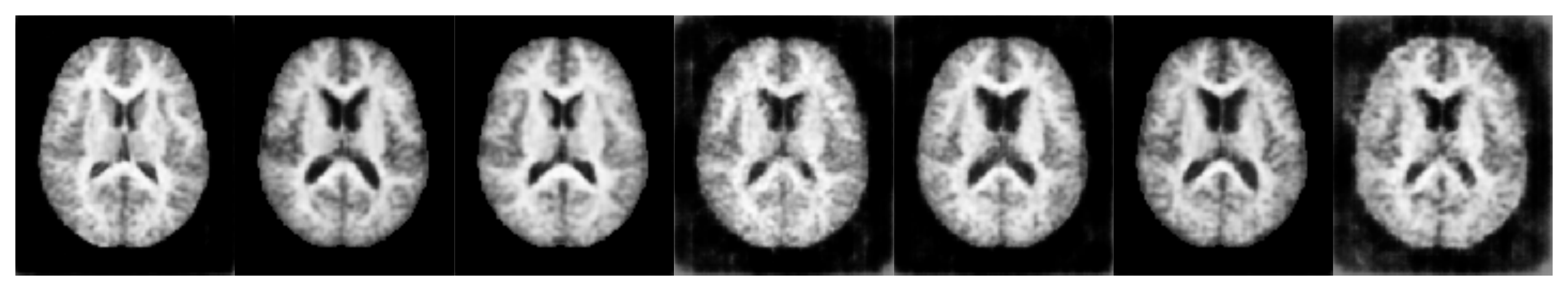}}\\
    \vspace{-5mm}
    \adjustbox{minipage=5em,raise=\dimexpr -5.\height}{\small WAE}
    \subfloat{\includegraphics[width=3.8in]{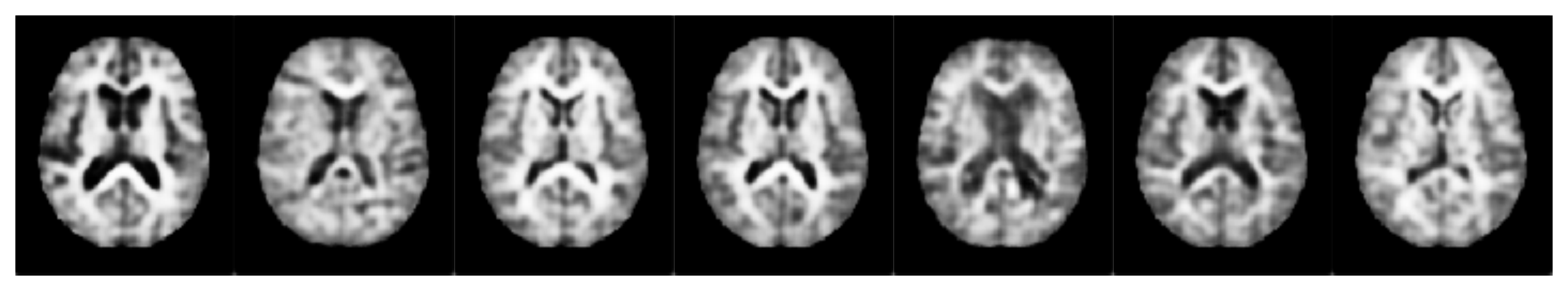}}\\
    \vspace{-5mm}
    \adjustbox{minipage=5em,raise=\dimexpr -5.\height}{\small VAMP}
    \subfloat{\includegraphics[width=3.8in]{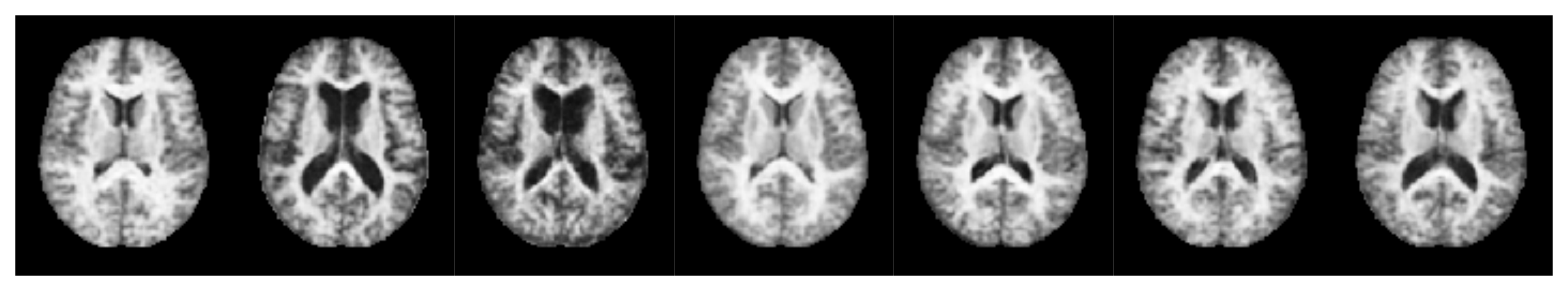}}\\
    \vspace{-5mm}
    \adjustbox{minipage=5em,raise=\dimexpr -5.\height}{\small HVAE}
    \subfloat{\includegraphics[width=3.8in]{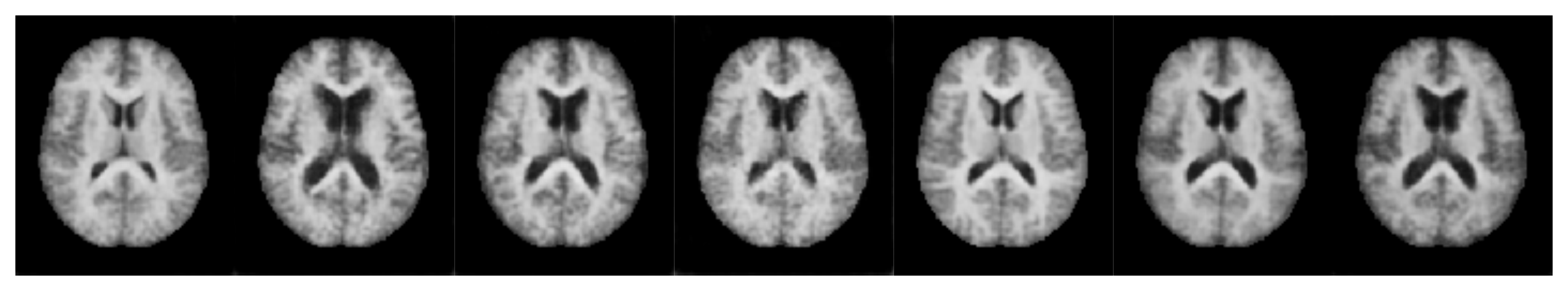}}\\
    \vspace{-5mm}
    \adjustbox{minipage=5em,raise=\dimexpr -5.\height}{\small RHVAE}
    \subfloat{\includegraphics[width=3.8in]{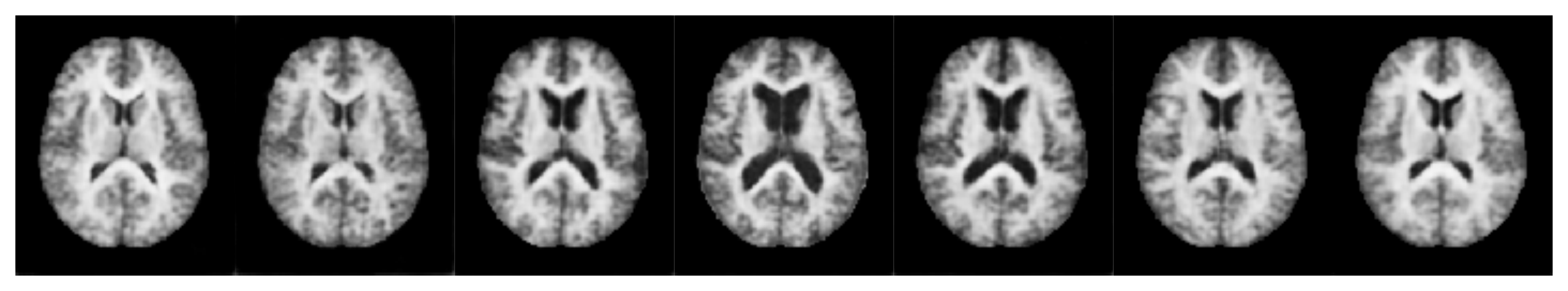}}\\
    \vspace{-5mm}
    \adjustbox{minipage=5em,raise=\dimexpr -5.\height}{\small VAE - GMM}
    \subfloat{\includegraphics[width=3.8in]{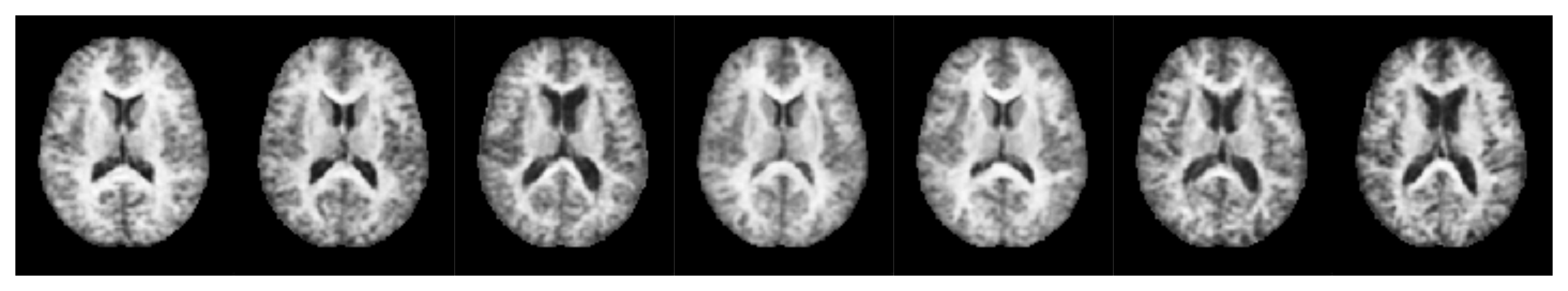}}\\
    \vspace{-5mm}
    \adjustbox{minipage=5em,raise=\dimexpr -5.\height}{\small RAE (GP)}
    \subfloat{\includegraphics[width=3.8in]{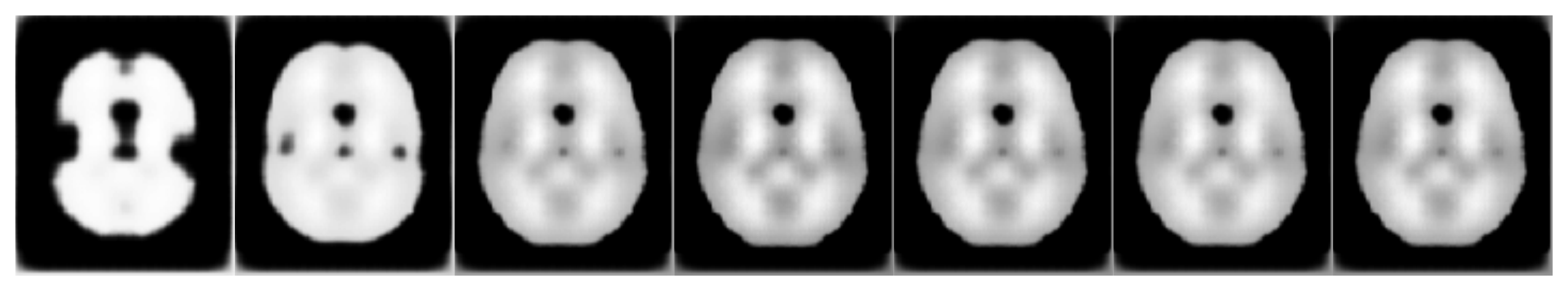}}\\
    \vspace{-5mm}
    \adjustbox{minipage=5em,raise=\dimexpr -5.\height}{\small RAE (L2)}
    \subfloat{\includegraphics[width=3.8in]{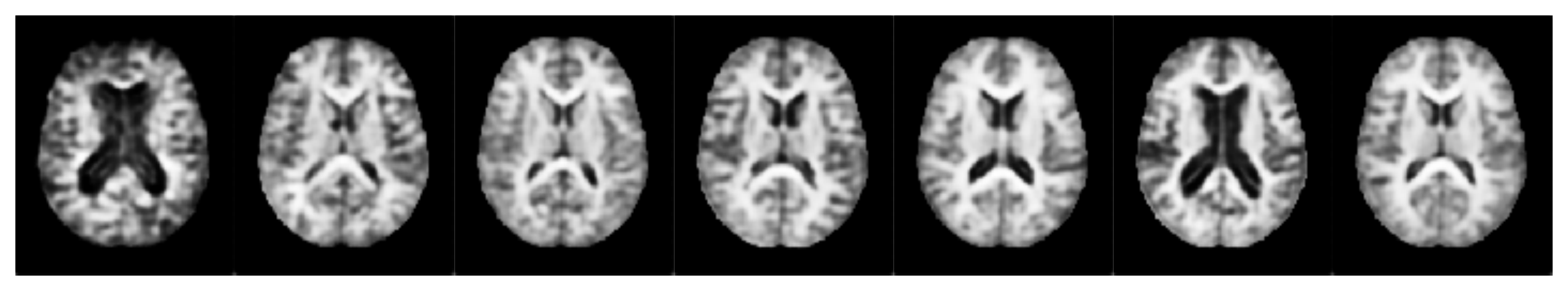}}\\
    \vspace{-5mm}
    \adjustbox{minipage=5em,raise=\dimexpr -5.\height}{\small RAE (SN)}
    \subfloat{\includegraphics[width=3.8in]{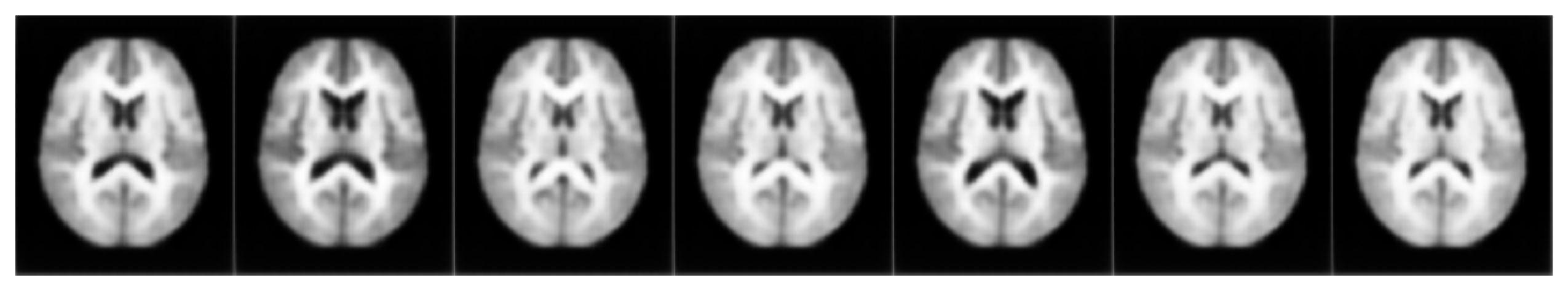}}\\
    \vspace{-5mm}
    \adjustbox{minipage=5em,raise=\dimexpr -5.\height}{\small RAE}
    \subfloat{\includegraphics[width=3.8in]{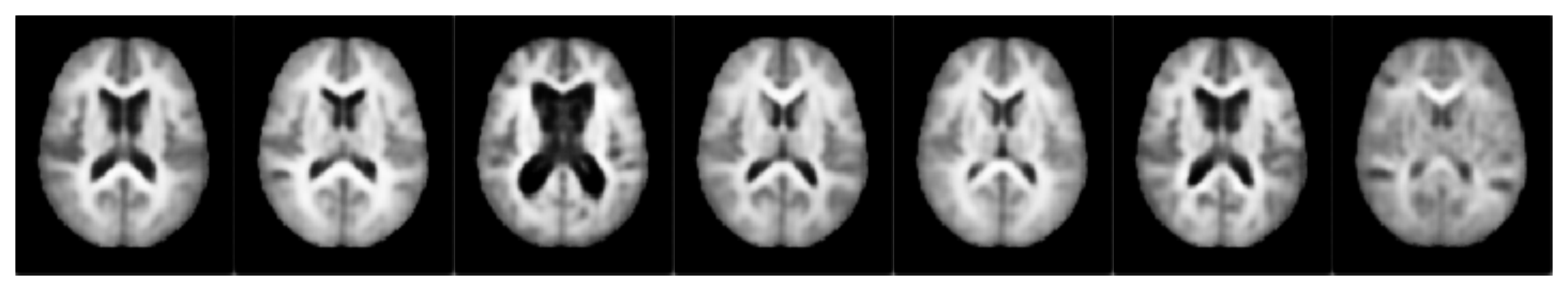}}\\
    \vspace{-5mm}
    \adjustbox{minipage=5em,raise=\dimexpr -5.\height}{\small VAE - Ours}
    \subfloat{\includegraphics[width=3.8in]{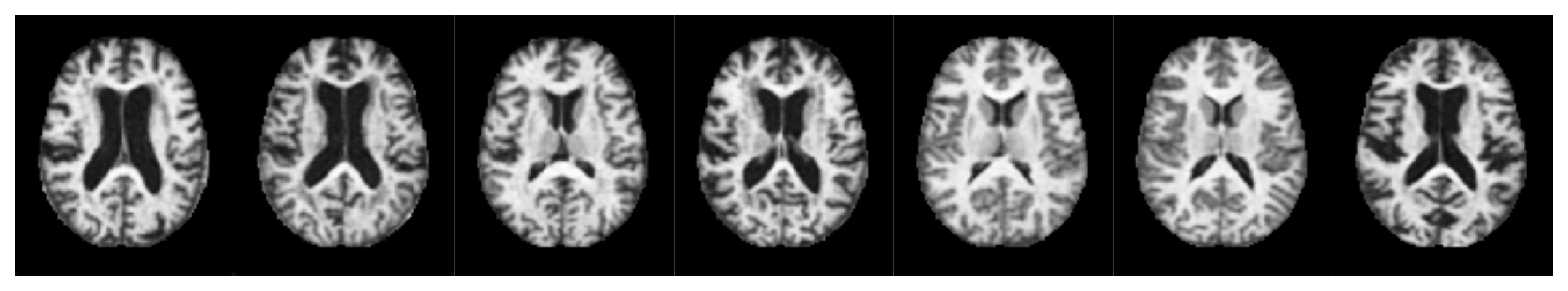}}
    
    \caption{Generated samples with different models and generation processes. }
    \label{fig: oasis generation}
    \end{figure}

Finally, we also propose to stress the proposed generation procedure in a day-to-day scenario where the limited data regime is more than common. To stress the model in such condition, we consider the publicly available OASIS database \cite{marcus_open_2007} composed of 416 MRI of patients, 100 of whom were diagnosed with Alzheimer disease (AD). Since both FID and PRD scores are not reliable when no large test set is available, we propose to assess quantitatively the generation quality with a data augmentation task. Hence, we split the dataset into a train (70\%), a validation (10\%) and a test set (20\%). Each model is trained on each label of the train set and used to generate 2k samples per class. Then a CNN classifier is trained on i) the original train set and ii) the 4k generated samples from the generative models and tested on the test set. Table~\ref{tab:classification results} shows classification results averaged across 20 runs for each considered model. The line \emph{raw (resampled)} corresponds to a case where the train set is obtained by balancing the classes with simple repetitions of the samples from the under-represented class. These metrics provide a way to assess i) if the model can generate data adding relevant information for classification and ii) allows to quantify the amount of overfitting. The proposed method is the only one allowing to achieve higher balanced accuracy and F1 scores for both labels than on the original (unbalanced) data meaning that the samples are relevant to the classifier and this is also sign of a good generalization. Moreover, we provide generated samples using each generation procedure in Figure~\ref{fig: oasis generation}. Again, the proposed method appears to produce visually the sharpest samples. However, such augmentation method for medical data requires caution and needs further assessment on the possibly induced biases before being used on a \emph{real-life} application case.

\begin{table}[ht]
    \caption{Classification results averaged on 20 independent runs. For the VAEs, the classifier is trained on 2K generated samples per class.}
    \vskip 0.15in
    \centering
    \scriptsize
    \begin{tabular}{l|ccccccc}
    \toprule
        \multirow{2}{*}{Generation method} & Balanced & \multicolumn{2}{c}{F1}&  \multicolumn{2}{c}{Precision} & \multicolumn{2}{c}{Recall}\\
        & Accuracy & AD & CN & AD & CN \\
        \midrule
        Original*           & 66.2 $\pm$ 7.6 & 47.6 $\pm$ 15.8 & 87.3 $\pm$ 2.0 & 74.7 $\pm$ 8.4 & 80.3 $\pm$ 4.0 & 35.7 $\pm$ 16.3 & 95.7 $\pm$ 1.5 \\
        Original (resampled) & 81.8 $\pm$ 2.6& 72.1 $\pm$ 3.6 & \textbf{88.0 $\boldsymbol{\pm}$ 2.3} & 67.0 $\pm$ 5.3 & 91.4 $\pm$ 1.8 & 78.5 $\pm$ 5.2 & 85.1 $\pm$ 4.2 \\
        \midrule
        AE - $\mathcal{N}$  & 50.0 $\pm$ 0.0 & 0.0 $\pm$ 0.0 & 84.1 $\pm$ 0.0 & 0.0 $\pm$ 0.0 & 72.6 $\pm$ 0.0 & 0.0 $\pm$ 0.0 & 100.0 $\pm$ 0.0 \\
        WAE                 & 57.4 $\pm$ 9.7 & 21.0$\pm$ 24.5 & 84.4 $\pm$ 2.3  & 48.5$\pm$ 42.8 & 76.7 $\pm$ 6.1 & 19.3 $\pm$ 27.5 & 95.4 $\pm$ 9.3 \\
        VAE - $\mathcal{N}$ & 51.8 $\pm$ 3.8 & 6.1 $\pm$ 11.8 & 84.6 $\pm$ 1.1 & 38.0 $\pm$ 47.3 & 73.4 $\pm$ 1.7 & 3.7 $\pm$ 7.8 & 99.8 $\pm$ 0.7 \\
        VAMP                & 83.1 $\pm$ 2.6 & 70.4 $\pm$ 3.6 & 82.2 $\pm$ 4.7 & 56.3 $\pm$ 5.2 & 97.5 $\pm$ 2.1 & 94.8 $\pm$ 4.7 & 71.5 $\pm$ 7.4 \\
        HVAE & 56.3 $\pm$ 7.9 & 19.6 $\pm$ 21.7& 85.4 $\pm$ 1.7 & 48.7 $\pm$ 41.7 & 75.5 $\pm$ 3.8 & 13.9 $\pm$ 17.6 & 98.6 $\pm$2.2\\
        RHVAE & 68.0 $\pm$ 10.9 & 47.0 $\pm$ 24.2& 85.1 $\pm$ 3.3 & 56.1 $\pm$ 25.3 & 83.0 $\pm$ 7.5 & 46.7 $\pm$ 30.2 & 89.2 $\pm$ 10.6\\
        \midrule
        AE - GMM  & 82.4 $\pm$ 2.3 & 69.5 $\pm$ 3.1 & 82.0 $\pm$ 3.6 & 55.8 $\pm$ 4.9  & 96.8 $\pm$ 2.4 & 93.3 $\pm$ 5.6 & 71.5 $\pm$ 6.2 \\
        RAE (GP)  & 63.9 $\pm$ 9.8 & 46.5 $\pm$ 15.9& 70.6 $\pm$ 19.6 & 45.3 $\pm$ 18.5 & 84.2 $\pm$ 8.6 & 60.9 $\pm$ 28.6& 67.0 $\pm$ 24.9 \\
        RAE (L2)  & 74.1 $\pm$ 6.0 & 60.6 $\pm$ 9.5 & 82.1 $\pm$ 5.9 & 57.8 $\pm$ 10.1 & 88.3 $\pm$ 5.2 & 70.0 $\pm$ 18.7& 78.3 $\pm$ 11.7 \\
        RAE (SN)  & 62.3 $\pm$ 8.9 & 37.8 $\pm$ 22.6& 80.1 $\pm$ 7.9 & 43.1 $\pm$ 24.9 & 80.6 $\pm$ 6.6 & 41.7 $\pm$ 30.1& 82.9 $\pm$ 16.4 \\
        RAE       & 69.3 $\pm$ 8.1 & 53.8 $\pm$ 12.9& 80.0 $\pm$ 10.7 & 56.2 $\pm$ 13.5 & 85.2 $\pm$ 6.2 & 60.0 $\pm$ 24.0& 78.5 $\pm$ 17.5 \\
        VAE - GMM & 83.0 $\pm$ 3.6 & 71.4 $\pm$ 4.3 & 85.3 $\pm$ 3.0 & 60.7 $\pm$ 5.4  & 94.9 $\pm$ 3.7 & 88.0 $\pm$ 9.5 & 77.9 $\pm$ 5.9 \\
        \midrule
        VAE - Ours & \textbf{85.4 $\boldsymbol{\pm}$ 2.5} & \textbf{74.7 $\boldsymbol{\pm}$ 3.5}  &  87.3 $\pm$ 2.7 & 64.0 $\pm$ 5.3 & 95.8 $\pm$ 2.2 & 90.4 $\pm$ 5.6 & 80.3 $\pm$ 5.1 \\
        \bottomrule
        \multicolumn{6}{l}{\small *unbalanced}
        
    \end{tabular}

    \label{tab:classification results}
\end{table}

\subsection{Wider hyper-parameter search}

As stated in the paper, for the experiments, we used the official implementation and hyper-parameters provided by the authors when available. However, we also propose to perform a hyper-parameter search for the models considered in the benchmark \emph{i.e.} WAE, VAMP-VAE, RAE-GP and RAE-L2 [3] on MNIST and CELEBA. Since both HVAE and RHVAE models have a very time consuming training, we propose to replace these approaches with models having the same objective (i.e. enriching the posterior distribution). Do to so we consider a VAE with inverse autoregressive flows \citep{kingma2016improved} (VAE-IAF) and a VAE with normalizing flows with radial/planar invertible transformations \citep{rezende_variational_2015} (VAE-NF).
          
We train these models with 10 different hyper-parameter configurations on MNIST and CELEBA. For the WAE, we vary the kernel bandwidth in \{0.01, 0.1, 0.5, 1, 2, 5\} and change the regularization factor weighting the reconstruction and regularization in \{0.01, 0.1, 1, 10, 100\}. For the RAEs, we vary the L2 latent code regularization factor and the factor before the explicit regularization in \{$1e^{-6}$, $1e^{-4}$, $1e^{-3}, 0.01, 0.1, 1\}$. For the VAMP we vary the number of pseudo-inputs in \{10, 20, 30, 50, 100, 150, 200, 250, 300,189
500\}. Finally, for the flow-based VAEs we vary the complexity of the flows with different number of IAF blocks (VAE-IAF) or different flow lengths (VAE-NF). To assess the influence of the neural architecture, the experiment is performed twice each time with a different neural network architecture (CNN in Table.~\ref{Tab: CNN Architectures} or a simpler ResNet). In Table.~\ref{tab:hp search}, we show the generation vs. test FID of the model achieving the lowest FID on the validation set.

\begin{table}[ht]
    \caption{FID (lower is better) for different models and datasets. For the mixture of Gaussian (GMM), we fit a 10-component mixture of Gaussian in the latent space.}
    \vskip 0.15in
    \centering
    \scriptsize
    \begin{sc}
    \begin{tabular}{l|cccc}
    \toprule
    Models                & \multicolumn{2}{c}{MNIST}  & \multicolumn{2}{c}{CELEBA} \\
    \midrule
    Nets         	& CNN    	& ResNet       & CNN		      	& ResNet      	\\
    \midrule
AE - N(0,1)  	& 46.4    	& 221.8     & 64.6     	 	    & 275.0    	\\
WAE          	& 18.9	  	& 20.3      & 54.6     	  	    & 67.1     	\\
VAE - N(0,1) 	& 40.7    	& 47.8      & 64.1     	  	    & 69.5     	\\
VAMP         	& 34.0    	& 34.5      & 56.0     	  	    & 67.2     	\\
VAE-NF       	& 29.3    	& 32.5      & 55.4     	  	    & 67.1     	\\
VAE-IAF      	& 27.5    	& 30.6      & 56.5     	  	    & 66.2     	\\
\midrule
AE - GMM     	& 9.6     	& 11.0      & 56.1     	  	    & 57.4     	\\
RAE-GP       	& 9.4     	& 11.4      & 52.5	   	  	    & 59.0     	\\
RAE-L2       	& 9.1	   	& 11.5      & 54.5     	  	    & 58.3     	\\
VAE - GMM    	& 13.1    	& 12.4      & 55.5     	  	    & 59.9     	\\
\midrule
Ours         	& \textbf{8.5}		& \textbf{10.7 }     & \textbf{48.7	} 			& \textbf{53.2}		\\
    \bottomrule

    \end{tabular}
    \end{sc}
    \label{tab:hp search}
\end{table}

\clearpage    
\section{Experimental set-up}\label{appD}
We compare the proposed sampling method to several VAE variants such as a Wasserstein Autoencoder (WAE) \cite{tolstikhin2018wasserstein}, Regularized Autoencoders \cite{ghosh_variational_2020} with either L2 decoder's parameters regularization (RAE-L2), gradient penalty (RAE-GP), spectral normalization (RAE-SN) or simple L2 latent code regularization (RAE), a vamp-prior VAE (VAMP) \cite{tomczak_vae_2018}, a Hamiltonian VAE (HVAE) \cite{caterini_hamiltonian_2018}, a geometry-aware VAE (RHVAE) \cite{chadebec_geometry-aware_2020} and an Autoencoder (AE). The RAEs, VAEs and AEs are trained for 100 epochs for SVHN, MNIST\footnote{MNIST images are re-scaled to 32x32 images with a 0 padding.} and CELEBA and 200 on CIFAR10. Each time we use the official train and test split of the data. For MNIST and SVHN, 10k samples out of the train set are reserved for validation and 40k for CIFAR10. As to CELEBA, we use the official validation set for validation. The model that is kept at the end of training is the one achieving the best validation loss. All the models are trained with a batch size of 100 and starting learning rate of $1e{-3}$ (but CIFAR where the learning rate is set to $5e{-4}$) with an Adam optimizer \cite{kingma_adam_2014}. We also use a scheduler decreasing the learning rate by half if the validation loss stops increasing for 5 epochs. For the experiments on the sensitivity to the training set size, we keep the same set-up. For each dataset we ensure that the validation set is $1/5^{\text{th}}$ the size of the train set but for CIFAR where we select the best model on the train set. The neural networks architectures can be found in Table~\ref{Tab: CNN Architectures} and are inspired by \cite{ghosh_variational_2020}. The metrics (FID and PRD scores) are computed with 10000 samples against the test set (for CELEBA we selected only the 10000 first samples of the official test set). The factor $\rho$ is set to $\rho = \max \limits _i \min \limits _{j\neq i} \Vert c_i -c_j\Vert_2$ to ensure some \emph{smoothness} of the manifold. For models coming from peers, we use the parameters and code provided by the authors when available and allowed by licenses.

For the data augmentation task, the generative models are trained on each class for 1000 epochs with a batch size of 100 and a starting learning rate of $1e{-4}$. Again a scheduler is used and the learning rate is cut by half if the loss does not improve for 20 epochs. All the models have the autoencoding architecture described in Table~\ref{Tab: CNN Architectures}. As to the classifier, it is trained with a batch size of 200 for 50 epochs with a starting learning rate of $1e{-4}$ and Adam optimizer. A scheduler reducing the learning rate by half every 5 epochs if the validation loss does not improve is again used. The best kept model is the one achieving the best balanced accuracy on the validation set. Its neural network architecture may be found in Table~\ref{Tab: Classifier Architecture}. MRIs are only pre-processed such that the maximum value of a voxel is 1 and the minimum 0 for each data point.

\begin{table}[!ht]
\caption{Neural networks used for the encoder and decoders of VAEs in the benchmarks}
\vskip 0.15in
\label{Tab: CNN Architectures}
    \begin{center}
    \tiny
    \begin{sc}
    \begin{tabular}{ccccc}
    \toprule
         & MNIST [CIFAR10] & SVHN & CELEBA & OASIS\\
    \midrule
    \midrule
    Encoder & (1[3], 32, 32) & (3, 32, 32) & (3, 64, 64) & (1, 208, 176)\\
    \midrule
    \multirow{3}{*}{Layer 1} & Conv(128, (4, 4), stride=2)          & Linear(1000) & Conv(128, (5, 5), stride=2) & Conv(64, (5, 5), stride=2)\\
                             & Batch normalization                  & Relu         & Batch normalization & Relu \\
                             & Relu                                 &              & Relu\\
    \midrule
    \multirow{3}{*}{Layer 2} & Conv(256, (4, 4), stride=2)          & Linear(500)  & Conv(256, (5, 5), stride=2) & Conv(128, (5, 5), stride=2)\\
                             & Batch normalization                  & Relu         & Batch normalization & Relu   \\
                             & Relu                                 &              & Relu \\
    \midrule
    \multirow{3}{*}{Layer 3} & Conv(512, (4, 4), stride=2) &\multirow{3}{*}{Linear(500, 16)} & Conv(512, (5, 5), stride=2) & Conv(256, (5, 5), stride=2) \\
                             & Batch normalization         &                                  & Batch normalization & Relu \\
                             & Relu                        &                                  & Relu \\
    \midrule
    \multirow{3}{*}{Layer 4} & Conv(1024, (4, 4), stride=2) &\multirow{3}{*}{-}   & Conv(1024, (5, 5), stride=2) & Conv(512, (5, 5), stride=2)\\
                             & Batch normalization          &                     & Batch normalization & Relu\\
                             & Relu                         &                     & Relu \\
    \midrule
    \multirow{2}{*}{Layer 5}                  & \multirow{2}{*}{Linear(4096, 16)}            &  \multirow{2}{*}{-}                  & \multirow{2}{*}{Linear(16384, 64)} & Conv(1024, (5, 5), stride=2)\\
                    & & & & Relu \\
            
    \midrule 
    Layer 6                  & - &- &- & Linear(4096, 16) \\
    \midrule
    \midrule
    Decoder                  & (16 [32])                              & (16)      & (64) & (16) \\
    \midrule
    \multirow{2}{*}{Layer 1} & Linear(65536)                        & Linear(500) & Linear(65536) & Linear(65536) \\
                             & Reshape(1024, 8, 8)                  & Relu        & Reshape(1024, 8, 8) & Reshape(1024, 8, 8)        \\
    \midrule 
    \multirow{3}{*}{Layer 2} & ConvT(512, (4, 4), stride=2)         & Linear (1000) & ConvT(512, (5, 5), stride=2) & ConvT(512, (5, 5), stride=(3, 2))\\
                             & Batch normalization                  & Relu          & Batch normalization & Relu\\
                             & Relu                                 &               & Relu\\
    \midrule
    \multirow{3}{*}{Layer 3} & ConvT(256, (4, 4), stride=2)         & Linear(3072)       & ConvT(256, (5, 5), stride=2) & ConvT(256, (5, 5), stride=2)\\
                             & Batch normalization                  & Reshape(3, 32, 32) & Batch normalization & Relu \\
                             & Relu                                 & Sigmoid            & Relu \\
    \midrule
    \multirow{3}{*}{Layer 4} & ConvT(3, (4, 4), stride=1)         & \multirow{3}{*}{-}   & ConvT(128, (5, 5), stride=2) & ConvT(128, (5, 5), stride=2)\\
                             & Batch normalization                &                      & Batch normalization & Relu \\
                             & Sigmoid                            &                      & Relu \\
    \midrule
    \multirow{3}{*}{Layer 5} & \multirow{3}{*}{-}                 & \multirow{3}{*}{-}   & ConvT(3, (5, 5), stride=1) & ConvT(64, (5, 5), stride=2)\\
                             &                                    &                      & Batch normalization & Relu\\
                             &                                    &                      & Sigmoid\\
    \midrule
    \multirow{2}{*}{Layer 6} & \multirow{2}{*}{-} & \multirow{2}{*}{-} & \multirow{2}{*}{-} & ConvT(1, (5, 5), stride=1)\\
        & & & & Relu\\
    \bottomrule
    \end{tabular}
\end{sc}
\end{center}
\end{table}

\begin{table}[t]
\caption{Neural Network used for the classifier in Sec.~\ref{Sec: genration with complex data}}
\vskip 0.15in
\label{Tab: Classifier Architecture}
\scriptsize
    \begin{center}
    \begin{sc}
    \begin{tabular}{cc}
    \toprule
         & OASIS Classifier\\
    \midrule
    \midrule
     Input Shape & (1, 208, 176)\\
    \midrule
    \multirow{4}{*}{Layer 1} & Conv(8, (3, 3), stride=1) \\
    & Batch normalization \\
    & LeakyRelu \\
    & MaxPool(2, stride=2) \\
    \midrule
    \multirow{4}{*}{Layer 2} & Conv(16, (3, 3), stride=1) \\
    & Batch normalization \\
    & LeakyRelu \\
    & MaxPool(2, stride=2) \\
    \midrule
    \multirow{4}{*}{Layer 3} & Conv(32, (3, 3), stride=2) \\
    & Batch normalization \\
    & LeakyRelu \\
    & MaxPool(2, stride=2) \\
    \midrule
    \multirow{4}{*}{Layer 4} & Conv(64, (3, 3), stride=2) \\
    & Batch normalization \\
    & LeakyRelu \\
    & MaxPool(2, stride=2) \\
    \midrule
    \multirow{2}{*}{Layer 5} & Linear(256, 100) \\
    & Relu \\
    \midrule
    \multirow{2}{*}{Layer 6} & Linear(100, 2) \\
    & Softmax \\
    
    \bottomrule
    \end{tabular}
\end{sc}
\end{center}
\end{table}

\clearpage
\section{Dataset size sensibility on SVHN}\label{appE}

\begin{figure}[ht]
    \centering
    \captionsetup[subfigure]{position=above, labelformat = empty}
    \subfloat[SVHN]{\includegraphics[width=4in]{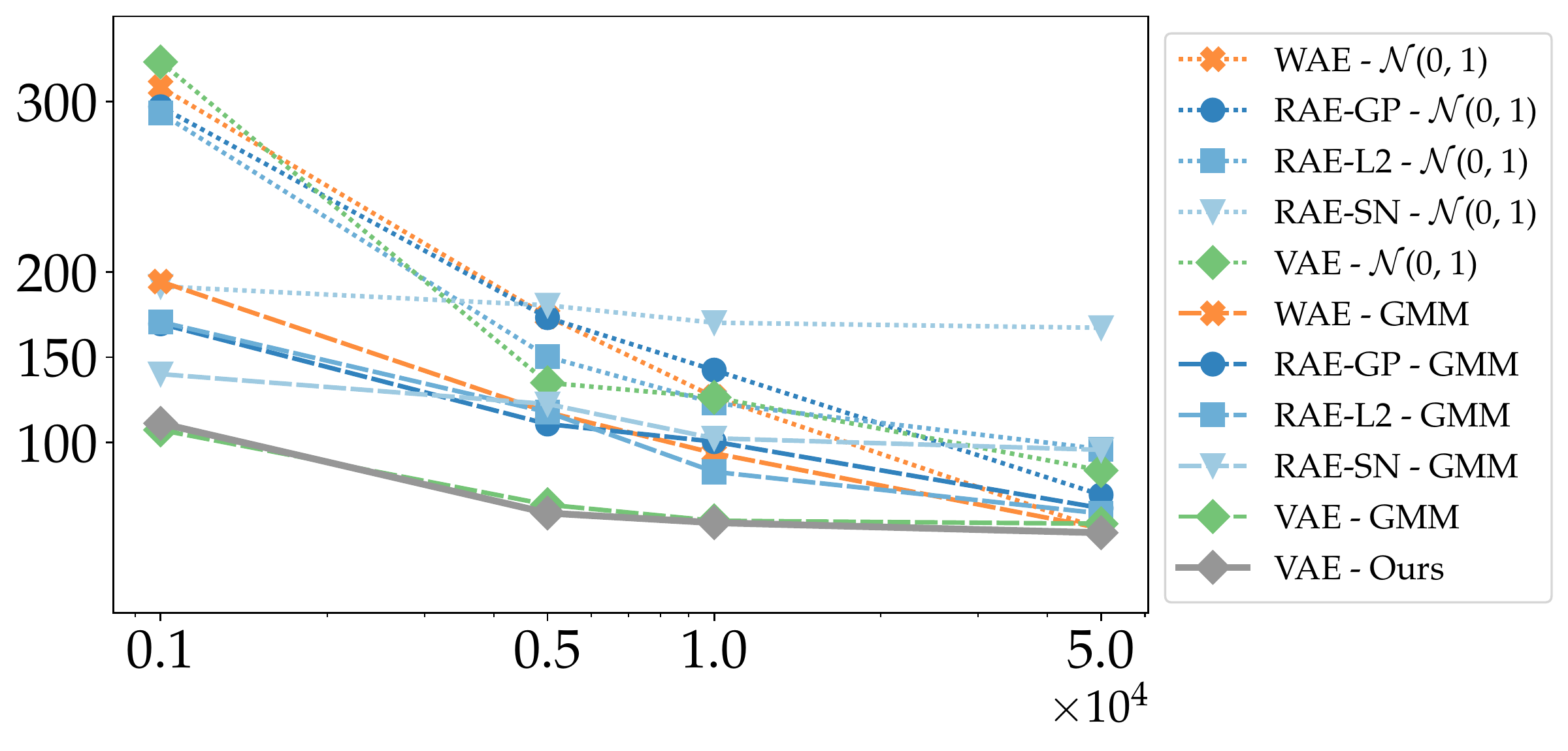}}
    \caption{FID score evolution according to the number of training samples.}
    \label{fig:fid evolution svhn}
    \end{figure}

In Figure~\ref{fig:fid evolution svhn}, we show the same plot for SVHN as in Sec.~5.2. Again the proposed method appears to be part of the most robust generation procedures to dataset size changes.

\clearpage
\section{Ablation study}\label{appF}

\subsection{Influence of the number of centroids in the metric}

\begin{figure}[ht]
    \centering
    \captionsetup[subfigure]{position=above, labelformat = empty}
    \subfloat{\includegraphics[width=2.7in]{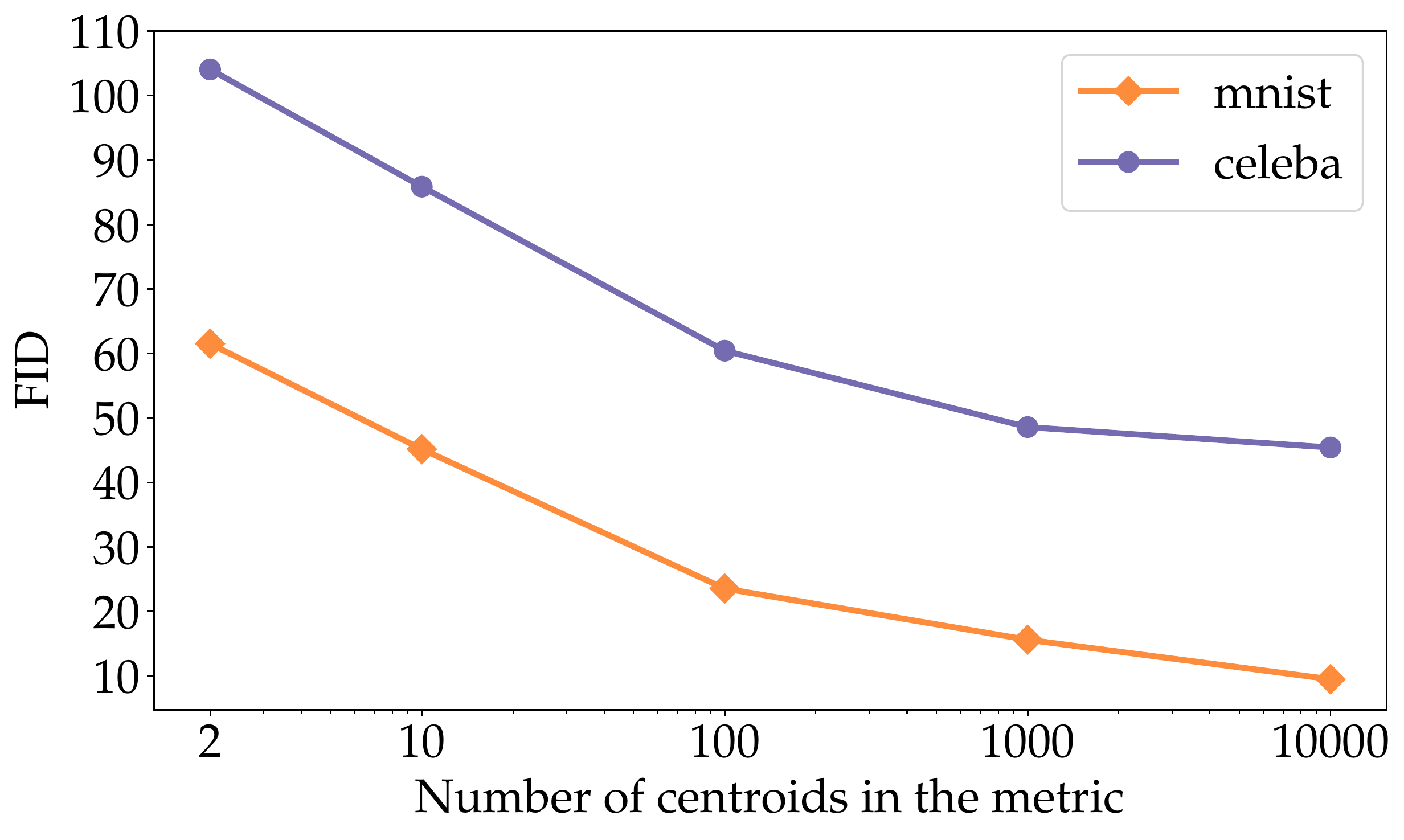}}
    \subfloat{\includegraphics[width=2.7in]{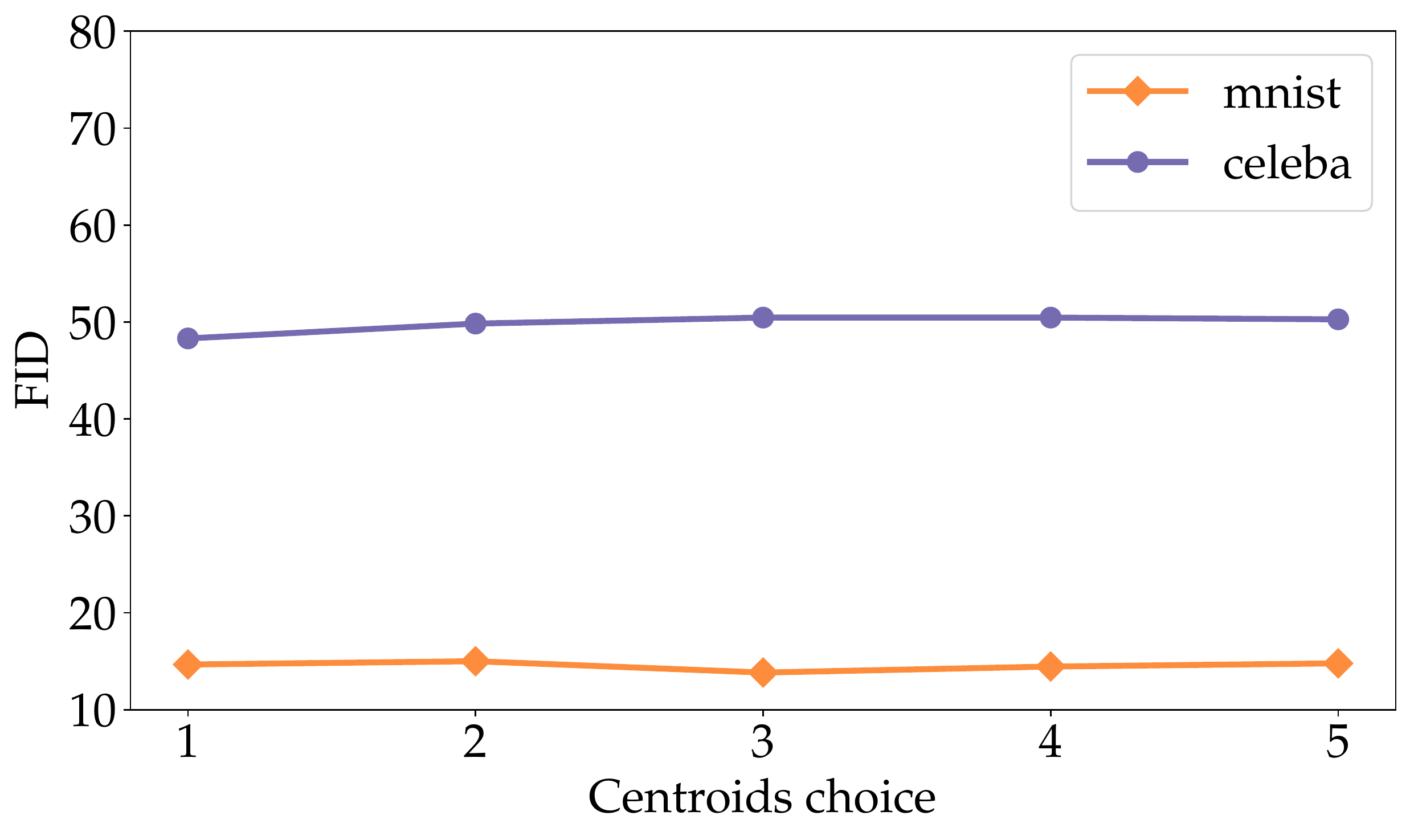}}
    \caption{\emph{Left:} FID score evolution according to the number of centroids in the metric (Eq.~\eqref{eg: Riemannian metric}). \emph{Right}: The FID variation with respect to the choice in centroids. We generate 10000 samples by selecting each time different centroids ($k=1000$).}
    \label{fig:fid evolution with centroids}
    \end{figure}

In order to assess the influence of the number of centroids and their choice in the metric in Eq.~\eqref{eg: Riemannian metric}, we show in Figure~\ref{fig:fid evolution with centroids} the evolution of the FID according to the number of centroids in the metric (left) and the variation of FID according to the choice in the centroids (right). As expected, choosing a small number of centroids will increase the value of the FID since it reduces the variability of the generated samples that will remain \emph{close} to the centroids. Nonetheless, as soon as the number of centroids is higher than 1000 the FID score is either competitive or better than peers and continues decreasing as the number of centroids increases.

\begin{figure}[ht]
    \centering
    \captionsetup[subfigure]{position=above, labelformat = empty}
     \subfloat[\centering
         Decoded centroid]{\includegraphics[width=1.0in]{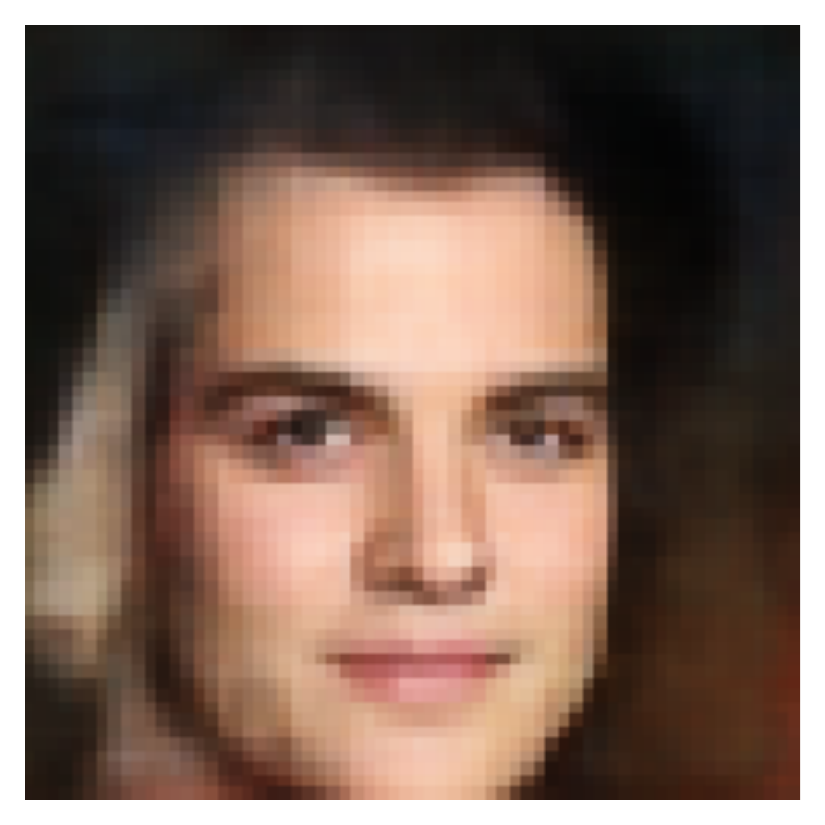}}
    \subfloat[\centering Nearest train image]{\includegraphics[width=1.0in]{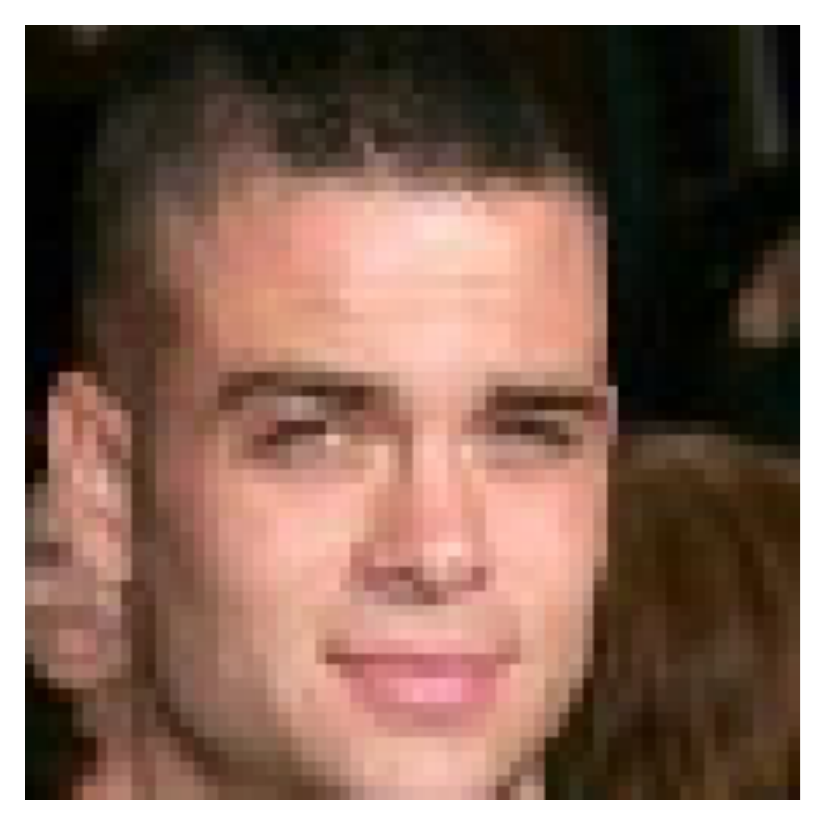}}
    \subfloat[Generated samples]{\includegraphics[width=3.35in]{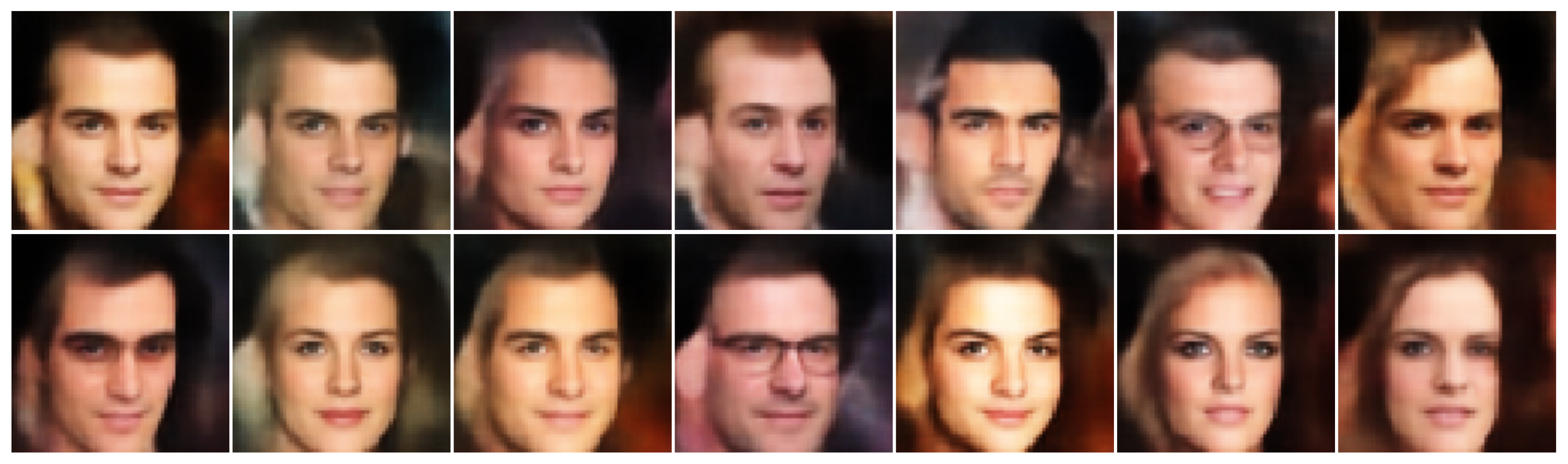}}
    \vspace{-1em}
    \subfloat{\includegraphics[width=1.0in]{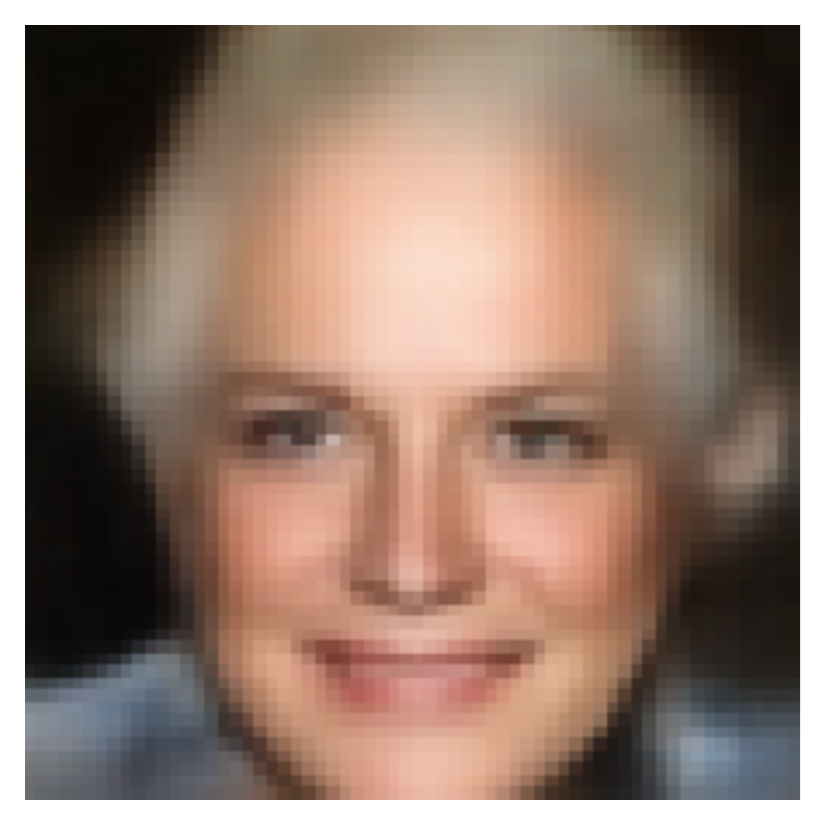}}
    \subfloat{\includegraphics[width=1.0in]{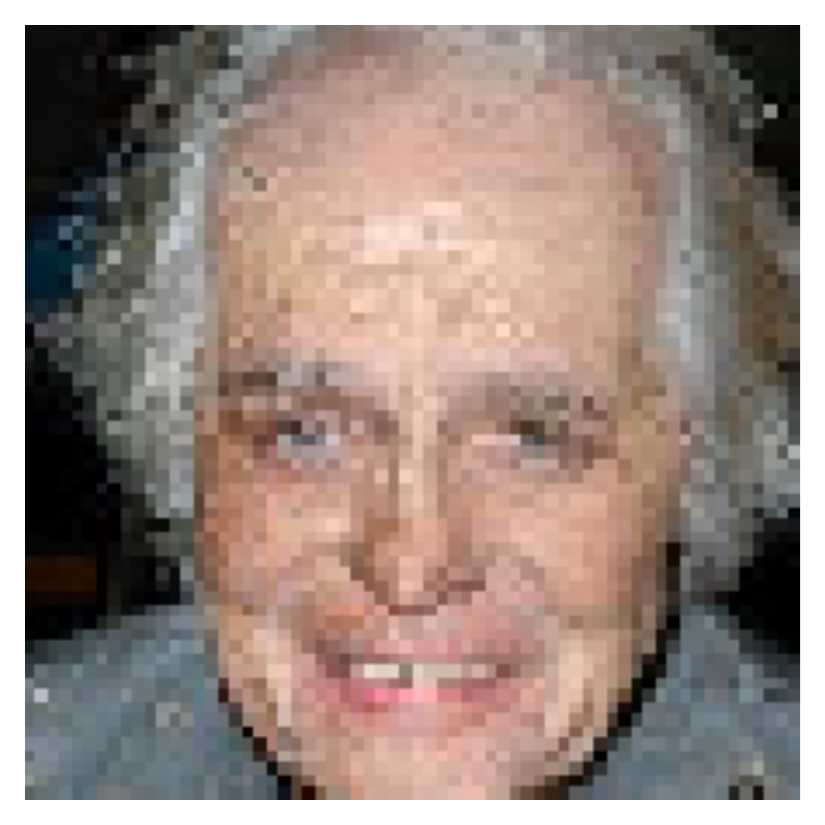}}
    \subfloat{\includegraphics[width=3.35in]{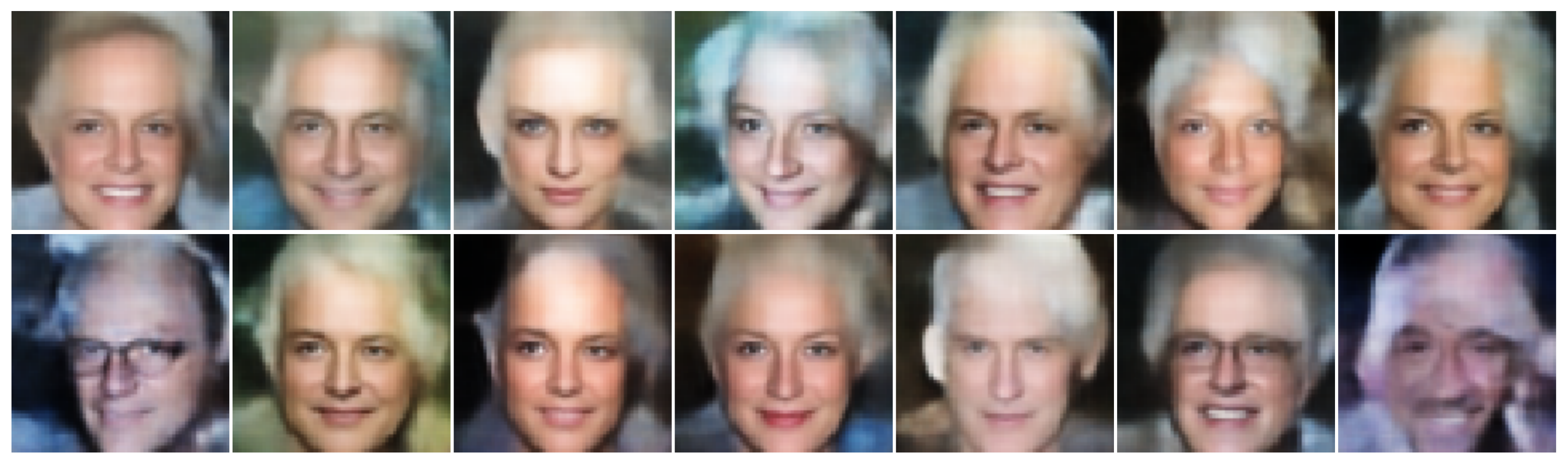}}
    \vfil
    \caption{Variability of the generated samples when only two centroids are considered in the metric. \emph{Left:} The image obtained by decoding the centroids. \emph{Middle}: The nearest image in the train set to the decoded centroids. \emph{Right:} Some generated samples. Each generated sample is assigned to the closest decoded centroid (top row for the first centroid and bottom row for the second one).}
    \label{fig: sampling 2 centroids}
    \end{figure}

To assess the variability of the generated samples, we propose to analyze some generated samples when only 2 centroids are considered. In Figure~\ref{fig: sampling 2 centroids}, we display on the left the decoded centroids along with the closest image to these decoded centroids in the train set. On the right are presented some generated samples. We place these samples in the top row if they are closer to the first decoded centroid and in the bottom row otherwise. Interestingly, even with a small number of centroids the proposed sampling scheme is able to access to a relatively good diversity of samples. These samples are not simply resampled train images or a simple interpolation between selected centroids as some of the generated samples have attributes such as glasses that are not present in the images of the decoded centroids.

\subsection{Influence of $\lambda$ in the metric}

\begin{figure}[ht]
    \centering
    \captionsetup[subfigure]{position=above, labelformat = empty}
\subfloat{\includegraphics[width=2.7in]{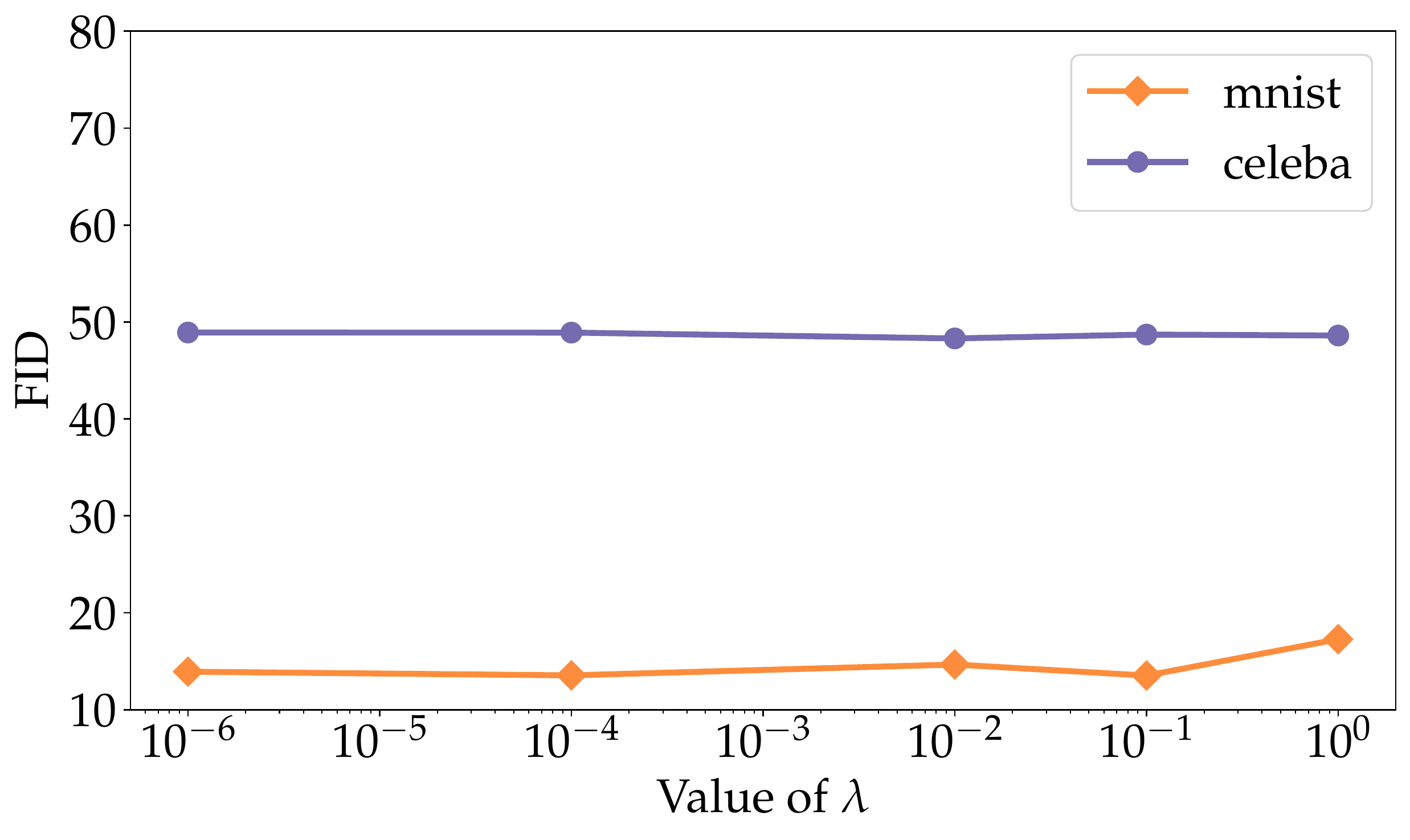}}
    \caption{ FID score evolution according to the value of $\lambda$ in the metric (Eq.~\eqref{eg: Riemannian metric}).}
    \label{fig:fid evolution with lambda}
    \end{figure}

In this section, we also assess the influence of the regularization factor $\lambda$ in Eq.~\eqref{eg: Riemannian metric} on the resulting sampling. To do so, we generate 10k samples using the proposed method on both MNIST and CELEBA datasets for values of $\lambda \in [1e^{-6}, 1e^{-4}, 1e^{-2}, 1e^{-1}, 1]$. Then, we compute the FID against the test set. Each time, we consider $k=1000$ centroids in the metric. As shown in Figure~\ref{fig:fid evolution with lambda}, the influence of $\lambda$ remains limited.  In the implementation, a typical choice for $\lambda$ is $1e^{-2}$.

\subsection{The choice of $\rho$}

In the experiments presented, the smoothing factor $\rho$ in Eq.~\eqref{eg: Riemannian metric} is set to the value of the maximum distance between two closest centroids  $\rho = \max \limits _i \min \limits _{j\neq i} \Vert c_j -c_i\Vert_2$. This choice is motivated by the fact that we wanted to build a smooth metric and so ensure some \emph{smoothness} of the manifold while trying to interpolate faithfully between the metric tensors $\mathbf{G}_i = \mathbf{\Sigma}(x_i)^{-1}$. In particular, a too small value of $\rho$ would have allowed disconnected regions and the sampling may have not prospected well the learned manifold and would have only become a resampling of the centroids. On the other hand, setting a high value for $\rho$ would have biased the interpolation and the value of the metric at a $\mu(x_i)$. As a result, $\mathbf{G}(\mu(x_i))$ might have been very different from the one observed $\mathbf{\Sigma}(x_i)^{-1}$ since the other $\mu(x_j)$ would have had a strong influence on its value. The proposed value for $\rho$ appeared to work well in practice.

\clearpage

\section{Can the method benefit more recent models ?}\label{appG}
Our method proposes to build a Riemannian metric using the covariances in the posterior distributions. Thus, it can be easily plugged into more recent models provided that they have a Gaussian posterior distribution. In order to assess how it would benefit to more recent VAE models, we train a VAMP-VAE \citep{tomczak_vae_2018}, a VAEGAN \citep{larsen_autoencoding_2015}, an Adversarial AE \citep{makhzani2015adversarial} and an IWAE \citep{burda_importance_2016} and compare the generation FID obtained 1) with the prior or 2) when plugging our method. For this experiment, we conduct a hyper-parameter search consisting in training each model with 10 different configurations. For the VAMP we vary the number of pseudo-inputs in \{10, 20, 30, 50, 100, 150, 200, 250, 300, 500\}. For the VAEGAN, we use a discriminator similar to the encoder described in Table.~\ref{Tab: CNN Architectures} and vary the layer depth considered for the reconstruction loss in \{2, 3, 4\} and the factor balancing reconstruction/generation for the decoder's loss in \{0.3, 0.5, 0.7, 0.8, 0.9, 0.99, 0.999\}. For the AAE, we change the factor balancing the reconstruction loss and the regularization in \{0.001, 0.01, 0.1, 0.25, 0.5, 0.75, 0.9, 0.95, 0.99, 0.999\}. Finally, for the IWAE, we vary the number of importance samples in \{2, 3, 4, 5, 6, 7, 8, 9, 10, 12\}. For each model and generation scheme, we report the results of the model achieving the lowest FID on the validation set. According to Table.~\ref{tab:benefit others}, the proposed generation method seems to benefit these models in almost all cases since the FID decreases when compared to the prior-based generation.

\begin{table}[ht]
    \caption{FID (lower is better) vs. the test set using either the prior (classic approach) or by plugging our generation method.}
    \vskip 0.15in
    \centering
    \scriptsize
    \begin{sc}
    \begin{tabular}{lc|cc}
    \toprule
    Model                & Generation  & MNIST & CELEBA \\
    \midrule
    \multirow{2}{*}{VAMP} & prior       & 34.5  & 67.2 \\
                          & ours        & \textbf{32.7} & \textbf{60.9}\\
                          \midrule
    \multirow{2}{*}{IWAE} & prior       & \textbf{32.4} & 67.6  \\
                          & ours        & 33.8          & \textbf{60.3}\\
    \midrule
    \multirow{2}{*}{AAE}  & prior       & 19.1  & 64.8  \\
                          & ours        & \textbf{11.7} & \textbf{51.4} \\
    \midrule
    \multirow{2}{*}{VAEGAN} & prior       & 8.7 & 39.7  \\
                          & ours          & \textbf{6.1} & \textbf{31.4} \\
    \bottomrule

    \end{tabular}
    \end{sc}
    \label{tab:benefit others}
\end{table}

Another approach that is interesting to compare to is the 2-stage VAE model proposed in \citep{dai_diagnosing_2018}. Our method can indeed be seen as part of the methods trying to counterbalance the poor expressiveness of the prior distribution. In \citep{dai_diagnosing_2018}, the authors argue that the actual distribution of the latent codes  (i.e. the aggregated posterior) is "likely not close to a standard Gaussian distribution" \citep{dai_diagnosing_2018} leading to a distribution mismatch degrading the generation capability of the model. To address this issue, they propose to use a second VAE to estimate the learned distribution of the latent variables. Our approach starts with the same observation that the latent codes have no reason to follow the prior. However, it differs since we propose to adopt a fully geometric perspective and propose instead a sampling scheme using the intrinsic uniform distribution defined on the learned Riemannian manifold.

We nonetheless compare our method with models obtained with the official implementation provided by the authors of \citep{dai_diagnosing_2018} on MNIST and CELEBA. To allow a fair comparison, we simply plug our method to the obtained trained models and build the metric using the posteriors coming from the 1$^{\mathrm{st}}$ stage VAE. In Table.~\ref{tab:compare 2stage}, we compare the FID obtained 1) with the first stage VAE (\emph{i.e.} prior), 2) with the second stage VAE \citep{dai_diagnosing_2018} and 3) with our method. Again, our proposed generation method allows to achieve lower FID results.

\begin{table}[ht]
    \caption{FID (lower is better) vs. the test set using the 2-stage VAE implementation \citep{dai_diagnosing_2018} for either the reconstructed samples (recon.), using the prior (1$^{\mathrm{st}}$ stage), using the 2-stage approach (2$^{\mathrm{nd}}$ stage) or by plugging our generation method.}
    \vskip 0.15in
    \centering
    \scriptsize
    \begin{sc}
    \begin{tabular}{cccccc}
    \toprule
    Dataset               & Nets  & Recon. & 1$^{\mathrm{st}}$ stage & 2$^{\mathrm{nd}}$ stage & Ours\\
    \midrule
    MNIST & similar to \citep{chen2016infogan} & 14.8 & 20.0  &   12.9 & \textbf{9.9} \\
    CELEBA & similar to \citep{chen2016infogan} & 44.9 & 67.8 & 53.3 & \textbf{49.6} \\
    CELEBA & similar to \citep{tolstikhin2018wasserstein} & 34.3 & 70.8 & 40.7 & \textbf{37.9}\\
    \bottomrule

    \end{tabular}
    \end{sc}
    \label{tab:compare 2stage}
\end{table}

\medskip

\end{document}